\documentclass{article} %
\usepackage{time_jepa_preprint,times}

\usepackage{amsmath,amsfonts,bm}

\DeclareMathSizes{8.5}{9}{6}{5}

\def\eqref#1{equation~\ref{#1}}

\def\1{\bm{1}}

\DeclareMathAlphabet{\mathsfit}{\encodingdefault}{\sfdefault}{m}{sl}
\SetMathAlphabet{\mathsfit}{bold}{\encodingdefault}{\sfdefault}{bx}{n}

\newcommand{\Ljepa}{\mathcal{L}_{\mathrm{JEPA}}}
\newcommand{\Ltime}{\mathcal{L}_{\mathrm{time}}}

\newcommand{\appmetric}[1]{\makebox[1.65em][c]{\ensuremath{#1}}\ensuremath{\uparrow}}

\usepackage{hyperref}
\usepackage{url}
\usepackage{graphicx}
\usepackage{booktabs}
\usepackage{colortbl}
\usepackage{tabularx}
\usepackage{longtable}
\usepackage{placeins}
\usepackage{flafter}
\usepackage{multirow}

\definecolor{appendixcardborder}{rgb}{0.78,0.82,0.87}
\definecolor{appendixcardbackground}{rgb}{0.97,0.98,0.99}
\definecolor{appendixcardtitle}{rgb}{0.16,0.30,0.43}
\newcommand{\appendixcard}[2]{%
    \begingroup
    \setlength{\fboxsep}{6pt}%
    \setlength{\fboxrule}{0.4pt}%
    \noindent\fcolorbox{appendixcardborder}{appendixcardbackground}{%
        \begin{minipage}{\dimexpr\linewidth-2\fboxsep-2\fboxrule\relax}
            \small\textbf{\color{appendixcardtitle}#1}\par\vspace{2pt}%
            #2
        \end{minipage}%
    }%
    \endgroup
}

\newcommand{\appendixcarditem}[2]{\noindent\textbf{#1} #2\par}

\title{Time-JEPA: Grounding Video World Models with Physical Time Series}

\author{%
Shenyi Li$^{1,3,*}$ \quad Ziyang Zhang$^{2,3,*}$ \quad Wen Wu$^{3}$\\
{\bfseries Ming Liu$^{1}$ \quad Chao Zhang$^{2,3,\dagger}$}\\[0.6em]
{\normalfont\small $^{1}$Harbin Institute of Technology}\\
{\normalfont\small $^{2}$Tsinghua University}\\
{\normalfont\small $^{3}$Shanghai Artificial Intelligence Laboratory}\\[0.4em]
{\normalfont\small $^{*}$Equal contribution. \quad $^{\dagger}$Corresponding author.}\\[0.4em]
{\normalfont\footnotesize \href{mailto:lishenyi@pjlab.org.cn}{\nolinkurl{lishenyi@pjlab.org.cn}},
\href{mailto:ziyang-z24@mails.tsinghua.edu.cn}{\nolinkurl{ziyang-z24@mails.tsinghua.edu.cn}}}\\
{\normalfont\footnotesize \href{mailto:wuwen@pjlab.org.cn}{\nolinkurl{wuwen@pjlab.org.cn}},
\href{mailto:liuming1981@hit.edu.cn}{\nolinkurl{liuming1981@hit.edu.cn}}}\\
{\normalfont\footnotesize Correspondence: \href{mailto:cz277@tsinghua.edu.cn}{\nolinkurl{cz277@tsinghua.edu.cn}}}
}

\hypersetup{
    pdftitle={Time-JEPA: Grounding Video World Models with Physical Time Series},
    pdfauthor={Shenyi Li, Ziyang Zhang, Wen Wu, Ming Liu, Chao Zhang},
    pdfsubject={Preprint}
}

\begin{document}

\maketitle
\begin{center}
\small Code: \url{https://github.com/Lishenyi2004/Time-JEPA}
\end{center}

\begin{abstract}
Video world models learn how visual environments evolve by predicting missing
or future content. However, physical-state information can provide useful context for predicting how environments evolve. Prior work incorporates this information through different mechanisms: domain-specific physical laws can complicate multi-domain training, while supervision on individual frames or frame pairs gives limited attention to how physical state evolves over time. We
introduce \textbf{Time-JEPA}, which fine-tunes a video world model using
low-dimensional, temporally aligned physical time series as auxiliary supervision.
Physical time series connect video representations with physical state at
corresponding temporal positions, encouraging the model to preserve how
physical state evolves. In multiple domains including autonomous driving,
robot manipulation, and human motion, Time-JEPA improves current-state
decoding, multi-horizon future-state prediction, and kinematic reconstruction
over matched video-fine-tuned controls across both condition and state-variable
distribution shifts. Multi-domain training provides aggregate gains over
matched single-domain training in all three tasks. External evaluations further
suggest that time-series supervision improves sensitivity to physical anomalies
and the representation of physical interactions, including in unseen scenarios.
Aggregate gains on a second video backbone indicate that the method can be applied to more backbones.
These results suggest that physical time series help video world models better
understand the physical world.
\end{abstract}

\section{Introduction}
\label{sec:intro}

Video world models learn how visual environments evolve by predicting missing
or future content. Common approaches include masked video autoencoders, which
reconstruct hidden pixels
\citep{tong2022videomae,feichtenhofer2022spatiotemporal}, and latent-prediction
video models such as V-JEPA \citep{bardes2024vjepa,assran2025vjepa2}. V-JEPA
builds on the joint-embedding predictive architecture (JEPA)
\citep{assran2023ijepa} to predict masked spatiotemporal target representations
from visible video context. The resulting representations encode objects,
motion, and events. However, predicting how an event unfolds also requires
information about physical state. A car can follow the same turn at different
speeds and produce different future motion, and visually similar grasps can
involve different joint configurations or forces. The speeds, joint
configurations, and forces in these examples form part of the physical state
and help video world models understand how events unfold.

Prior work incorporates physical state into video world models in two main ways.
Some methods use physical laws or system-specific dynamical priors to constrain
how physical state evolves \citep{leguen2020phydnet,mao2026piwm}. Because these
constraints rely on domain-specific formulations, equations, and knowledge,
one training design cannot easily cover multiple domains. Moreover, certain
physical constraints can make the model focus on the dynamics they describe
and miss other changes in videos. Others use geometric structure, motion cues,
or privileged state as prediction or alignment targets to capture
different aspects of physical state
\citep{mustafa2024dynamics,shang2025roboscape,shen2026phantom,zeng2026phylatent,yan2026psgjepa,wen2026xpjepa}. This way reduces reliance
on explicit physical laws. However, this way often focuses supervision on individual frames or frame pairs, giving limited attention to how physical state evolves within a video’s full temporal context. Besides, such methods often require additional input signals, making the training setup more complex. Therefore, overcoming these limitations requires a common interface and
readily available data that can serve as physical state.

We introduce \textbf{Time-JEPA}, a method that aligns physical-state with the temporal structure of video. We
apply a shared interface for multiple domains and use physical time series
that sensors often easily provide alongside video.
It is noted that time series only serve as auxiliary supervision during training and video remains as the model's
only input during evaluation. We evaluate Time-JEPA in
multiple domains including autonomous driving, robot manipulation, and human
motion. Compared with matched video-fine-tuned controls, Time-JEPA improves
current-state decoding, multi-horizon future-state prediction, and kinematic
reconstruction. The gains hold under two levels of
out-of-distribution evaluation: unseen test conditions, and state variables
not used as auxiliary training targets. Multi-domain training provides
aggregate gains over matched single-domain training in all three tasks. External
evaluations further suggest that time-series supervision improves sensitivity
to physical anomalies and the representation of physical interactions, including
in unseen scenarios. Aggregate gains on a second video backbone indicate that the method can be applied to more backbones. These results indicate that physical time series help
video world models better understand the physical world.

\begin{samepage}
We make two contributions:
\begin{enumerate}
    \item \textbf{Physical time-series supervision for video world models.}
    We introduce Time-JEPA, which combines latent prediction with temporally
    aligned physical time-series supervision through a shared interface.
    We set up in-distribution (ID) and out-of-distribution (OOD) settings at
    two levels, including conditions and variables.

    \item \textbf{Datasets and toolkit in multiple domains.}
    We construct datasets for multiple domains including autonomous driving,
    robot manipulation, and human motion. We develop a toolkit for video
    world models that combines current-state decoding, multi-horizon
    future-state prediction, and kinematic reconstruction to more thoroughly
    assess their ability to represent physical states, including physical
    relations and state evolution.
\end{enumerate}
\end{samepage}

\section{Related Work}
\label{sec:related_work}

\paragraph{Predictive video representations and world models.}
Video world models learn representations of visual environments through pixel
reconstruction or latent prediction.
For pixel reconstruction, masked video autoencoders
\citep{tong2022videomae,feichtenhofer2022spatiotemporal} reconstruct hidden
pixels, and VideoMAE~V2~\citep{wang2023videomaev2} scales this objective through
dual masking.
For latent prediction, V-JEPA~\citep{bardes2024vjepa} follows
JEPA~\citep{assran2023ijepa} and predicts target-encoder representations at
masked spatiotemporal positions.
V-JEPA~2~\citep{assran2025vjepa2} and
V-JEPA~2.1~\citep{murlabadia2026vjepa2_1} develop this family for motion
understanding and dense feature extraction.
InternVideo-Next~\citep{wang2025internvideonext} combines reconstruction and
latent prediction.
Dreamer~\citep{hafner2020dreamer} and iVideoGPT~\citep{wu2024ivideogpt} learn
action-conditioned dynamics for control and interactive video prediction.
Time-JEPA builds on V-JEPA's masked latent-prediction objective and adds
time-series supervision during fine-tuning.

\paragraph{Signals for visual learning.}
Prior work uses auxiliary supervision from scene state, egomotion, and other
modalities to shape visual representations
\citep{chen2020learningbycheating,agrawal2015moving,gupta2016crossmodal}.
One route incorporates physical equations or dynamical priors into latent
prediction, as in PhyDNet~\citep{leguen2020phydnet} and
PIWM~\citep{mao2026piwm}.
A second route adds prediction or alignment targets, including physical state
in Dynamics Mapping~\citep{mustafa2024dynamics}, depth and keypoint trajectories
in RoboScape~\citep{shang2025roboscape}, and V-JEPA-derived latents in
PHANTOM~\citep{shen2026phantom}.
PhyLatent~\citep{zeng2026phylatent} and PSG-JEPA~\citep{yan2026psgjepa} add
physical-state supervision to action-conditioned JEPA prediction, while
JEPA-x~\citep{wen2026xpjepa} couples visual and physical trajectories through
cross-modal future prediction.
Time-JEPA uses low-dimensional, temporally aligned physical time series as
auxiliary supervision through a shared interface across autonomous
driving, robot manipulation, and human motion.
Sensors or synchronized annotations supply these time series, and the encoder uses
video as its only input during fine-tuning and evaluation.

\begin{figure}[!t]
    \centering
    \includegraphics[width=\textwidth]{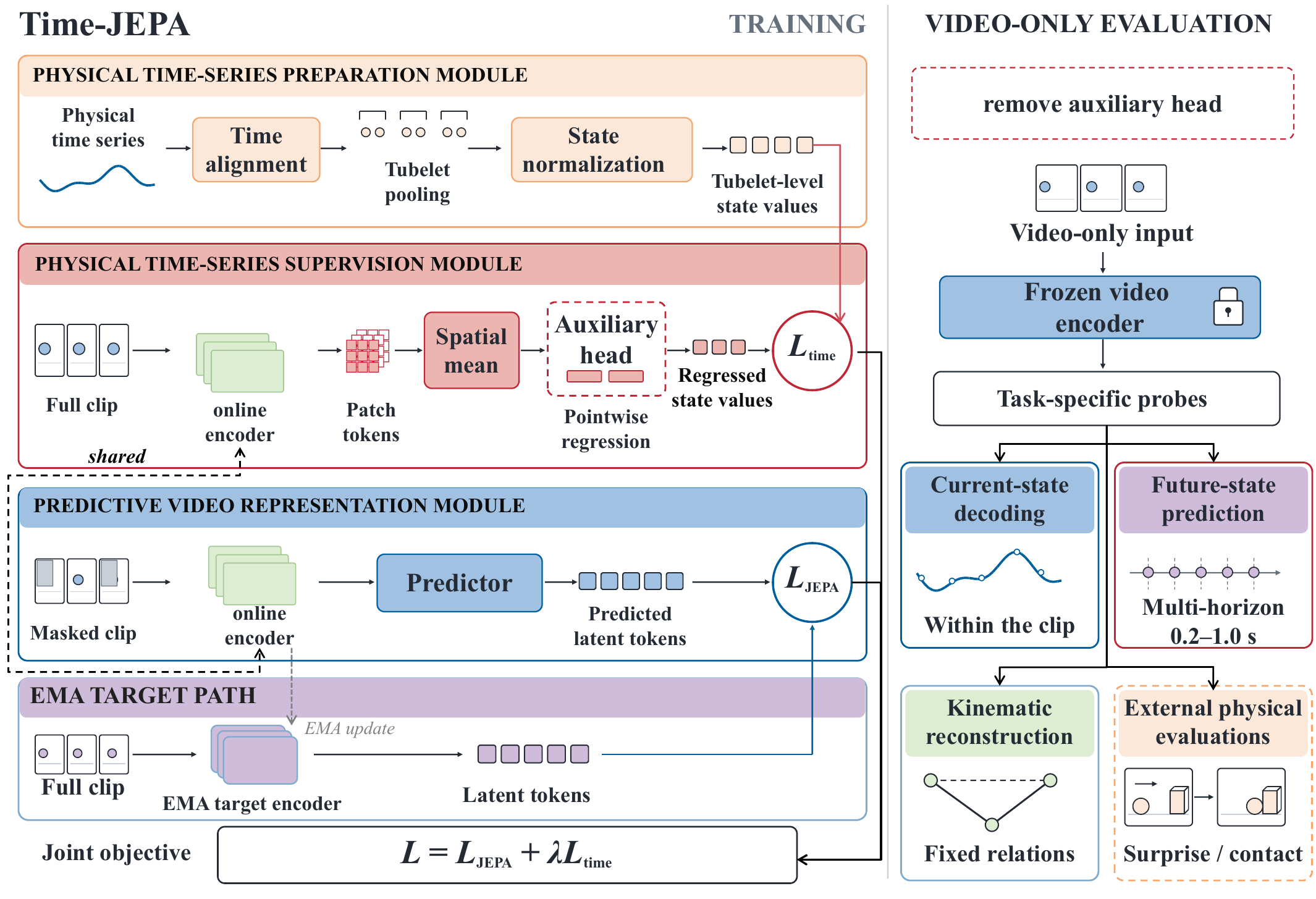}
    \caption{\textbf{Time-JEPA training and evaluation.}
    Training combines masked latent prediction with regression of aligned,
    pooled, and normalized physical state values from full-clip video features.
    The EMA target encoder supplies latent supervision. Video-only evaluation
    covers current-state decoding, multi-horizon future-state prediction,
    kinematic reconstruction, and external physical evaluations.}
    \label{fig:time_jepa_overview}
\end{figure}

\paragraph{Evaluating physical information in video representations.}
PHYRE~\citep{bakhtin2019phyre} and CLEVRER~\citep{yi2020clevrer} evaluate
physical reasoning about interacting objects and collision events.
Physion~\citep{bear2021physion} tests future-contact prediction, and
GRASP~\citep{jassim2024grasp} evaluates intuitive-physics properties with
matched possible and impossible videos.
Studies of self-supervised video predictors examine intuitive physics and
probe physical-state information in frozen video
representations \citep{garrido2025intuitivephysics,
joseph2026interpretingphysics,punzo2026videophysics}.
Our evaluation combines continuous current-state decoding, kinematic
reconstruction, and multi-horizon future-state prediction with external tests on
GRASP and Physion.

\section{Method: Time-JEPA}
\label{sec:method}

\paragraph{Overview.}
We propose Time-JEPA, a fine-tuning approach that uses temporally aligned
physical time-series supervision to encourage predictive video representations
to preserve information about physical state as it evolves.
As shown in Figure~\ref{fig:time_jepa_overview}, Time-JEPA couples two learning
paths through a shared online video encoder.
The predictive video representation module builds on V-JEPA and predicts latent
representations at masked positions. An exponential-moving-average (EMA) target
encoder computes these representations from the full clip.
The physical time-series preparation module maps synchronized measurements to video
time, and the physical time-series supervision module regresses the aligned
state from full-clip video features.
Both objectives update the same online encoder, coupling masked latent
prediction with tubelet-level physical-state supervision.
The physical time series provide auxiliary supervision during fine-tuning, and
video remains the encoder input throughout fine-tuning and evaluation.

\subsection{Predictive Video Representation}
\label{sec:method_preliminaries}

The predictive video representation module builds on the masked
latent-prediction objective of V-JEPA~\citep{bardes2024vjepa}.
For each clip, the online encoder transforms a masked video view into latent
patch tokens.
Its final-layer representation indexes these tokens by temporal position and
spatial patch location.
The predictor estimates latent tokens at the masked positions.
The EMA target encoder processes the full clip and provides the corresponding
latent tokens.
The JEPA loss $\Ljepa$ measures prediction error at the masked positions and
updates the online encoder and predictor.
The exponential moving average updates the target encoder parameters from the
online encoder.

\subsection{Physical Time-Series Preparation}
\label{sec:method_state_targets}

The physical time-series preparation module maps physical time series from sensors or
synchronized annotations to the encoder's temporal grid.
These time series record measurements such as vehicle speed, robot joint
configuration, and human joint angle over time.
We associate each sampled video time with nearby physical measurements using
timestamps or frame indices, then group the aligned measurements according to
the encoder's tubelet structure.
Within each tubelet, we pool each type of physical measurement across its sampled times.
We then normalize the pooled measurements to place different numerical scales
in a common space, producing a sequence of tubelet-level state values.
This preparation aligns each temporal video feature with the corresponding
physical measurements even when their sampling rates differ.
Appendix~\ref{app:time_series_preprocessing} gives the shared alignment,
pooling, validity, and normalization details.

\subsection{Physical Time-Series Supervision}
\label{sec:method_time_supervision}

In the physical time-series supervision module, we process the unmasked video
clip with the shared online encoder.
The online encoder's final layer organizes patch tokens along temporal
positions and spatial patch locations.
At each temporal position, Time-JEPA then applies mean pooling across spatial
patch locations to produce one pooled temporal feature.
An auxiliary physical-state regression head applies a two-layer pointwise nonlinear
projection independently to each pooled temporal feature and regresses
normalized state values for the corresponding tubelet.
This minimal pointwise design concentrates temporal representation learning in
the online encoder.
We compute the time-series regression loss $\Ltime$ as a validity-masked
Smooth-L1 error between the auxiliary-head outputs and their aligned state values.

We combine the two learning paths through the joint objective
$\mathcal L=\Ljepa+\lambda\Ltime$, where $\lambda$ controls the contribution of
the time-series loss.
Under this objective, the JEPA loss guides masked latent prediction, while the
time-series loss encourages the online encoder to model how physical state
evolves across the video clip.

\subsection{Evaluation with Video-Only Input}
\label{sec:method_evaluation}

After fine-tuning, we discard the auxiliary head and freeze the online encoder.
We train state probes on its final-layer features from video-only input.
For current-state decoding and future-state prediction, these probes apply
spatial mean pooling followed by two pointwise convolutional layers.
Current-state decoding measures how accurately the encoders recover physical states
within the observed clip.
Multi-horizon future-state prediction estimates physical states at multiple
times beyond the observed clip.
Kinematic reconstruction evaluates physical relations among the outputs of
separate state probes.
External physical evaluations assess transfer to physical tasks beyond the
fine-tuning datasets.
Sections~\ref{sec:experimental_protocol}--\ref{sec:external_transfer} detail the
evaluation protocols.

\section{Experimental Setup}
\label{sec:experimental_protocol}

\setlength{\textfloatsep}{0pt plus 0pt minus 0pt}
\begin{figure}[!t]
    \centering
    \includegraphics[width=0.92\textwidth]{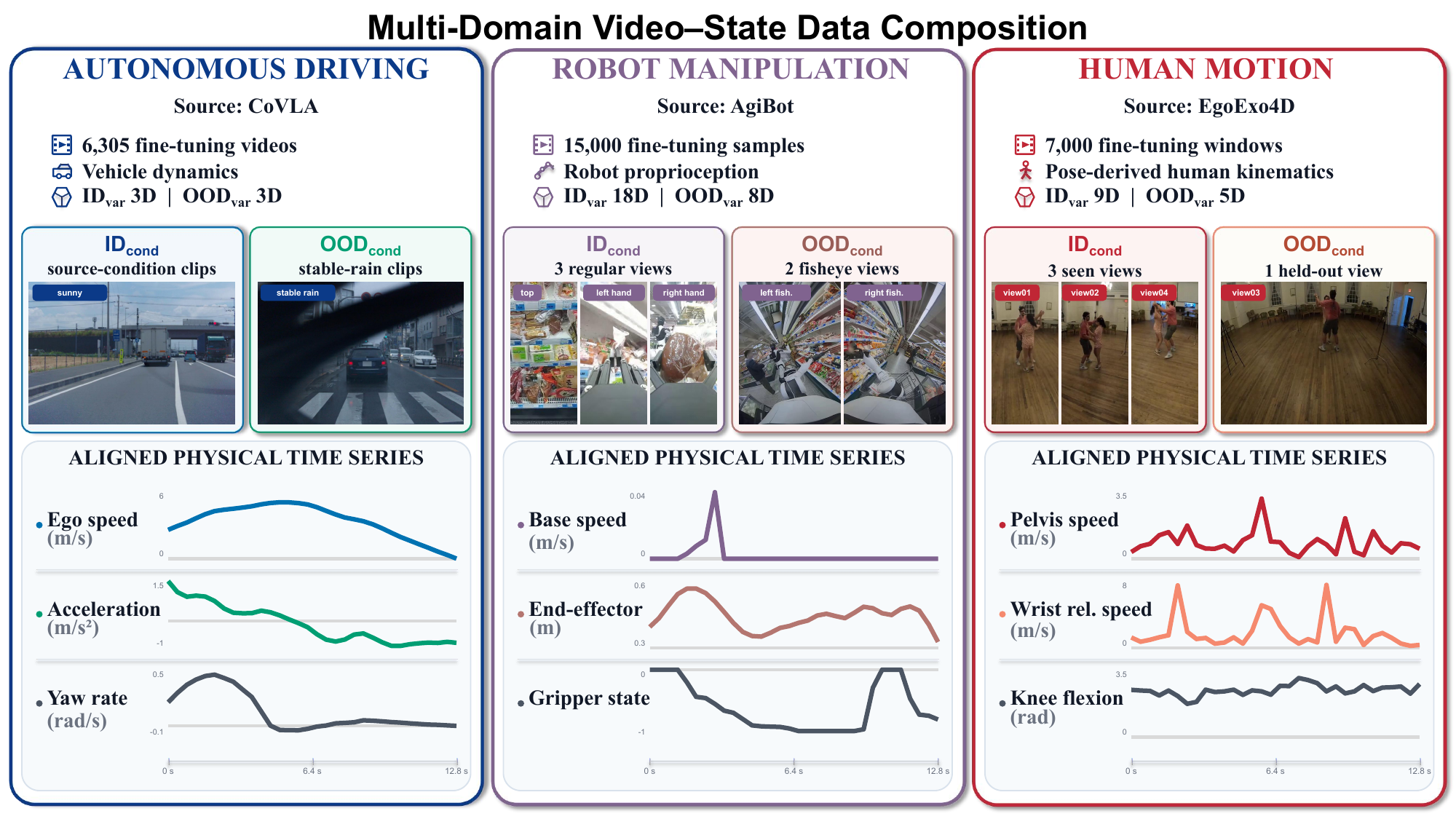}
    \caption{\textbf{Video--state data composition.}
    Each column summarizes one domain's video data for fine-tuning,
    two-level ID/OOD settings for conditions and state variables, and
    representative physical time series.}
    \label{fig:data_composition}
\end{figure}
\vspace{3pt}

\subsection{Comparisons}
\label{sec:protocol_tasks}

To evaluate the effects of video fine-tuning and time-series supervision,
we compare three configurations based on the released V-JEPA~2.1 ViT-g/16
model~\citep{murlabadia2026vjepa2_1}: (i)
The pretrained V-JEPA model (\emph{original V-JEPA});
(ii) V-JEPA fine-tuned  using video data and
$\Ljepa$ (\emph{video-fine-tuned control});
(iii) \emph{Time-JEPA} that uses the same video data and fine-tuning protocol, while adds
supervision from temporally aligned physical time series and optimizes the
time-series regression loss $\Ltime$ together with $\Ljepa$.

\subsection{Domains, Variables, and Splits}
\label{sec:protocol_domains}

We study autonomous driving, robot manipulation, and human motion using the
CoVLA~\citep{arai2025covla}, AgiBot~\citep{bu2025agibot}, and
EgoExo4D~\citep{grauman2024egoexo4d} datasets, respectively.
Across the three domains, we pair video with physical time series and define two
ID/OOD axes: conditions and state variables.
Seen and held-out conditions or camera views define
$\mathrm{ID}_{\mathrm{cond}}$ and $\mathrm{OOD}_{\mathrm{cond}}$, respectively.
State variables used and not used in auxiliary time-series supervision define
$\mathrm{ID}_{\mathrm{var}}$ and $\mathrm{OOD}_{\mathrm{var}}$, respectively.

Autonomous driving uses speed and acceleration for vehicle dynamics,
caption-defined stable rain for $\mathrm{OOD}_{\mathrm{cond}}$, and steering
and lateral motion for $\mathrm{OOD}_{\mathrm{var}}$.
Robot manipulation uses robot motion and gripper state for robot proprioception,
held-out fisheye views for $\mathrm{OOD}_{\mathrm{cond}}$, and joint state,
effort, and wrench for $\mathrm{OOD}_{\mathrm{var}}$.
Human motion uses body motion and joint angles for pose-derived human kinematics,
the held-out exocentric \texttt{view03} for $\mathrm{OOD}_{\mathrm{cond}}$, and
pose-derived body geometry for $\mathrm{OOD}_{\mathrm{var}}$.
Figure~\ref{fig:data_composition} summarizes the video data and physical
time series.
Appendix~\ref{app:datasets} gives the state-variable definitions and split
construction.

\subsection{Common Experimental Settings}
\label{sec:protocol_probes}
\label{sec:implementation_details}

We sample 64 frames at 5~FPS per clip for fine-tuning and evaluation in multiple domains.
We use task-specific probes throughout our evaluations.
We first train each probe on the probe-training split and then test it on the
corresponding test split.
We normalize state values in the probe-training, probe-validation, and test
splits using statistics from the probe-training split to keep each variable
on a consistent scale. Appendix~\ref{app:reproducibility} gives the full fine-tuning and probe protocols.

For multi-domain Time-JEPA, we share one encoder in multiple domains and
assign an auxiliary head to each domain.
each domain follows the same per-domain training settings as its matched single-domain run, while the shared encoder processes the combined stream.
Appendix~\ref{app:mix_details} gives the multi-domain scheduling.

\section{Physical-State Decoding and Prediction}
\label{sec:single_domain_results}

\begin{figure}[!b]
    \centering
    \includegraphics[width=0.90\textwidth]{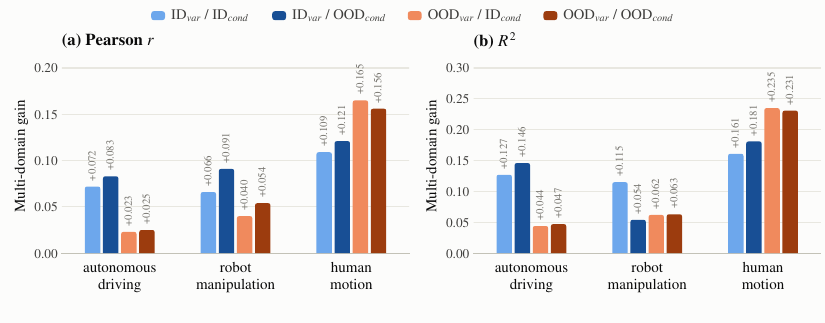}
    \caption{\textbf{Current-state decoding gains from multi-domain training.}
    (a) Pearson $r$ gains and (b) $R^2$ gains of multi-domain Time-JEPA over
    matched single-domain training. Colors identify the four
    variable/condition settings in the legend. Labels report absolute gains.}
    \label{fig:multidomain_gain}
\end{figure}

\begin{table}[!tb]
    \caption{\textbf{Current-state decoding across variables and conditions.}
    We compare the three V-JEPA configurations on
    $\mathrm{ID}_{\mathrm{var}}$ and $\mathrm{OOD}_{\mathrm{var}}$ state
    variables under $\mathrm{ID}_{\mathrm{cond}}$ and
    $\mathrm{OOD}_{\mathrm{cond}}$ conditions.
    We compute metrics for each variable and camera view, then average equally
    over views and over variables.
    Bold marks the highest score in each comparison.
    Appendix~\ref{app:target_matrices} gives variable-level and domain results.}
    \label{tab:single_domain_main}
    \centering
    \fontsize{8.5}{10}\selectfont
    \setlength{\tabcolsep}{1.75pt}
    \begin{tabular*}{\textwidth}{@{\extracolsep{\fill}}lllccc@{\hspace{8pt}}ccc@{\hspace{8pt}}ccc@{}}
        \toprule
        & & & \multicolumn{3}{c}{\textbf{autonomous driving}}
          & \multicolumn{3}{c}{\textbf{robot manipulation}}
          & \multicolumn{3}{c}{\textbf{human motion}} \\
        \cmidrule(lr){4-6}\cmidrule(lr){7-9}\cmidrule(l){10-12}
        Condition & Variables & Metric
        & Original & \shortstack{Video\\[-1pt]FT} & \shortstack{Time-\\[-1pt]JEPA}
        & Original & \shortstack{Video\\[-1pt]FT} & \shortstack{Time-\\[-1pt]JEPA}
        & Original & \shortstack{Video\\[-1pt]FT} & \shortstack{Time-\\[-1pt]JEPA} \\
        \midrule
        \multirow{4}{*}{\textbf{$\mathrm{ID}_{\mathrm{cond}}$}}
        & \multirow{2}{*}{$\mathrm{ID}_{\mathrm{var}}$}
               & \makebox[1.2em][r]{$r$}\(\uparrow\)
               & 0.804 & 0.817 & \textbf{0.916}
               & 0.609 & 0.639 & \textbf{0.845}
               & 0.500 & 0.503 & \textbf{0.589} \\
        & & \makebox[1.2em][r]{$R^2$}\(\uparrow\)
               & 0.698 & 0.713 & \textbf{0.849}
               & 0.389 & 0.424 & \textbf{0.725}
               & 0.228 & 0.231 & \textbf{0.345} \\
        \addlinespace[2pt]
        & \multirow{2}{*}{$\mathrm{OOD}_{\mathrm{var}}$}
               & \makebox[1.2em][r]{$r$}\(\uparrow\)
               & 0.879 & 0.885 & \textbf{0.920}
               & 0.641 & 0.654 & \textbf{0.743}
               & 0.434 & 0.434 & \textbf{0.504} \\
        & & \makebox[1.2em][r]{$R^2$}\(\uparrow\)
               & 0.771 & 0.783 & \textbf{0.846}
               & 0.422 & 0.440 & \textbf{0.572}
               & 0.132 & 0.133 & \textbf{0.222} \\
        \midrule
        \multirow{4}{*}{\textbf{$\mathrm{OOD}_{\mathrm{cond}}$}}
        & \multirow{2}{*}{$\mathrm{ID}_{\mathrm{var}}$}
               & \makebox[1.2em][r]{$r$}\(\uparrow\)
               & 0.800 & 0.812 & \textbf{0.902}
               & 0.236 & 0.243 & \textbf{0.492}
               & 0.480 & 0.486 & \textbf{0.581} \\
        & & \makebox[1.2em][r]{$R^2$}\(\uparrow\)
               & 0.685 & 0.699 & \textbf{0.824}
               & 0.002 & \(-\)0.001 & \textbf{0.085}
               & 0.189 & 0.199 & \textbf{0.332} \\
        \addlinespace[2pt]
        & \multirow{2}{*}{$\mathrm{OOD}_{\mathrm{var}}$}
               & \makebox[1.2em][r]{$r$}\(\uparrow\)
               & 0.894 & 0.901 & \textbf{0.937}
               & 0.373 & 0.402 & \textbf{0.555}
               & 0.450 & 0.452 & \textbf{0.515} \\
        & & \makebox[1.2em][r]{$R^2$}\(\uparrow\)
               & 0.799 & 0.812 & \textbf{0.876}
               & 0.093 & 0.127 & \textbf{0.205}
               & 0.127 & 0.135 & \textbf{0.225} \\
        \bottomrule
    \end{tabular*}
\end{table}

\subsection{Time-Series Supervision Across Variables and Conditions}
\label{sec:single_domain_main}

This section investigates how accurately the encoders recover physical states
within the observed clip.
We compare the three configurations from Section~\ref{sec:protocol_tasks} across
the state-variable and condition axes defined in Section~\ref{sec:protocol_domains}.
Table~\ref{tab:single_domain_main} reports Pearson $r$ and $R^2$ in three domain
column groups, with $\mathrm{ID}_{\mathrm{cond}}$ results in the upper block and
$\mathrm{OOD}_{\mathrm{cond}}$ results in the lower block.

\setlength{\textfloatsep}{0pt plus 0pt minus 0pt}
\vspace{3pt}

Across all three domains, Time-JEPA improves both Pearson $r$ and $R^2$ over
the video-fine-tuned control for $\mathrm{ID}_{\mathrm{var}}$ under
$\mathrm{ID}_{\mathrm{cond}}$.
These gains extend to $\mathrm{OOD}_{\mathrm{var}}$,
$\mathrm{OOD}_{\mathrm{cond}}$, and their combination.
For $\mathrm{OOD}_{\mathrm{var}}$ under $\mathrm{OOD}_{\mathrm{cond}}$, $R^2$
rises from 0.812 to 0.876, 0.127 to 0.205, and 0.135 to 0.225 in the three
domains.
The joint-OOD gains extend the current-state decoding benefit to simultaneous
variable- and condition-level distribution shifts.

Table~\ref{tab:ivideogpt_aggregate} aggregates the matched iVideoGPT~\citep{wu2024ivideogpt}
comparison across domains using the same settings from Time-JEPA. We retain iVideoGPT's video-prediction objective, add an auxiliary head to regress temporally aligned physical states during fine-tuning, and remove the head before video-only probing. Relative to matched Video FT, our method improves both metrics in all four aggregate settings and 10 of 12 domain panels (Appendix~\ref{app:target_matrices}).
These results suggest that the method has some transferability across video world models and can be applied to more backbones.

To gauge the difficulty of current-state decoding under held-out conditions, we
also evaluate two original video backbones, InternVideo-Next-L~\citep{wang2025internvideonext}
and VideoMAE V2-G~\citep{wang2023videomaev2}.
Appendix~\ref{app:target_matrices} gives their results.
In Appendix~\ref{app:alternative_supervision}, we compare Time-JEPA with methods supervised on individual
frames or frame pairs, and separately evaluate the effect of adding explicit
physical-law constraints to Time-JEPA. These comparisons are consistent with the concerns raised in the Introduction: frame- and frame-pair supervision may underuse full-video temporal context, while a specific physical-law constraint can improve its targeted relation without improving broader physical-state estimates. We also report component ablations and sensor sensitivity in
Appendices~\ref{app:ablations} and~\ref{app:sensitivity_analysis}.

\subsection{Multi-Domain Training}
\label{sec:mix_results}

To evaluate Time-JEPA's shared target interface for multi-domain training, we
jointly fine-tune one encoder with physical time-series supervision from all
three domains. We compare multi-domain Time-JEPA with matched single-domain encoders under the same per-domain training schedules in Figure~\ref{fig:multidomain_gain}.
Across all three domains, multi-domain Time-JEPA improves both metrics for
$\mathrm{ID}_{\mathrm{var}}$ under $\mathrm{ID}_{\mathrm{cond}}$.
The gains extend to $\mathrm{OOD}_{\mathrm{var}}$,
$\mathrm{OOD}_{\mathrm{cond}}$, and their combination.
Specifically, for $\mathrm{OOD}_{\mathrm{var}}$ under
$\mathrm{OOD}_{\mathrm{cond}}$, $R^2$ increases by 0.047, 0.063, and 0.231 in
autonomous driving, robot manipulation, and human motion, respectively.
These results show that Time-JEPA can use a shared interface to benefit
from multi-domain physical time-series supervision and further improve
physical-state decoding.
Appendix~\ref{app:mix_details} provides detailed results and
further analysis. The results show the same multi-domain advantage for future-state prediction and kinematic reconstruction, so we use the multi-domain setting for all subsequent tasks.

\subsection{Future-State Prediction}
\label{sec:future_state}

Section~\ref{sec:single_domain_main} evaluates current-state decoding within
the observed clip.
Multi-horizon future-state prediction extends this evaluation beyond the clip
and tests whether the representation captures how physical state evolves.
Let $t$ denote the time of the final observed frame.
A probe on features from each frozen online encoder
predicts physical states at $t+\{0.2,0.4,0.6,0.8,1.0\}$~s.

We use the three configurations from Section~\ref{sec:protocol_tasks} and the
multi-domain training setting from Section~\ref{sec:protocol_probes}.
We train probes with future labels for the state-variable groups
defined in Section~\ref{sec:protocol_domains}.
Table~\ref{tab:future_aggregate} reports nRMSE and $R^2$ across both the
state-variable and condition axes.

\begin{table}[!tb]
    \centering
    \begin{minipage}[t]{0.485\textwidth}
        \begin{minipage}[t][74pt][t]{\linewidth}
            \caption{\raggedright\textbf{iVideoGPT current-state decoding.}
            Setting: variable$*$condition. We average views and variables, then
            weight domains by 30 $\mathrm{ID}_{\mathrm{var}}$ or 16
            $\mathrm{OOD}_{\mathrm{var}}$ dimensions.}
            \label{tab:ivideogpt_aggregate}
        \end{minipage}
        \centering
        \fontsize{8.5}{10}\selectfont
        \setlength{\tabcolsep}{0pt}
        \begin{tabular*}{\linewidth}{@{\extracolsep{\fill}}llccc@{}}
            \toprule
        Setting & Metric & Original & \shortstack{Video\\[-1pt]FT} & Time \\
            \midrule
            \multirow{2}{*}{{\fontsize{8}{9}\selectfont$\mathrm{ID}_{\mathrm{var}}\!\ast\!\mathrm{ID}_{\mathrm{cond}}$}}
            & \makebox[1.5em][l]{$r$}\(\uparrow\) & 0.360 & 0.441 & \textbf{0.557} \\
            & \makebox[1.5em][l]{$R^2$}\(\uparrow\) & 0.136 & 0.208 & \textbf{0.328} \\
            \multirow{2}{*}{{\fontsize{8}{9}\selectfont$\mathrm{OOD}_{\mathrm{var}}\!\ast\!\mathrm{ID}_{\mathrm{cond}}$}}
            & \makebox[1.5em][l]{$r$}\(\uparrow\) & 0.415 & 0.481 & \textbf{0.561} \\
            & \makebox[1.5em][l]{$R^2$}\(\uparrow\) & 0.184 & 0.253 & \textbf{0.341} \\
            \midrule
            \multirow{2}{*}{{\fontsize{8}{9}\selectfont$\mathrm{ID}_{\mathrm{var}}\!\ast\!\mathrm{OOD}_{\mathrm{cond}}$}}
            & \makebox[1.5em][l]{$r$}\(\uparrow\) & 0.157 & 0.165 & \textbf{0.219} \\
            & \makebox[1.5em][l]{$R^2$}\(\uparrow\) & $-0.026$ & 0.007 & \textbf{0.042} \\
            \multirow{2}{*}{{\fontsize{8}{9}\selectfont$\mathrm{OOD}_{\mathrm{var}}\!\ast\!\mathrm{OOD}_{\mathrm{cond}}$}}
            & \makebox[1.5em][l]{$r$}\(\uparrow\) & 0.214 & 0.253 & \textbf{0.313} \\
            & \makebox[1.5em][l]{$R^2$}\(\uparrow\) & 0.014 & 0.070 & \textbf{0.113} \\
            \bottomrule
        \end{tabular*}
    \end{minipage}\hspace{0.02\textwidth}%
    \begin{minipage}[t]{0.485\textwidth}
        \begin{minipage}[t][74pt][t]{\linewidth}
            \caption{\raggedright\textbf{Multi-domain V-JEPA future-state prediction.}
            Setting: variable$*$condition. We average five horizons, then weight
            domains by 30 $\mathrm{ID}_{\mathrm{var}}$ or 16
            $\mathrm{OOD}_{\mathrm{var}}$ dimensions.}
            \label{tab:future_aggregate}
        \end{minipage}
        \centering
        \fontsize{8.5}{10}\selectfont
        \setlength{\tabcolsep}{0pt}
        \begin{tabular*}{\linewidth}{@{\extracolsep{\fill}}llccc@{}}
            \toprule
            Setting & Metric & Original & \shortstack{Video\\[-1pt]FT} & \shortstack{Time-\\[-1pt]JEPA} \\
            \midrule
            \multirow{2}{*}{{\fontsize{8}{9}\selectfont$\mathrm{ID}_{\mathrm{var}}\!\ast\!\mathrm{ID}_{\mathrm{cond}}$}}
            & \makebox[3.5em][l]{nRMSE}\(\downarrow\) & 0.857 & 0.846 & \textbf{0.651} \\
            & \makebox[3.5em][l]{$R^2$}\(\uparrow\) & 0.280 & 0.296 & \textbf{0.550} \\
            \multirow{2}{*}{{\fontsize{8}{9}\selectfont$\mathrm{OOD}_{\mathrm{var}}\!\ast\!\mathrm{ID}_{\mathrm{cond}}$}}
            & \makebox[3.5em][l]{nRMSE}\(\downarrow\) & 0.873 & 0.861 & \textbf{0.732} \\
            & \makebox[3.5em][l]{$R^2$}\(\uparrow\) & 0.254 & 0.272 & \textbf{0.463} \\
            \midrule
            \multirow{2}{*}{{\fontsize{8}{9}\selectfont$\mathrm{ID}_{\mathrm{var}}\!\ast\!\mathrm{OOD}_{\mathrm{cond}}$}}
            & \makebox[3.5em][l]{nRMSE}\(\downarrow\) & 0.953 & 0.949 & \textbf{0.875} \\
            & \makebox[3.5em][l]{$R^2$}\(\uparrow\) & 0.100 & 0.104 & \textbf{0.202} \\
            \multirow{2}{*}{{\fontsize{8}{9}\selectfont$\mathrm{OOD}_{\mathrm{var}}\!\ast\!\mathrm{OOD}_{\mathrm{cond}}$}}
            & \makebox[3.5em][l]{nRMSE}\(\downarrow\) & 0.930 & 0.911 & \textbf{0.844} \\
            & \makebox[3.5em][l]{$R^2$}\(\uparrow\) & 0.117 & 0.153 & \textbf{0.235} \\
            \bottomrule
        \end{tabular*}
    \end{minipage}
\end{table}

Relative to the video-fine-tuned control, Time-JEPA reduces aggregate nRMSE and
increases aggregate $R^2$ for $\mathrm{ID}_{\mathrm{var}}$ under
$\mathrm{ID}_{\mathrm{cond}}$.
These improvements extend to $\mathrm{OOD}_{\mathrm{var}}$,
$\mathrm{OOD}_{\mathrm{cond}}$, and their combination.
Under $\mathrm{OOD}_{\mathrm{cond}}$, nRMSE falls from 0.949 to 0.875 for
$\mathrm{ID}_{\mathrm{var}}$ and from 0.911 to 0.844 for
$\mathrm{OOD}_{\mathrm{var}}$.
These results extend the current-state decoding gains to future-state
prediction, including the joint
$\mathrm{OOD}_{\mathrm{var}}/\mathrm{OOD}_{\mathrm{cond}}$ setting.
These results definitely demonstrate a better understanding of state evolution.
Appendix~\ref{app:future_state} provides further details on these comparisons.

\begin{table}[!b]
    \caption{\textbf{Primary kinematic reconstruction errors.}
    Mean $\pm$ sample standard deviation across five probe seeds.
    Within each seed, trajectory, reach-rate, and distance MAE average three
    horizons, two arms, and four limbs, respectively.
    Video FT and Time-JEPA use multi-domain training.
    Bold marks the lowest error per row.}
    \label{tab:grounded_consistency}
    \centering
    \fontsize{8.5}{10.5}\selectfont
    \setlength{\tabcolsep}{5pt}
    \begin{tabularx}{\textwidth}{@{}cl>{\raggedright\arraybackslash}Xccc@{}}
        \toprule
        Condition & Domain & Metric $\downarrow$
        & Original & Video FT & Time-JEPA \\
        \midrule
        \multirow{3}{*}{\textbf{$\mathrm{ID}_{\mathrm{cond}}$}} & autonomous driving & Trajectory MAE (m)
        & 4.423\(\,\pm\,\)0.115 & 3.932\(\,\pm\,\)0.123
        & \textbf{1.795\(\,\pm\,\)0.027} \\
        & robot manipulation & Reach-rate MAE (cm/s)
        & 5.523\(\,\pm\,\)0.010 & 5.372\(\,\pm\,\)0.008
        & \textbf{2.290\(\,\pm\,\)0.006} \\
        & human motion & Distance MAE (cm)
        & 6.877\(\,\pm\,\)0.093 & 6.815\(\,\pm\,\)0.110
        & \textbf{4.619\(\,\pm\,\)0.065} \\
        \midrule
        \multirow{3}{*}{\textbf{$\mathrm{OOD}_{\mathrm{cond}}$}} & autonomous driving & Trajectory MAE (m)
        & 4.829\(\,\pm\,\)0.199 & 4.150\(\,\pm\,\)0.146
        & \textbf{1.858\(\,\pm\,\)0.144} \\
        & robot manipulation & Reach-rate MAE (cm/s)
        & 5.901\(\,\pm\,\)0.030 & 5.822\(\,\pm\,\)0.035
        & \textbf{4.171\(\,\pm\,\)0.046} \\
        & human motion & Distance MAE (cm)
        & 7.229\(\,\pm\,\)0.177 & 7.120\(\,\pm\,\)0.210
        & \textbf{4.557\(\,\pm\,\)0.053} \\
        \bottomrule
    \end{tabularx}
\end{table}

\section{Kinematic Reconstruction}
\label{sec:grounded_consistency}

Beyond decoding individual physical states, we evaluate whether the video
representation captures the physical relations that connect them.
We use the three configurations from Section~\ref{sec:protocol_tasks} and
the multi-domain training setting from Section~\ref{sec:protocol_probes}.
Kinematic reconstruction applies fixed physical laws to predictions from
separate probes and compares the reconstructed states with ground
truth.
Following Section~\ref{sec:method_evaluation}, we train these probes on features
from the frozen online encoder.
Each probe learns its own parameters to predict one physical state and we use the physical laws only at evaluation, after encoder fine-tuning and
probe training.
Table~\ref{tab:grounded_consistency} reports the primary reconstruction error
for the three configurations under $\mathrm{ID}_{\mathrm{cond}}$ and
$\mathrm{OOD}_{\mathrm{cond}}$ in all three domains.
Each domain uses its own physical law and unit, so we compare the
configurations within each row.

For autonomous driving, integrated trajectory reconstruction follows the
physical laws linking vehicle velocity and yaw rate to changes in position
and heading.
We integrate longitudinal and lateral velocity predictions with yaw-rate
predictions from separate probes to recover trajectory endpoints.
We average endpoint MAE across three horizons.
Relative to the video-fine-tuned control, Time-JEPA lowers trajectory MAE in
both condition settings, with $\mathrm{OOD}_{\mathrm{cond}}$ MAE falling from
4.150 to 1.858~m.

For robot manipulation, reach-rate reconstruction estimates how quickly the
distance between the shoulder and hand changes.
We use joint-angle predictions and the robot's fixed geometry to determine
the shoulder-to-hand direction, then project hand-velocity predictions onto
this direction.
We average signed reach-rate MAE across both arms within each probe seed.
Time-JEPA also lowers reach-rate MAE under both condition settings.
Under $\mathrm{OOD}_{\mathrm{cond}}$, it reduces the error from 5.822~cm/s for
the video-fine-tuned control to 4.171~cm/s.

For human motion, the cosine law connects joint angles and limb segment
lengths to distances between limb endpoints.
We therefore combine elbow and knee angle predictions with segment lengths to recover shoulder-to-wrist and hip-to-ankle distances
through cosine-law distance reconstruction.
We average distance MAE across both arms and both legs.
Time-JEPA achieves lower distance MAE than the video-fine-tuned control in
both condition settings, including a decrease from 7.120 to 4.557~cm under
$\mathrm{OOD}_{\mathrm{cond}}$.

These smaller errors show that Time-JEPA's encoder learns video representations
that support more accurate kinematic reconstructions and capture physical
relations more effectively.
Appendix~\ref{app:grounded_details} provides the equations, ground-truth
validation, and detailed results.

\begin{figure}[!t]
    \centering
    \includegraphics[width=0.90\textwidth]{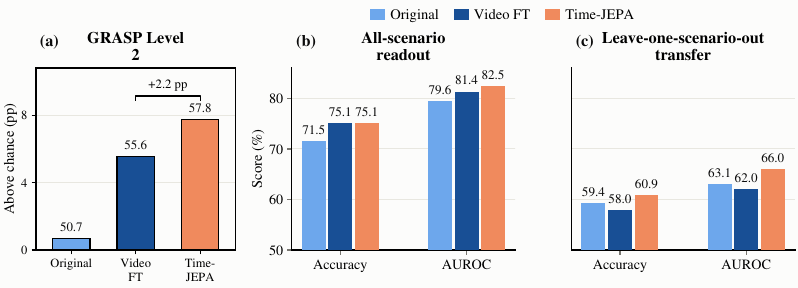}
    \caption{\textbf{External physical evaluations.}
    (a) GRASP Level 2 eight-property macro pair accuracy over 2,048 matched
    possible--impossible pairs; bars show points above 50\% chance and labels
    give absolute accuracy. (b,c) Physion v1.5 scenario-macro Accuracy and
    AUROC (\%) with chronological input, averaged equally over seven scenarios.}
    \label{fig:external_physical}
\end{figure}

\section{External Physical Evaluations}
\label{sec:external_transfer}

To examine transfer to external physical tasks, we evaluate Time-JEPA on
GRASP and Physion.
GRASP tests sensitivity to physical anomalies, and Physion tests
future-contact prediction (Figure~\ref{fig:external_physical}).
These two evaluations complement the decoding, prediction, and
reconstruction tasks and broaden our assessment of Time-JEPA's physical
generalization.

\subsection{GRASP: Violation-of-Expectation Prediction}
\label{sec:grasp}

GRASP Level 2 contains 2,048 matched possible--impossible video pairs from
16 test families spanning eight intuitive-physics properties
\citep{jassim2024grasp}. We use the three configurations from
Section~\ref{sec:protocol_tasks} and the multi-domain training setting from
Section~\ref{sec:protocol_probes}. We freeze the online encoder, target encoder,
and predictor for every configuration.

We compute latent $L_1$ prediction error between predicted and target
latent features, then average the window errors to obtain \emph{AverageSurprise}
for each video. For each pair, we classify the video with higher
\emph{AverageSurprise} as impossible. Our primary metric averages pair accuracy
equally across the eight properties. Figure~\ref{fig:external_physical} reports the
three external-physical evaluation panels.
Time-JEPA reaches $57.8\%$, exceeding the matched video-fine-tuned control by
$2.2$ percentage points. This gain suggests that time-series supervision
improves sensitivity to physical anomalies.
Appendix~\ref{app:external_transfer} provides the full protocol and
property-level results.

\subsection{Physion: Future Contact}
\label{sec:physion}

On Physion~\citep{bear2021physion}, we use the same configurations to predict
whether two designated objects will make contact after four observed frames.
We provide the frames in chronological order and train the probe on frozen target-encoder features.
The \emph{all-scenario readout} includes all seven scenario types in probe
training, while \emph{leave-one-scenario-out transfer} trains each probe on six
types and tests it on the remaining type.

Figure~\ref{fig:external_physical} reports Accuracy and AUROC for both
Physion readout protocols.
AUROC (area under the receiver operating characteristic curve) measures how
often the probe assigns a higher contact probability to a true contact case
than to a true non-contact case.
Although Time-JEPA achieves nearly identical Accuracy to the video-fine-tuned
control in the all-scenario readout, it increases AUROC by $1.182$ percentage
points. The gains extend to both metrics in leave-one-scenario-out transfer,
where Time-JEPA improves Accuracy and AUROC by $2.945$ and $3.997$ percentage
points over the same control. These results suggest that time-series
supervision supports a better understanding of physical interactions,
including in unseen scenarios.
Appendix~\ref{app:external_transfer} provides further results and protocol details.
\section{Conclusion}
\label{sec:conclusion}

We introduced Time-JEPA, which shapes video representations using low-dimensional, temporally aligned physical time series as auxiliary supervision during fine-tuning. Its shared interface supports multi-domain training, and video remains the encoder's only input. In multiple domains, Time-JEPA improves current-state decoding, multi-horizon future-state prediction, and kinematic reconstruction over matched video-fine-tuned controls across both condition and state-variable distribution shifts. Multi-domain training provides aggregate gains over matched single-domain training in all three tasks. External evaluations further suggest that time-series supervision improves sensitivity to physical anomalies and the representation of physical interactions, including in unseen scenarios. Aggregate gains on a second video backbone indicate that the method can be applied to more backbones. These findings suggest that time series help video world models better understand how events unfold. Future work can extend this interface to richer physical time series and action-conditioned settings.

\clearpage
\subsection*{AI Use Statement}
We used generative AI tools to assist with writing code and polishing the
wording of this manuscript. The authors reviewed the AI-assisted code and text
and take responsibility for the final content of this work.

\subsection*{Ethics Statement}
We use previously released datasets, including human activity videos and
associated pose annotations. Such recordings can contain identifiable people;
reuse of the source data and any derived materials should respect the datasets'
access terms and participant privacy. Our study reports aggregate results from
offline video evaluations and does not test real-world vehicle or robot control.
The reported scores do not establish safety for deployment in those settings.

\subsection*{Reproducibility Statement}
Appendix~\ref{app:reproducibility} documents the fine-tuning and probing setup
and checkpoint selection. Appendices~\ref{app:target_matrices}--\ref{app:alternative_supervision} provide
the variable-level results, common time-series preprocessing, dataset definitions and
splits, kinematic-reconstruction formulas, multi-domain training details, future and
downstream protocols, additional-backbone comparisons, and ablations
needed to reproduce the reported comparisons.

\bibliography{references}
\bibliographystyle{time_jepa}

\clearpage
\appendix
\renewcommand{\thetable}{\Alph{section}\arabic{table}}
\renewcommand{\thefigure}{\Alph{section}\arabic{figure}}
\renewcommand{\theHtable}{\Alph{section}.\arabic{table}}
\renewcommand{\theHfigure}{\Alph{section}.\arabic{figure}}

\renewcommand{\topfraction}{0.95}
\renewcommand{\bottomfraction}{0.95}
\renewcommand{\textfraction}{0.05}
\renewcommand{\floatpagefraction}{0.8}
\setcounter{topnumber}{5}
\setcounter{bottomnumber}{5}
\setcounter{totalnumber}{8}
\setlength{\textfloatsep}{12pt plus 2pt minus 2pt}
\setlength{\floatsep}{10pt plus 2pt minus 2pt}
\setlength{\intextsep}{10pt plus 2pt minus 2pt}
\makeatletter
\setlength{\@fptop}{0pt}
\setlength{\@fpsep}{12pt}
\setlength{\@fpbot}{0pt plus 1fil}
\makeatother

\section{Implementation and Reproducibility}
\label{app:reproducibility}
\setcounter{table}{0}

\subsection{Fine-Tuning Protocol}

All main V-JEPA comparisons start from the same pretrained model.
The video-fine-tuned control and Time-JEPA share the configuration and sampling
distributions, with independent draws of clips, crops, and JEPA masks.
Time-JEPA adds the auxiliary head and time-series loss.

\begin{table}[htbp]
    \caption{\textbf{Matched V-JEPA fine-tuning recipe.}
    Time-JEPA adds the time-series supervision branch and its loss to the shared settings.
    Domain steps count batches from the corresponding domain.}
    \label{tab:fine_tuning_recipe}
    \centering
    \small
    \setlength{\tabcolsep}{0pt}
    \renewcommand{\arraystretch}{1.0}
    \begin{tabularx}{\textwidth}{@{}X@{}}
        \appendixcard{Video input}{64 frames at 5 FPS, $384\!\times\!384$ crop, patch size 16, tubelet size 2, 32 temporal tokens, and strict temporal sampling.} \\
        \noalign{\vskip 3pt}
        \appendixcard{Sampling}{All domains use the same 64-frame, 5-FPS temporal context (12.6 s between the first and last sampled frames). Valid start times are constructed from each source timeline: a random offset within each selected autonomous-driving video, or a timestamp-aligned grid with 12.8 s spacing for robot manipulation and human motion.} \\
        \noalign{\vskip 3pt}
        \appendixcard{Augmentation}{Random resized crop with scale $[0.3,1.0]$ and aspect ratio $[0.75,1.35]$. We retain the original horizontal orientation.} \\
        \noalign{\vskip 3pt}
        \appendixcard{Optimization}{20 epochs, batch size 8 per GPU, AdamW, learning rate $3\!\times\!10^{-6}\rightarrow3\!\times\!10^{-5}\rightarrow10^{-6}$ with one warmup epoch and cosine decay, weight decay 0.01, bfloat16, and EMA momentum $0.9995\rightarrow0.9999$.} \\
        \noalign{\vskip 3pt}
        \appendixcard{JEPA masks}{Eight block masks with spatial scale 0.15 and two with spatial scale 0.70. All use temporal scale 1.0.} \\
        \noalign{\vskip 3pt}
        \appendixcard{Time-series supervision}{A separate unmasked online-encoder pass supplies the final-layer patch grid. A two-layer pointwise nonlinear auxiliary head with GELU regresses aligned states under a masked Smooth-L1 loss ($\beta=0.5$). The loss weight warms up linearly over the first 250 domain steps.} \\
        \noalign{\vskip 3pt}
        \appendixcard{Time-series loss weight}{$\lambda=0.2$ for autonomous driving and $\lambda=0.1$ for robot manipulation and human motion. Multi-domain training retains these domain-specific values.} \\
        \noalign{\vskip 3pt}
        \appendixcard{Random seed}{239 for every main fine-tuning run.}
    \end{tabularx}
\end{table}

\subsection{Probe Training on Frozen Encoders and Aggregation}

For each encoder and current-state task, we freeze the encoder and train a
new pointwise nonlinear probe for 15 epochs. The probe uses AdamW with a
learning rate of $10^{-4}$, cosine decay to $10^{-6}$, one warmup epoch, and
weight decay 0.01. Batch size is 16 for autonomous driving and 8 for robot
manipulation and human motion. We select each probe by validation loss, then
evaluate it under $\mathrm{ID}_{\mathrm{cond}}$ and
$\mathrm{OOD}_{\mathrm{cond}}$. Both tests reuse the probe-training
normalization statistics and fitted probe parameters.

We compute each metric per state variable and view, average views equally, then
average variables equally. We average these eight view--variable entries equally. Kinematic-reconstruction
probes use the same optimizer schedule and domain-specific batch sizes.

\FloatBarrier
\section{Full Current-State Results}
\label{app:target_matrices}
\setcounter{table}{0}

Table~\ref{tab:app_ivideogpt_current_state} gives the iVideoGPT~\citep{wu2024ivideogpt}
domain scores behind Table~\ref{tab:ivideogpt_aggregate}.
Time-JEPA improves Pearson $r$ and $R^2$ over Video FT in 10 of 12 domain panels.
The two exceptions occur under the held-out fisheye views in robot manipulation.
Table~\ref{tab:app_additional_backbones} uses the original
InternVideo-Next-L~\citep{wang2025internvideonext} and
VideoMAE V2-G~\citep{wang2023videomaev2} encoders to gauge the difficulty of
held-out current-state decoding, alongside the Time-JEPA encoder for comparison.
Both original backbones score lower under the robot manipulation fisheye views
than under the seen views across both variable groups and metrics.
Their lower held-out-condition scores support the difficulty of our task
setting.

Tables~\ref{tab:app_agibot_readout_targets}--\ref{tab:app_egoexo_shot_targets}
expand Table~\ref{tab:single_domain_main} to variable-level V-JEPA results and
include nRMSE.
$\mathrm{ID}_{\mathrm{var}}$ denotes state variables included in auxiliary
time-series supervision, and $\mathrm{OOD}_{\mathrm{var}}$ denotes those
excluded from it.
Each result comes from one supervised probe on a frozen encoder.
Bold marks the best unrounded value for each variable, condition, and metric.
Time-JEPA improves all three variable-macro metrics in all 12 V-JEPA panels.

\begin{table}[!htbp]
    \caption{\textbf{Current-state decoding and transfer with iVideoGPT.}
    The three configurations use one matched probe run at native resolution and
    Table~\ref{tab:single_domain_main}'s view/variable aggregation. Bold marks the best unrounded value.}
    \label{tab:app_ivideogpt_current_state}
    \centering
    \fontsize{8.5}{10.5}\selectfont
    \setlength{\tabcolsep}{1.5pt}
    \renewcommand{\arraystretch}{1.06}
    \begin{tabular*}{\textwidth}{@{\extracolsep{\fill}}cccccc@{\hspace{8pt}}ccc@{\hspace{8pt}}ccc@{}}
        \toprule
         &  &  & \multicolumn{3}{c}{\textbf{autonomous driving}} & \multicolumn{3}{c}{\textbf{robot manipulation}} & \multicolumn{3}{c}{\textbf{human motion}} \\
        \cmidrule(lr){4-6}\cmidrule(lr){7-9}\cmidrule(lr){10-12}
        Condition & Variables & Metric & Original & \shortstack{Video\\[-1pt]FT} & Time & Original & \shortstack{Video\\[-1pt]FT} & Time & Original & \shortstack{Video\\[-1pt]FT} & Time \\
        \midrule
        \multirow{4}{*}{\textbf{$\mathrm{ID}_{\mathrm{cond}}$}} & \multirow{2}{*}{$\mathrm{ID}_{\mathrm{var}}$} & \appmetric{r} & 0.466 & 0.563 & \textbf{0.757} & 0.362 & 0.457 & \textbf{0.570} & 0.320 & 0.370 & \textbf{0.462} \\
        & & \appmetric{R^2} & 0.260 & 0.403 & \textbf{0.654} & 0.145 & 0.223 & \textbf{0.342} & 0.076 & 0.115 & \textbf{0.193} \\
        & \multirow{2}{*}{$\mathrm{OOD}_{\mathrm{var}}$} & \appmetric{r} & 0.576 & 0.730 & \textbf{0.857} & 0.425 & 0.495 & \textbf{0.564} & 0.303 & 0.311 & \textbf{0.379} \\
        & & \appmetric{R^2} & 0.322 & 0.526 & \textbf{0.736} & 0.199 & 0.265 & \textbf{0.339} & 0.078 & 0.068 & \textbf{0.108} \\
        \midrule
        \multirow{4}{*}{\textbf{$\mathrm{OOD}_{\mathrm{cond}}$}} & \multirow{2}{*}{$\mathrm{ID}_{\mathrm{var}}$} & \appmetric{r} & 0.441 & 0.552 & \textbf{0.727} & \textbf{0.077} & 0.040 & 0.034 & 0.223 & 0.286 & \textbf{0.419} \\
        & & \appmetric{R^2} & 0.225 & 0.380 & \textbf{0.608} & \textbf{\(-\)0.065} & \(-\)0.068 & \(-\)0.103 & \(-\)0.031 & 0.032 & \textbf{0.144} \\
        & \multirow{2}{*}{$\mathrm{OOD}_{\mathrm{var}}$} & \appmetric{r} & 0.525 & 0.728 & \textbf{0.858} & \textbf{0.109} & 0.097 & 0.090 & 0.195 & 0.218 & \textbf{0.344} \\
        & & \appmetric{R^2} & 0.269 & 0.522 & \textbf{0.735} & \(-\)0.060 & \textbf{\(-\)0.057} & \(-\)0.096 & \(-\)0.019 & 0.002 & \textbf{0.075} \\
        \bottomrule
    \end{tabular*}
\end{table}

\begin{table}[!htbp]
    \caption{\textbf{Current-state task difficulty and Time-JEPA comparison.}
    We evaluate original pretrained InternVideo-Next-L and VideoMAE V2-G encoders
    alongside the Time-JEPA encoder at their native input resolutions, using
    Table~\ref{tab:single_domain_main}'s splits, supervised-probe protocol, and aggregation.
    The column labels ``I-Next-L,'' ``VMAE V2-G,'' and ``Time-JEPA'' denote
    InternVideo-Next-L, VideoMAE V2-G, and Time-JEPA, respectively.
    Variable and condition notation follows Section~\ref{sec:protocol_domains}.
    Bold marks the best unrounded value among the listed encoders.}
    \label{tab:app_additional_backbones}
    \centering
    \fontsize{8.5}{10.5}\selectfont
    \setlength{\tabcolsep}{0pt}
    \renewcommand{\arraystretch}{1.06}
    \begin{tabular*}{\textwidth}{@{\extracolsep{\fill}}ccc@{\hspace{-1.5pt}}ccc@{\hspace{-1.5pt}}ccc@{\hspace{-1.5pt}}ccc@{}}
        \toprule
         &  &  & \multicolumn{3}{c}{\textbf{autonomous driving}} & \multicolumn{3}{c}{\textbf{robot manipulation}} & \multicolumn{3}{c}{\textbf{human motion}} \\
        \cmidrule(lr){4-6}\cmidrule(lr){7-9}\cmidrule(lr){10-12}
        Condition & Variables & Metric & \shortstack{I-Next-\\L} & \shortstack{VMAE\\V2-G} & \shortstack{Time-\\JEPA} & \shortstack{I-Next-\\L} & \shortstack{VMAE\\V2-G} & \shortstack{Time-\\JEPA} & \shortstack{I-Next-\\L} & \shortstack{VMAE\\V2-G} & \shortstack{Time-\\JEPA} \\
        \midrule
        \multirow{4}{*}{\textbf{$\mathrm{ID}_{\mathrm{cond}}$}} & \multirow{2}{*}{$\mathrm{ID}_{\mathrm{var}}$} & \appmetric{r} & 0.662 & 0.613 & \textbf{0.916} & 0.557 & 0.324 & \textbf{0.845} & 0.536 & 0.354 & \textbf{0.589} \\
        & & \appmetric{R^2} & 0.475 & 0.429 & \textbf{0.849} & 0.345 & 0.128 & \textbf{0.725} & 0.279 & 0.113 & \textbf{0.345} \\
        & \multirow{2}{*}{$\mathrm{OOD}_{\mathrm{var}}$} & \appmetric{r} & 0.510 & 0.652 & \textbf{0.920} & 0.608 & 0.407 & \textbf{0.743} & 0.485 & 0.327 & \textbf{0.504} \\
        & & \appmetric{R^2} & 0.260 & 0.417 & \textbf{0.846} & 0.377 & 0.172 & \textbf{0.572} & 0.214 & 0.096 & \textbf{0.222} \\
        \midrule
        \multirow{4}{*}{\textbf{$\mathrm{OOD}_{\mathrm{cond}}$}} & \multirow{2}{*}{$\mathrm{ID}_{\mathrm{var}}$} & \appmetric{r} & 0.637 & 0.573 & \textbf{0.902} & 0.240 & 0.140 & \textbf{0.492} & 0.539 & 0.323 & \textbf{0.581} \\
        & & \appmetric{R^2} & 0.434 & 0.358 & \textbf{0.824} & \(-\)0.089 & \(-\)0.008 & \textbf{0.085} & 0.283 & 0.073 & \textbf{0.332} \\
        & \multirow{2}{*}{$\mathrm{OOD}_{\mathrm{var}}$} & \appmetric{r} & 0.432 & 0.602 & \textbf{0.937} & 0.369 & 0.251 & \textbf{0.555} & 0.495 & 0.277 & \textbf{0.515} \\
        & & \appmetric{R^2} & 0.187 & 0.361 & \textbf{0.876} & 0.046 & 0.055 & \textbf{0.205} & 0.233 & 0.048 & \textbf{0.225} \\
        \bottomrule
    \end{tabular*}
\end{table}

\begin{table}[!htbp]
    \caption{\textbf{robot manipulation: $\mathrm{ID}_{\mathrm{var}}$ variables.}}
    \label{tab:app_agibot_readout_targets}
    \centering
    \fontsize{8.5}{10.5}\selectfont
    \setlength{\tabcolsep}{1pt}
    \renewcommand{\arraystretch}{1.28}
    \begin{tabular*}{\textwidth}{@{\extracolsep{\fill}}c>{\centering\arraybackslash}p{0.23\textwidth}ccc@{\hspace{5pt}}ccc@{\hspace{5pt}}ccc@{}}
        \toprule
         & & \multicolumn{3}{c}{Original} & \multicolumn{3}{c}{Video FT} & \multicolumn{3}{c}{Time-JEPA} \\
        \cmidrule(lr){3-5}\cmidrule(lr){6-8}\cmidrule(lr){9-11}
        Condition & State variable & $r\uparrow$ & $R^2\uparrow$ & nRMSE $\downarrow$ & $r\uparrow$ & $R^2\uparrow$ & nRMSE $\downarrow$ & $r\uparrow$ & $R^2\uparrow$ & nRMSE $\downarrow$ \\
        \midrule
        \multirow{19}{*}{\textbf{$\mathrm{ID}_{\mathrm{cond}}$}} & Base translation speed & 0.459 & 0.210 & 0.907 & 0.461 & 0.212 & 0.906 & \textbf{0.529} & \textbf{0.279} & \textbf{0.867} \\
        & Base angular speed & 0.465 & 0.216 & 0.793 & 0.474 & 0.224 & 0.789 & \textbf{0.596} & \textbf{0.352} & \textbf{0.721} \\
        & Waist PC1 & 0.763 & 0.582 & 0.647 & 0.782 & 0.611 & 0.624 & \textbf{0.912} & \textbf{0.833} & \textbf{0.407} \\
        & Waist PC2 & 0.803 & 0.644 & 0.608 & 0.813 & 0.661 & 0.593 & \textbf{0.899} & \textbf{0.807} & \textbf{0.449} \\
        & Left arm position $x$ & 0.608 & 0.371 & 0.768 & 0.637 & 0.406 & 0.746 & \textbf{0.847} & \textbf{0.718} & \textbf{0.510} \\
        & Left arm position $y$ & 0.800 & 0.640 & 0.567 & 0.814 & 0.663 & 0.548 & \textbf{0.929} & \textbf{0.863} & \textbf{0.349} \\
        & Left arm position $z$ & 0.639 & 0.410 & 0.701 & 0.663 & 0.441 & 0.682 & \textbf{0.846} & \textbf{0.718} & \textbf{0.476} \\
        & Right arm position $x$ & 0.709 & 0.502 & 0.710 & 0.735 & 0.539 & 0.684 & \textbf{0.889} & \textbf{0.791} & \textbf{0.456} \\
        & Right arm position $y$ & 0.743 & 0.553 & 0.642 & 0.768 & 0.589 & 0.616 & \textbf{0.920} & \textbf{0.845} & \textbf{0.376} \\
        & Right arm position $z$ & 0.665 & 0.445 & 0.714 & 0.688 & 0.476 & 0.693 & \textbf{0.860} & \textbf{0.742} & \textbf{0.475} \\
        & Left end velocity $x$ & 0.426 & 0.186 & 0.913 & 0.486 & 0.241 & 0.882 & \textbf{0.854} & \textbf{0.728} & \textbf{0.528} \\
        & Left end velocity $y$ & 0.455 & 0.207 & 0.873 & 0.527 & 0.277 & 0.834 & \textbf{0.880} & \textbf{0.772} & \textbf{0.468} \\
        & Left end velocity $z$ & 0.370 & 0.133 & 0.925 & 0.419 & 0.171 & 0.904 & \textbf{0.826} & \textbf{0.680} & \textbf{0.557} \\
        & Right end velocity $x$ & 0.596 & 0.360 & 0.781 & 0.647 & 0.420 & 0.744 & \textbf{0.910} & \textbf{0.828} & \textbf{0.405} \\
        & Right end velocity $y$ & 0.563 & 0.324 & 0.809 & 0.624 & 0.392 & 0.769 & \textbf{0.906} & \textbf{0.821} & \textbf{0.417} \\
        & Right end velocity $z$ & 0.530 & 0.275 & 0.836 & 0.567 & 0.318 & 0.811 & \textbf{0.868} & \textbf{0.754} & \textbf{0.483} \\
        & Left gripper state & 0.718 & 0.519 & 0.693 & 0.731 & 0.537 & 0.679 & \textbf{0.878} & \textbf{0.771} & \textbf{0.480} \\
        & Right gripper state & 0.651 & 0.432 & 0.732 & 0.670 & 0.458 & 0.714 & \textbf{0.869} & \textbf{0.754} & \textbf{0.485} \\
        \midrule
        & \textbf{Variable macro} & 0.609 & 0.389 & 0.757 & 0.639 & 0.424 & 0.734 & \textbf{0.845} & \textbf{0.725} & \textbf{0.495} \\
        \midrule
        \multirow{19}{*}{\textbf{$\mathrm{OOD}_{\mathrm{cond}}$}} & Base translation speed & 0.211 & \textbf{0.007} & \textbf{1.017} & 0.155 & \(-\)0.016 & 1.029 & \textbf{0.355} & 0.006 & 1.017 \\
        & Base angular speed & 0.265 & \textbf{\(-\)0.033} & \textbf{0.911} & 0.207 & \(-\)0.072 & 0.928 & \textbf{0.347} & \(-\)0.048 & 0.917 \\
        & Waist PC1 & 0.511 & \textbf{0.145} & \textbf{0.927} & 0.506 & 0.081 & 0.960 & \textbf{0.540} & 0.073 & 0.968 \\
        & Waist PC2 & 0.449 & \textbf{0.153} & \textbf{0.940} & 0.412 & 0.137 & 0.949 & \textbf{0.617} & \(-\)0.112 & 1.076 \\
        & Left arm position $x$ & 0.319 & \textbf{\(-\)0.084} & \textbf{1.009} & 0.324 & \(-\)0.109 & 1.021 & \textbf{0.420} & \(-\)0.292 & 1.087 \\
        & Left arm position $y$ & 0.404 & 0.131 & 0.880 & 0.409 & 0.137 & 0.877 & \textbf{0.754} & \textbf{0.303} & \textbf{0.786} \\
        & Left arm position $z$ & 0.163 & \textbf{\(-\)0.004} & \textbf{0.916} & 0.144 & \(-\)0.047 & 0.935 & \textbf{0.244} & \(-\)0.063 & 0.943 \\
        & Right arm position $x$ & 0.256 & \(-\)0.074 & 1.044 & 0.264 & \textbf{0.010} & \textbf{1.002} & \textbf{0.463} & \(-\)0.031 & 1.023 \\
        & Right arm position $y$ & 0.259 & \(-\)0.012 & 0.969 & 0.274 & 0.029 & 0.949 & \textbf{0.638} & \textbf{0.321} & \textbf{0.791} \\
        & Right arm position $z$ & 0.207 & 0.017 & 0.952 & 0.243 & \textbf{0.050} & \textbf{0.935} & \textbf{0.245} & \(-\)0.256 & 1.061 \\
        & Left end velocity $x$ & 0.091 & \(-\)0.026 & 1.028 & 0.107 & \(-\)0.030 & 1.030 & \textbf{0.525} & \textbf{0.330} & \textbf{0.809} \\
        & Left end velocity $y$ & 0.157 & 0.006 & 0.980 & 0.197 & \(-\)0.007 & 0.986 & \textbf{0.587} & \textbf{0.227} & \textbf{0.859} \\
        & Left end velocity $z$ & 0.062 & \(-\)0.017 & 1.002 & 0.084 & \(-\)0.015 & 1.001 & \textbf{0.355} & \textbf{0.133} & \textbf{0.920} \\
        & Right end velocity $x$ & 0.107 & \(-\)0.077 & 1.017 & 0.126 & \(-\)0.126 & 1.040 & \textbf{0.530} & \textbf{0.332} & \textbf{0.781} \\
        & Right end velocity $y$ & 0.213 & 0.032 & 0.974 & 0.258 & 0.058 & 0.961 & \textbf{0.596} & \textbf{0.394} & \textbf{0.750} \\
        & Right end velocity $z$ & 0.102 & \(-\)0.028 & 0.996 & 0.109 & \(-\)0.041 & 1.003 & \textbf{0.462} & \textbf{0.175} & \textbf{0.886} \\
        & Left gripper state & 0.275 & 0.055 & 0.980 & 0.343 & 0.070 & 0.972 & \textbf{0.604} & \textbf{0.254} & \textbf{0.862} \\
        & Right gripper state & 0.199 & \(-\)0.151 & 1.056 & 0.220 & \textbf{\(-\)0.125} & \textbf{1.044} & \textbf{0.577} & \(-\)0.209 & 1.074 \\
        \midrule
        & \textbf{Variable macro} & 0.236 & 0.002 & 0.978 & 0.243 & \(-\)0.001 & 0.979 & \textbf{0.492} & \textbf{0.085} & \textbf{0.923} \\
        \bottomrule
    \end{tabular*}
\end{table}

\begin{table}[!htbp]
    \caption{\textbf{robot manipulation: $\mathrm{OOD}_{\mathrm{var}}$ variables.}}
    \label{tab:app_agibot_shot_targets}
    \centering
    \fontsize{8.5}{10.5}\selectfont
    \setlength{\tabcolsep}{1pt}
    \renewcommand{\arraystretch}{1.06}
    \begin{tabular*}{\textwidth}{@{\extracolsep{\fill}}c>{\centering\arraybackslash}p{0.23\textwidth}ccc@{\hspace{5pt}}ccc@{\hspace{5pt}}ccc@{}}
        \toprule
         & & \multicolumn{3}{c}{Original} & \multicolumn{3}{c}{Video FT} & \multicolumn{3}{c}{Time-JEPA} \\
        \cmidrule(lr){3-5}\cmidrule(lr){6-8}\cmidrule(lr){9-11}
        Condition & State variable & $r\uparrow$ & $R^2\uparrow$ & nRMSE $\downarrow$ & $r\uparrow$ & $R^2\uparrow$ & nRMSE $\downarrow$ & $r\uparrow$ & $R^2\uparrow$ & nRMSE $\downarrow$ \\
        \midrule
        \multirow{9}{*}{\textbf{$\mathrm{ID}_{\mathrm{cond}}$}} & Left joint-position norm & 0.764 & 0.583 & 0.616 & 0.775 & 0.600 & 0.603 & \textbf{0.856} & \textbf{0.732} & \textbf{0.494} \\
        & Right joint-position norm & 0.715 & 0.511 & 0.685 & 0.736 & 0.541 & 0.664 & \textbf{0.850} & \textbf{0.722} & \textbf{0.517} \\
        & Left joint-velocity norm & 0.723 & 0.521 & 0.705 & 0.756 & 0.569 & 0.669 & \textbf{0.907} & \textbf{0.821} & \textbf{0.432} \\
        & Right joint-velocity norm & 0.783 & 0.613 & 0.610 & 0.806 & 0.649 & 0.581 & \textbf{0.920} & \textbf{0.845} & \textbf{0.386} \\
        & Left joint-effort norm & 0.533 & 0.283 & 0.853 & 0.544 & 0.294 & 0.846 & \textbf{0.649} & \textbf{0.420} & \textbf{0.766} \\
        & Right joint-effort norm & 0.518 & 0.268 & 0.923 & 0.534 & 0.284 & 0.913 & \textbf{0.650} & \textbf{0.422} & \textbf{0.820} \\
        & Left wrench-force norm & 0.515 & 0.255 & 0.913 & 0.518 & 0.254 & 0.913 & \textbf{0.553} & \textbf{0.293} & \textbf{0.889} \\
        & Right wrench-force norm & 0.505 & 0.252 & 1.043 & 0.509 & 0.256 & 1.040 & \textbf{0.545} & \textbf{0.291} & \textbf{1.015} \\
        \midrule
        & \textbf{Variable macro} & 0.632 & 0.411 & 0.793 & 0.647 & 0.431 & 0.778 & \textbf{0.741} & \textbf{0.568} & \textbf{0.665} \\
        \midrule
        \multirow{9}{*}{\textbf{$\mathrm{OOD}_{\mathrm{cond}}$}} & Left joint-position norm & 0.458 & 0.048 & 0.931 & 0.515 & \textbf{0.087} & \textbf{0.912} & \textbf{0.545} & \(-\)0.036 & 0.971 \\
        & Right joint-position norm & 0.375 & \(-\)0.048 & 1.004 & 0.442 & 0.108 & 0.927 & \textbf{0.668} & \textbf{0.365} & \textbf{0.782} \\
        & Left joint-velocity norm & 0.407 & 0.146 & 0.944 & 0.471 & 0.208 & 0.909 & \textbf{0.886} & \textbf{0.782} & \textbf{0.477} \\
        & Right joint-velocity norm & 0.591 & 0.336 & 0.801 & 0.605 & 0.356 & 0.789 & \textbf{0.858} & \textbf{0.666} & \textbf{0.568} \\
        & Left joint-effort norm & 0.284 & 0.031 & 0.991 & 0.321 & \textbf{0.052} & \textbf{0.980} & \textbf{0.392} & \(-\)0.151 & 1.080 \\
        & Right joint-effort norm & 0.263 & 0.060 & 1.046 & 0.289 & \textbf{0.082} & \textbf{1.034} & \textbf{0.439} & 0.055 & 1.049 \\
        & Left wrench-force norm & 0.327 & \textbf{0.103} & \textbf{1.002} & 0.307 & 0.078 & 1.015 & \textbf{0.396} & 0.064 & 1.023 \\
        & Right wrench-force norm & \textbf{0.280} & \textbf{0.067} & \textbf{1.165} & 0.262 & 0.050 & 1.175 & 0.254 & \(-\)0.106 & 1.268 \\
        \midrule
        & \textbf{Variable macro} & 0.373 & 0.093 & 0.985 & 0.402 & 0.127 & 0.968 & \textbf{0.555} & \textbf{0.205} & \textbf{0.902} \\
        \bottomrule
    \end{tabular*}
\end{table}

\begin{table}[!htbp]
    \normalsize
    \caption{\textbf{autonomous driving: $\mathrm{ID}_{\mathrm{var}}$ variables.}}
    \label{tab:app_covla_readout_targets}
    \centering
    \fontsize{8.5}{10.5}\selectfont
    \setlength{\tabcolsep}{1pt}
    \renewcommand{\arraystretch}{1.06}
    \begin{tabular*}{\textwidth}{@{\extracolsep{\fill}}c>{\centering\arraybackslash}p{0.23\textwidth}ccc@{\hspace{5pt}}ccc@{\hspace{5pt}}ccc@{}}
        \toprule
         & & \multicolumn{3}{c}{Original} & \multicolumn{3}{c}{Video FT} & \multicolumn{3}{c}{Time-JEPA} \\
        \cmidrule(lr){3-5}\cmidrule(lr){6-8}\cmidrule(lr){9-11}
        Condition & State variable & $r\uparrow$ & $R^2\uparrow$ & nRMSE $\downarrow$ & $r\uparrow$ & $R^2\uparrow$ & nRMSE $\downarrow$ & $r\uparrow$ & $R^2\uparrow$ & nRMSE $\downarrow$ \\
        \midrule
        \multirow{4}{*}{\textbf{$\mathrm{ID}_{\mathrm{cond}}$}} & Longitudinal velocity & 0.972 & 0.942 & 0.213 & 0.975 & 0.949 & 0.200 & \textbf{0.990} & \textbf{0.979} & \textbf{0.127} \\
        & Longitudinal acceleration & 0.477 & 0.225 & 0.886 & 0.510 & 0.256 & 0.868 & \textbf{0.771} & \textbf{0.594} & \textbf{0.642} \\
        & Yaw rate & 0.963 & 0.926 & 0.280 & 0.967 & 0.934 & 0.266 & \textbf{0.986} & \textbf{0.973} & \textbf{0.170} \\
        \midrule
        & \textbf{Variable macro} & 0.804 & 0.698 & 0.460 & 0.817 & 0.713 & 0.445 & \textbf{0.916} & \textbf{0.849} & \textbf{0.313} \\
        \midrule
        \multirow{4}{*}{\textbf{$\mathrm{OOD}_{\mathrm{cond}}$}} & Longitudinal velocity & 0.961 & 0.919 & 0.261 & 0.966 & 0.926 & 0.249 & \textbf{0.986} & \textbf{0.972} & \textbf{0.152} \\
        & Longitudinal acceleration & 0.488 & 0.236 & 0.848 & 0.515 & 0.262 & 0.833 & \textbf{0.740} & \textbf{0.542} & \textbf{0.656} \\
        & Yaw rate & 0.950 & 0.900 & 0.351 & 0.954 & 0.908 & 0.337 & \textbf{0.980} & \textbf{0.958} & \textbf{0.228} \\
        \midrule
        & \textbf{Variable macro} & 0.800 & 0.685 & 0.487 & 0.812 & 0.699 & 0.473 & \textbf{0.902} & \textbf{0.824} & \textbf{0.346} \\
        \bottomrule
    \end{tabular*}
\end{table}

\begin{table}[!htbp]
    \normalsize
    \caption{\textbf{autonomous driving: $\mathrm{OOD}_{\mathrm{var}}$ variables.}}
    \label{tab:app_covla_shot_targets}
    \centering
    \fontsize{8.5}{10.5}\selectfont
    \setlength{\tabcolsep}{1pt}
    \renewcommand{\arraystretch}{1.06}
    \begin{tabular*}{\textwidth}{@{\extracolsep{\fill}}c>{\centering\arraybackslash}p{0.23\textwidth}ccc@{\hspace{5pt}}ccc@{\hspace{5pt}}ccc@{}}
        \toprule
         & & \multicolumn{3}{c}{Original} & \multicolumn{3}{c}{Video FT} & \multicolumn{3}{c}{Time-JEPA} \\
        \cmidrule(lr){3-5}\cmidrule(lr){6-8}\cmidrule(lr){9-11}
        Condition & State variable & $r\uparrow$ & $R^2\uparrow$ & nRMSE $\downarrow$ & $r\uparrow$ & $R^2\uparrow$ & nRMSE $\downarrow$ & $r\uparrow$ & $R^2\uparrow$ & nRMSE $\downarrow$ \\
        \midrule
        \multirow{4}{*}{\textbf{$\mathrm{ID}_{\mathrm{cond}}$}} & Steering angle & 0.848 & 0.717 & 0.586 & 0.853 & 0.725 & 0.578 & \textbf{0.884} & \textbf{0.781} & \textbf{0.516} \\
        & Lateral velocity & 0.927 & 0.859 & 0.390 & 0.932 & 0.867 & 0.378 & \textbf{0.951} & \textbf{0.904} & \textbf{0.320} \\
        & Lateral acceleration & 0.861 & 0.736 & 0.479 & 0.872 & 0.756 & 0.460 & \textbf{0.925} & \textbf{0.853} & \textbf{0.357} \\
        \midrule
        & \textbf{Variable macro} & 0.879 & 0.771 & 0.485 & 0.885 & 0.783 & 0.472 & \textbf{0.920} & \textbf{0.846} & \textbf{0.398} \\
        \midrule
        \multirow{4}{*}{\textbf{$\mathrm{OOD}_{\mathrm{cond}}$}} & Steering angle & 0.887 & 0.786 & 0.423 & 0.892 & 0.795 & 0.414 & \textbf{0.924} & \textbf{0.854} & \textbf{0.350} \\
        & Lateral velocity & 0.928 & 0.860 & 0.390 & 0.932 & 0.868 & 0.377 & \textbf{0.954} & \textbf{0.910} & \textbf{0.312} \\
        & Lateral acceleration & 0.868 & 0.751 & 0.521 & 0.880 & 0.771 & 0.499 & \textbf{0.932} & \textbf{0.865} & \textbf{0.384} \\
        \midrule
        & \textbf{Variable macro} & 0.894 & 0.799 & 0.445 & 0.901 & 0.812 & 0.430 & \textbf{0.937} & \textbf{0.876} & \textbf{0.349} \\
        \bottomrule
    \end{tabular*}
\end{table}

\begin{table}[!htbp]
    \caption{\textbf{human motion: $\mathrm{ID}_{\mathrm{var}}$ variables.}}
    \label{tab:app_egoexo_readout_targets}
    \centering
    \fontsize{8.5}{10.5}\selectfont
    \setlength{\tabcolsep}{1pt}
    \renewcommand{\arraystretch}{1.06}
    \begin{tabular*}{\textwidth}{@{\extracolsep{\fill}}c>{\centering\arraybackslash}p{0.23\textwidth}ccc@{\hspace{5pt}}ccc@{\hspace{5pt}}ccc@{}}
        \toprule
         & & \multicolumn{3}{c}{Original} & \multicolumn{3}{c}{Video FT} & \multicolumn{3}{c}{Time-JEPA} \\
        \cmidrule(lr){3-5}\cmidrule(lr){6-8}\cmidrule(lr){9-11}
        Condition & State variable & $r\uparrow$ & $R^2\uparrow$ & nRMSE $\downarrow$ & $r\uparrow$ & $R^2\uparrow$ & nRMSE $\downarrow$ & $r\uparrow$ & $R^2\uparrow$ & nRMSE $\downarrow$ \\
        \midrule
        \multirow{10}{*}{\textbf{$\mathrm{ID}_{\mathrm{cond}}$}} & Pelvis speed & 0.582 & 0.334 & 0.652 & 0.589 & 0.343 & 0.648 & \textbf{0.696} & \textbf{0.472} & \textbf{0.581} \\
        & Left wrist relative speed & 0.518 & 0.259 & 0.682 & 0.519 & 0.260 & 0.682 & \textbf{0.559} & \textbf{0.306} & \textbf{0.660} \\
        & Right wrist relative speed & 0.519 & 0.267 & 0.613 & 0.522 & 0.269 & 0.613 & \textbf{0.557} & \textbf{0.304} & \textbf{0.598} \\
        & Left ankle relative speed & 0.256 & 0.058 & 0.736 & 0.257 & 0.058 & 0.736 & \textbf{0.323} & \textbf{0.095} & \textbf{0.721} \\
        & Right ankle relative speed & 0.437 & 0.167 & 0.404 & 0.442 & 0.168 & 0.404 & \textbf{0.534} & \textbf{0.234} & \textbf{0.388} \\
        & Left elbow flexion & 0.328 & 0.057 & 1.017 & 0.329 & 0.061 & 1.015 & \textbf{0.465} & \textbf{0.194} & \textbf{0.940} \\
        & Right elbow flexion & 0.436 & 0.122 & 0.988 & 0.436 & 0.126 & 0.985 & \textbf{0.557} & \textbf{0.294} & \textbf{0.886} \\
        & Left knee flexion & 0.743 & 0.493 & 0.744 & 0.747 & 0.501 & 0.738 & \textbf{0.829} & \textbf{0.666} & \textbf{0.604} \\
        & Right knee flexion & 0.681 & 0.290 & 0.858 & 0.684 & 0.297 & 0.853 & \textbf{0.783} & \textbf{0.544} & \textbf{0.687} \\
        \midrule
        & \textbf{Variable macro} & 0.500 & 0.228 & 0.744 & 0.503 & 0.231 & 0.742 & \textbf{0.589} & \textbf{0.345} & \textbf{0.674} \\
        \midrule
        \multirow{10}{*}{\textbf{$\mathrm{OOD}_{\mathrm{cond}}$}} & Pelvis speed & 0.558 & 0.303 & 0.667 & 0.564 & 0.309 & 0.665 & \textbf{0.689} & \textbf{0.461} & \textbf{0.587} \\
        & Left wrist relative speed & 0.483 & 0.229 & 0.696 & 0.486 & 0.232 & 0.694 & \textbf{0.539} & \textbf{0.287} & \textbf{0.669} \\
        & Right wrist relative speed & 0.487 & 0.232 & 0.628 & 0.490 & 0.236 & 0.627 & \textbf{0.548} & \textbf{0.296} & \textbf{0.601} \\
        & Left ankle relative speed & 0.258 & 0.054 & 0.737 & 0.262 & 0.056 & 0.737 & \textbf{0.320} & \textbf{0.085} & \textbf{0.725} \\
        & Right ankle relative speed & 0.436 & 0.162 & 0.406 & 0.442 & 0.167 & 0.405 & \textbf{0.549} & \textbf{0.254} & \textbf{0.383} \\
        & Left elbow flexion & 0.341 & 0.078 & 1.006 & 0.353 & 0.095 & 0.997 & \textbf{0.468} & \textbf{0.200} & \textbf{0.937} \\
        & Right elbow flexion & 0.424 & 0.127 & 0.985 & 0.432 & 0.140 & 0.978 & \textbf{0.562} & \textbf{0.307} & \textbf{0.878} \\
        & Left knee flexion & 0.695 & 0.378 & 0.825 & 0.702 & 0.400 & 0.810 & \textbf{0.799} & \textbf{0.606} & \textbf{0.657} \\
        & Right knee flexion & 0.638 & 0.140 & 0.944 & 0.639 & 0.158 & 0.934 & \textbf{0.757} & \textbf{0.489} & \textbf{0.728} \\
        \midrule
        & \textbf{Variable macro} & 0.480 & 0.189 & 0.766 & 0.486 & 0.199 & 0.761 & \textbf{0.581} & \textbf{0.332} & \textbf{0.685} \\
        \bottomrule
    \end{tabular*}
\end{table}

\begin{table}[!htbp]
    \caption{\textbf{human motion: $\mathrm{OOD}_{\mathrm{var}}$ variables.}}
    \label{tab:app_egoexo_shot_targets}
    \centering
    \fontsize{8.5}{10.5}\selectfont
    \setlength{\tabcolsep}{1pt}
    \renewcommand{\arraystretch}{1.06}
    \begin{tabular*}{\textwidth}{@{\extracolsep{\fill}}c>{\centering\arraybackslash}p{0.23\textwidth}ccc@{\hspace{5pt}}ccc@{\hspace{5pt}}ccc@{}}
        \toprule
         & & \multicolumn{3}{c}{Original} & \multicolumn{3}{c}{Video FT} & \multicolumn{3}{c}{Time-JEPA} \\
        \cmidrule(lr){3-5}\cmidrule(lr){6-8}\cmidrule(lr){9-11}
        Condition & State variable & $r\uparrow$ & $R^2\uparrow$ & nRMSE $\downarrow$ & $r\uparrow$ & $R^2\uparrow$ & nRMSE $\downarrow$ & $r\uparrow$ & $R^2\uparrow$ & nRMSE $\downarrow$ \\
        \midrule
        \multirow{6}{*}{\textbf{$\mathrm{ID}_{\mathrm{cond}}$}} & Inter-wrist distance & 0.573 & 0.316 & 0.893 & 0.575 & 0.320 & 0.891 & \textbf{0.643} & \textbf{0.400} & \textbf{0.836} \\
        & Left wrist--shoulder distance & 0.287 & 0.000 & 0.669 & 0.290 & 0.000 & 0.669 & \textbf{0.385} & \textbf{0.074} & \textbf{0.644} \\
        & Right wrist--shoulder distance & 0.472 & 0.177 & 0.635 & 0.471 & 0.176 & 0.635 & \textbf{0.533} & \textbf{0.262} & \textbf{0.601} \\
        & Inter-ankle distance & 0.223 & \(-\)0.109 & 0.370 & 0.215 & \(-\)0.121 & 0.372 & \textbf{0.247} & \textbf{\(-\)0.089} & \textbf{0.367} \\
        & Pelvis support height & 0.614 & 0.277 & 0.451 & 0.621 & 0.289 & 0.447 & \textbf{0.710} & \textbf{0.462} & \textbf{0.389} \\
        \midrule
        & \textbf{Variable macro} & 0.434 & 0.132 & 0.603 & 0.434 & 0.133 & 0.603 & \textbf{0.504} & \textbf{0.222} & \textbf{0.567} \\
        \midrule
        \multirow{6}{*}{\textbf{$\mathrm{OOD}_{\mathrm{cond}}$}} & Inter-wrist distance & 0.571 & 0.318 & 0.891 & 0.573 & 0.320 & 0.890 & \textbf{0.640} & \textbf{0.396} & \textbf{0.839} \\
        & Left wrist--shoulder distance & 0.361 & 0.086 & 0.640 & 0.365 & 0.091 & 0.638 & \textbf{0.428} & \textbf{0.129} & \textbf{0.624} \\
        & Right wrist--shoulder distance & 0.492 & 0.219 & 0.618 & 0.494 & 0.220 & 0.618 & \textbf{0.548} & \textbf{0.277} & \textbf{0.595} \\
        & Inter-ankle distance & 0.232 & \(-\)0.162 & 0.379 & 0.228 & \(-\)0.150 & 0.377 & \textbf{0.258} & \textbf{\(-\)0.103} & \textbf{0.369} \\
        & Pelvis support height & 0.593 & 0.173 & 0.483 & 0.601 & 0.194 & 0.477 & \textbf{0.704} & \textbf{0.426} & \textbf{0.402} \\
        \midrule
        & \textbf{Variable macro} & 0.450 & 0.127 & 0.602 & 0.452 & 0.135 & 0.600 & \textbf{0.515} & \textbf{0.225} & \textbf{0.566} \\
        \bottomrule
    \end{tabular*}
\end{table}

\section{Time-Series Alignment and Preprocessing}
\label{app:time_series_preprocessing}
\setcounter{table}{0}

This appendix defines the temporal construction shared across the three
domains. Appendix~\ref{app:datasets} describes the data sources, state
variables, and domain-specific splits.

\subsection{Common Temporal Construction}

The main V-JEPA experiments sample 64 RGB frames at 5 FPS.
Each video tubelet contains two sampled frames, producing state values at 32
temporal positions.
For scalar dimension $j$, let $y_{r,j}$ denote the physical-state value at
sampled frame $r$ and $v_{r,j}\in\{0,1\}$ its validity indicator.
For tubelet $k$, the aligned state value is
\begin{equation}
    \bar y_{k,j}
    = \frac{\sum_{r\in\{2k-1,2k\}} v_{r,j}y_{r,j}}
           {\sum_{r\in\{2k-1,2k\}} v_{r,j}},
    \qquad
    \sum_{r\in\{2k-1,2k\}} v_{r,j}>0.
    \label{eq:tubelet_target}
\end{equation}
We mask each tubelet--variable entry that lacks valid frames and compute
training-set means and standard deviations from valid values. We standardize
each available state value as
$\widetilde y_{k,j}=(\bar y_{k,j}-\mu_j)/\sigma_j$ and retain the validity mask
through normalization and the Smooth-L1 objective. Current-state probes on
frozen encoders predict the 32 aligned time steps.
Appendix~\ref{app:future_state} defines the future-state labels.

\section{Data Sources, State Variables, and Splits}
\label{app:datasets}
\setcounter{table}{0}

This appendix defines the data sources, physical state variables, and split
construction for Section~\ref{sec:experimental_protocol}. We assign disjoint
split roles to encoder fine-tuning, probe training, probe validation, and
testing. For robot manipulation and human motion, we test the same physical
windows through $\mathrm{ID}_{\mathrm{cond}}$ and
$\mathrm{OOD}_{\mathrm{cond}}$ cameras. We estimate normalization statistics
from the relevant training split and track validity separately for each state
variable.

\begin{table}[htbp]
    \caption{\textbf{Data roles and evaluated views.}
    $n\!\times\!v$ denotes $n$ physical windows with $v$ views each.
    For robot manipulation, we expand probe splits over views. For human motion,
    we sample one seen view per training access, fix one per validation window,
    and evaluate all three seen views for $\mathrm{ID}_{\mathrm{cond}}$ testing. Each autonomous driving
    clip provides one view.}
    \label{tab:app_dataset_splits}
    \centering
    \fontsize{8.5}{10.5}\selectfont
    \setlength{\tabcolsep}{0pt}
    \begin{tabularx}{\textwidth}{@{}X@{}}
        \appendixcard{autonomous driving}{\textbf{Fine-tuning:} 6,305 clips. \textbf{Probe train:} 1,349 clips. \textbf{Probe validation:} 449 clips. \textbf{$\mathrm{ID}_{\mathrm{cond}}$ test:} 900 clips. \textbf{$\mathrm{OOD}_{\mathrm{cond}}$ test:} 521 clips from date-disjoint caption-defined stable rain.} \\
        \noalign{\vskip 3pt}
        \appendixcard{robot manipulation}{\textbf{Fine-tuning:} 15,000 windows. \textbf{Probe train:} $1{,}500\!\times\!3$. \textbf{Probe validation:} $500\!\times\!3$. \textbf{$\mathrm{ID}_{\mathrm{cond}}$ test:} $1{,}000\!\times\!3$. \textbf{$\mathrm{OOD}_{\mathrm{cond}}$ test:} $1{,}000\!\times\!2$ from held-out left/right fisheye cameras.} \\
        \noalign{\vskip 3pt}
        \appendixcard{human motion}{\textbf{Fine-tuning:} 7,000 windows. \textbf{Probe train:} 1,500 windows. \textbf{Probe validation:} 500 windows. \textbf{$\mathrm{ID}_{\mathrm{cond}}$ test:} $1{,}000\!\times\!3$. \textbf{$\mathrm{OOD}_{\mathrm{cond}}$ test:} the same 1,000 test windows from held-out exocentric \texttt{view03}.}
    \end{tabularx}
\end{table}

\subsection{CoVLA Data Source}

\paragraph{Source data.}
CoVLA~\citep{arai2025covla} contains 10,000 30-s forward-facing driving videos
with temporally aligned CAN and inertial states, trajectories, and scene
descriptions.
\par\noindent\url{https://huggingface.co/datasets/turing-motors/CoVLA-Dataset}\par
We use RGB video, vehicle states, and trajectories, and construct
the weather split from captions.

\paragraph{State-variable selection.}
The three $\mathrm{ID}_{\mathrm{var}}$ variables cover longitudinal speed, longitudinal
acceleration, and turning motion. We use calibrated-frame $z$ angular
velocity for yaw because it agrees with heading change. We reserve steering
angle, lateral velocity, and lateral acceleration as
$\mathrm{OOD}_{\mathrm{var}}$ variables and use
the future trajectory as an independent reference for reconstruction.

\paragraph{Split and weather criterion.}
We separate CoVLA split roles at both video and date levels.
The $\mathrm{OOD}_{\mathrm{cond}}$ split uses caption-defined rain: at least eight of the ten caption
frames $0,60,\ldots,540$ must describe rainy weather.
This criterion yields the 521-clip $\mathrm{OOD}_{\mathrm{cond}}$ test set.

\paragraph{Alignment.}
CoVLA records video at 20 FPS. Sampling every fourth frame gives 64 frames
at 5 FPS, spanning 12.6 s. We map frame indices to timestamped vehicle-state
records before tubelet averaging and use the sampled timestamps for
trajectory integration.

\begin{table}[htbp]
    \caption{\textbf{autonomous driving state variables.}
    Variable groups follow Section~\ref{sec:protocol_domains}. Both groups use
    supervised probes on frozen encoders.}
    \label{tab:app_covla_targets}
    \centering
    \setlength{\tabcolsep}{0pt}
    \begin{tabularx}{\textwidth}{@{}X@{}}
        \appendixcard{$\mathrm{ID}_{\mathrm{var}}$ (3 variables)}{%
            \appendixcarditem{Longitudinal ego velocity}{(m/s)}
            \appendixcarditem{Longitudinal ego acceleration}{(m/s$^2$)}
            \appendixcarditem{Calibrated-frame yaw angular velocity}{(rad/s)}
        } \\
        \noalign{\vskip 3pt}
        \appendixcard{$\mathrm{OOD}_{\mathrm{var}}$ (3 variables)}{%
            \appendixcarditem{Steering-wheel angle}{(control quantity, degree)}
            \appendixcarditem{Calibrated-frame lateral velocity}{(m/s)}
            \appendixcarditem{Calibrated-frame lateral acceleration}{(m/s$^2$)}
        }
    \end{tabularx}
\end{table}

\subsection{AgiBot Data Source}

\paragraph{Source data.}
AgiBot~\citep{bu2025agibot} provides synchronized multi-camera video and
full-body robot state.
\par\noindent\url{https://huggingface.co/datasets/agibot-world/AgiBotWorld2026}\par
The G2 state stream includes base and waist pose,
bilateral joint position, velocity and effort, end-effector pose and wrench,
and gripper state. The encoder receives selected RGB views.

\paragraph{State-variable selection.}
The 18 $\mathrm{ID}_{\mathrm{var}}$ variables cover base motion, torso configuration, bilateral
end-effector position and signed velocity, and gripper state. Two principal
components summarize the 5D waist state. Signed velocity preserves the
direction needed for reach-rate reconstruction. The eight
$\mathrm{OOD}_{\mathrm{var}}$ variables
summarize bilateral arm configuration, joint motion, effort, and contact force.

\paragraph{Views and leakage control.}
The seen views comprise one overhead and two hand-mounted cameras.
$\mathrm{OOD}_{\mathrm{cond}}$ uses two held-out fisheye cameras. We assign split roles by a
camera-agnostic episode/window key before expanding views, keeping every
recording of a physical interval in one role. $\mathrm{ID}_{\mathrm{cond}}$ and
$\mathrm{OOD}_{\mathrm{cond}}$ test the same
1,000 physical windows.

\paragraph{State alignment and derived quantities.}
We align video timestamps to the LeRobot state timeline within 0.05 s before
tubelet averaging. We fit waist PCA on the fine-tuning split and retain its
first two axes. End-effector velocities use finite differences of arm
position on the native state timeline.

\begin{table}[htbp]
    \caption{\textbf{robot manipulation state variables.}
    Variable groups follow Section~\ref{sec:protocol_domains}. Parentheses give the number of
    scalar dimensions in each bilateral or vector group.}
    \label{tab:app_agibot_targets}
    \centering
    \fontsize{8.5}{10.5}\selectfont
    \setlength{\tabcolsep}{0pt}
    \begin{tabularx}{\textwidth}{@{}X@{}}
        \appendixcard{$\mathrm{ID}_{\mathrm{var}}$ (18 variables)}{%
            \appendixcarditem{Planar base translation-speed magnitude}{(1D, m/s)}
            \appendixcarditem{Base angular-speed magnitude}{(1D, rad/s)}
            \appendixcarditem{First two frozen PCA scores of the 5D waist configuration}{(2D, PCA score)}
            \appendixcarditem{Signed left/right end-effector $xyz$ position}{(6D, m)}
            \appendixcarditem{Signed left/right end-effector $xyz$ velocity in the recorded frame}{(6D, m/s)}
            \appendixcarditem{Bilateral gripper opening state}{(2D, native scalar)}
        } \\
        \noalign{\vskip 3pt}
        \appendixcard{$\mathrm{OOD}_{\mathrm{var}}$ (8 variables)}{%
            \appendixcarditem{L2 norm of each arm's 7D joint configuration}{(2D, rad)}
            \appendixcarditem{L2 norm of each arm's joint velocity}{(2D, rad/s)}
            \appendixcarditem{L2 norm of bilateral joint effort}{(2D, native effort)}
            \appendixcarditem{Bilateral end-effector force magnitude}{(2D, native force)}
        }
    \end{tabularx}
\end{table}

\subsection{EgoExo4D Data Source}

\paragraph{Source data.}
EgoExo4D~\citep{grauman2024egoexo4d} provides synchronized egocentric and
exocentric recordings of skilled activities.
\par\noindent\url{https://docs.ego-exo4d-data.org/download/}\par
We use exocentric RGB video and 3D body-pose annotations.

\paragraph{Variables and splits.}
The $\mathrm{ID}_{\mathrm{var}}$ panel captures whole-body translation, local limb
motion, and joint articulation. The $\mathrm{OOD}_{\mathrm{var}}$ panel covers
bimanual coordination, arm extension, stance width, and support height.
Fine-tuning and $\mathrm{ID}_{\mathrm{cond}}$ probing use exocentric
\texttt{view01}, \texttt{view02}, and \texttt{view04};
$\mathrm{OOD}_{\mathrm{cond}}$ uses held-out exocentric \texttt{view03}.
Split roles are isolated at the take, participant, and capture levels, and both
condition settings evaluate the same 1,000 physical test windows.

\paragraph{Temporal processing.}
We apply fixed smoothing and temporal alignment, retain valid measurements with
at least two supporting views, and average them within each two-frame tubelet.
Coordinates and distances use meters, speeds use m/s, and flexion angles use
radians.

\begin{table}[htbp]
    \caption{\textbf{human motion state variables.}
    Variable groups follow Section~\ref{sec:protocol_domains}.}
    \label{tab:app_egoexo_targets}
    \centering
    \fontsize{8.5}{10.5}\selectfont
    \setlength{\tabcolsep}{0pt}
    \begin{tabularx}{\textwidth}{@{}X@{}}
        \appendixcard{$\mathrm{ID}_{\mathrm{var}}$ (9 variables)}{%
            \appendixcarditem{Pelvis speed:}{speed of the midpoint between the left and right hips (1D, m/s)}
            \appendixcarditem{Left/right wrist relative speed:}{wrist speed after subtracting pelvis translation (2D, m/s)}
            \appendixcarditem{Left/right ankle relative speed:}{ankle speed after subtracting pelvis translation (2D, m/s)}
            \appendixcarditem{Left/right elbow flexion:}{interior shoulder--elbow--wrist angle (2D, rad)}
            \appendixcarditem{Left/right knee flexion:}{interior hip--knee--ankle angle (2D, rad)}
        } \\
        \noalign{\vskip 3pt}
        \appendixcard{$\mathrm{OOD}_{\mathrm{var}}$ (5 variables)}{%
            \appendixcarditem{Inter-wrist distance:}{Euclidean distance between the two wrists (1D, m)}
            \appendixcarditem{Left/right wrist--shoulder distance:}{Euclidean shoulder-to-wrist endpoint distance (2D, m)}
            \appendixcarditem{Inter-ankle distance:}{Euclidean distance between the two ankles (1D, m)}
            \appendixcarditem{Pelvis support height:}{pelvis height relative to the two-foot support reference (1D, m)}
        }
    \end{tabularx}
\end{table}

\FloatBarrier
\section{Kinematic Reconstruction}
\label{app:grounded_details}
\setcounter{table}{0}

Section~\ref{sec:grounded_consistency} reports one primary kinematic-reconstruction error for each domain. This appendix explains each domain-specific reconstruction and provides finer-grained results. For each encoder, we train separate probes on frozen video features to predict the physical quantities required by a fixed law. We use these predictions to reconstruct a trajectory endpoint, reach rate, or limb-endpoint distance, then compare the result with independently constructed ground truth. For autonomous driving and human motion, another probe predicts the same endpoint or distance directly. The tables label this error as Direct. As an additional diagnostic, we compare each reconstruction with fixed alternatives: incorrect physical relations for autonomous driving and human motion, and cross-arm joint-state–velocity pairings for robot manipulation. The score records how often the correct relation or same-arm pairing gives the lower error. Applying the same computation to ground-truth inputs reveals error introduced by integration, calibration, or pose annotations. Training and validation data determine the coordinate conventions, formulas, thresholds, and alternatives before test evaluation. The tables report results by trajectory horizon, camera, or limb. We select each probe by validation loss and report the mean and sample standard deviation across five probe seeds (0--4).

Tables~\ref{tab:app_covla_physics_horizons}--\ref{tab:app_egoexo_physics_limbs}
give single-domain diagnostics. Table~\ref{tab:grounded_consistency} reports
the multi-domain comparison, and Table~\ref{tab:mix_reconstruction} compares
single-domain and multi-domain Time-JEPA.

\subsection{autonomous driving: Integrated Trajectory Reconstruction}

Using probe-training ground truth, The resulting coordinate convention is
\begin{equation*}
    \widetilde{\mathbf v}_k
    = \begin{bmatrix}\widehat v_{x,k} & -\widehat v_{y,k}\end{bmatrix}^{\!\top},
    \qquad
    \widetilde\omega_k=-\widehat\omega_{z,k}.
\end{equation*}
Starting from $\mathbf p_0=\mathbf 0$ and $\phi_0=0$, the measured interval
$\Delta t_k$ gives
\begin{align}
    \phi_{k+1}
    &= \phi_k + \frac{\Delta t_k}{2}
       \left(\widetilde\omega_k+\widetilde\omega_{k+1}\right),
       \label{eq:covla_yaw_integration}\\
    \mathbf p_{k+1}
    &= \mathbf p_k + \frac{\Delta t_k}{2}
       \left[R(\phi_k)\widetilde{\mathbf v}_k+
       R(\phi_{k+1})\widetilde{\mathbf v}_{k+1}\right],
       \label{eq:covla_position_integration}
\end{align}
where $R(\phi)$ is the planar rotation matrix.
Recorded trajectory indices 20, 40, and 59 correspond to 1.00, 2.00, and
2.95 s.
A separate 2D probe predicts each direct endpoint independently of the three
state probes used for integration.

For law discrimination, the positive candidate uses the frozen convention.
The negative pool contains: reversed yaw sign, reversed lateral-velocity sign,
a one-token temporal offset, state sequences from another video, and an endpoint
evaluated at the wrong horizon.
If $C$ and $N$ are the positive and negative endpoint errors, respectively, the
reported score is
\begin{equation*}
    \operatorname{AUROC}=P(C<N)+\tfrac12 P(C=N),
\end{equation*}

\begin{table}[htbp]
    \caption{\textbf{autonomous driving trajectory reconstruction by horizon.}
    Direct and integrated errors use endpoint MAE in meters. AUROC compares the
    correct integration law with the fixed alternatives. Values give mean
    $\pm$ sample standard deviation across five probe seeds.
    Bold marks the best encoder mean.}
    \label{tab:app_covla_physics_horizons}
    \centering
    \fontsize{8.5}{10.5}\selectfont
    \setlength{\tabcolsep}{5pt}
    \begin{tabular}{@{}cccccc@{}}
        \toprule
        Split & Horizon & Encoder & Direct $\downarrow$ & Integrated $\downarrow$
        & AUROC $\uparrow$ \\
        \midrule
        $\mathrm{ID}_{\mathrm{cond}}$ & 1.00 s & Original & 2.226\(\,\pm\,\)0.092 & 2.199\(\,\pm\,\)0.057 & 0.642\(\,\pm\,\)0.002 \\
        $\mathrm{ID}_{\mathrm{cond}}$ & 1.00 s & Video FT & 2.124\(\,\pm\,\)0.108 & 2.083\(\,\pm\,\)0.064 & 0.649\(\,\pm\,\)0.003 \\
        $\mathrm{ID}_{\mathrm{cond}}$ & 1.00 s & Time-JEPA & \textbf{1.408\(\,\pm\,\)0.096} & \textbf{1.242\(\,\pm\,\)0.050} & \textbf{0.707\(\,\pm\,\)0.004} \\
        \addlinespace[2pt]
        $\mathrm{ID}_{\mathrm{cond}}$ & 2.00 s & Original & 4.728\(\,\pm\,\)0.097 & 4.436\(\,\pm\,\)0.116 & 0.636\(\,\pm\,\)0.004 \\
        $\mathrm{ID}_{\mathrm{cond}}$ & 2.00 s & Video FT & 4.528\(\,\pm\,\)0.098 & 4.194\(\,\pm\,\)0.129 & 0.647\(\,\pm\,\)0.004 \\
        $\mathrm{ID}_{\mathrm{cond}}$ & 2.00 s & Time-JEPA & \textbf{3.000\(\,\pm\,\)0.126} & \textbf{2.467\(\,\pm\,\)0.097} & \textbf{0.724\(\,\pm\,\)0.004} \\
        \addlinespace[2pt]
        $\mathrm{ID}_{\mathrm{cond}}$ & 2.95 s & Original & 6.994\(\,\pm\,\)0.128 & 6.634\(\,\pm\,\)0.172 & 0.632\(\,\pm\,\)0.006 \\
        $\mathrm{ID}_{\mathrm{cond}}$ & 2.95 s & Video FT & 6.656\(\,\pm\,\)0.120 & 6.262\(\,\pm\,\)0.192 & 0.645\(\,\pm\,\)0.007 \\
        $\mathrm{ID}_{\mathrm{cond}}$ & 2.95 s & Time-JEPA & \textbf{4.407\(\,\pm\,\)0.154} & \textbf{3.648\(\,\pm\,\)0.139} & \textbf{0.732\(\,\pm\,\)0.004} \\
        \midrule
        $\mathrm{OOD}_{\mathrm{cond}}$ & 1.00 s & Original & 2.424\(\,\pm\,\)0.110 & 2.394\(\,\pm\,\)0.101 & 0.635\(\,\pm\,\)0.004 \\
        $\mathrm{OOD}_{\mathrm{cond}}$ & 1.00 s & Video FT & 2.317\(\,\pm\,\)0.081 & 2.248\(\,\pm\,\)0.094 & 0.641\(\,\pm\,\)0.004 \\
        $\mathrm{OOD}_{\mathrm{cond}}$ & 1.00 s & Time-JEPA & \textbf{1.645\(\,\pm\,\)0.206} & \textbf{1.428\(\,\pm\,\)0.156} & \textbf{0.687\(\,\pm\,\)0.009} \\
        \addlinespace[2pt]
        $\mathrm{OOD}_{\mathrm{cond}}$ & 2.00 s & Original & 4.983\(\,\pm\,\)0.398 & 4.836\(\,\pm\,\)0.199 & 0.628\(\,\pm\,\)0.004 \\
        $\mathrm{OOD}_{\mathrm{cond}}$ & 2.00 s & Video FT & 4.787\(\,\pm\,\)0.377 & 4.532\(\,\pm\,\)0.184 & 0.637\(\,\pm\,\)0.004 \\
        $\mathrm{OOD}_{\mathrm{cond}}$ & 2.00 s & Time-JEPA & \textbf{3.306\(\,\pm\,\)0.643} & \textbf{2.841\(\,\pm\,\)0.305} & \textbf{0.705\(\,\pm\,\)0.009} \\
        \addlinespace[2pt]
        $\mathrm{OOD}_{\mathrm{cond}}$ & 2.95 s & Original & 7.651\(\,\pm\,\)0.701 & 7.256\(\,\pm\,\)0.296 & 0.614\(\,\pm\,\)0.009 \\
        $\mathrm{OOD}_{\mathrm{cond}}$ & 2.95 s & Video FT & 7.206\(\,\pm\,\)0.608 & 6.789\(\,\pm\,\)0.274 & 0.627\(\,\pm\,\)0.009 \\
        $\mathrm{OOD}_{\mathrm{cond}}$ & 2.95 s & Time-JEPA & \textbf{5.152\(\,\pm\,\)1.047} & \textbf{4.211\(\,\pm\,\)0.442} & \textbf{0.707\(\,\pm\,\)0.010} \\
        \bottomrule
    \end{tabular}
\end{table}

Trapezoidal integration of ground-truth velocity and yaw-rate sequences gives
deterministic endpoint MAEs of 0.321/0.639/0.953 m at 1.00/2.00/2.95 s under
$\mathrm{ID}_{\mathrm{cond}}$, and 0.273/0.526/0.757 m under rainy
$\mathrm{OOD}_{\mathrm{cond}}$.

Time-JEPA improves direct error, integrated-trajectory error, and law AUROC over the video-fine-tuned control at
every horizon under $\mathrm{ID}_{\mathrm{cond}}$ and rainy
$\mathrm{OOD}_{\mathrm{cond}}$. Errors increase with horizon for every row.

\subsection{robot manipulation: Reach-Rate Reconstruction}

For arm $a$, the official URDF maps the independently decoded 7D joint state to
the hand position.
We estimate rigid calibration and shoulder points from the fine-tuning split
and fix them for evaluation:
\begin{equation*}
    \begin{aligned}
    \widehat{\mathbf p}^{\mathrm{URDF}}_{a,t}
        &= FK_a^{\mathrm{URDF}}(\widehat{\mathbf q}_{a,t}),\\
    \widehat{\mathbf p}_{a,t}
        &= R_a\widehat{\mathbf p}^{\mathrm{URDF}}_{a,t}+\mathbf b_a,\\
    \widehat{\mathbf u}_{a,t}
        &= \frac{\widehat{\mathbf p}_{a,t}-\mathbf s_a}
        {\|\widehat{\mathbf p}_{a,t}-\mathbf s_a\|_2}.
    \end{aligned}
\end{equation*}
\begin{table}[htbp]
    \caption{\textbf{robot manipulation reach-rate reconstruction by camera.}
    Left/right columns give MAE in cm/s. Pair rate averages the two arms per
    camera. Values give mean $\pm$ sample standard deviation across five probe
    seeds. Bold marks the best encoder mean per camera and metric.}
    \label{tab:app_agibot_physics_views}
    \centering
    \fontsize{8.5}{10.5}\selectfont
    \setlength{\tabcolsep}{5pt}
    \begin{tabular}{@{}cccccc@{}}
        \toprule
        Split & Camera & Encoder & Left $\downarrow$ & Right $\downarrow$
        & Pair rate $\uparrow$ \\
        \midrule
        $\mathrm{ID}_{\mathrm{cond}}$ & Top head & Original & 4.130\(\,\pm\,\)0.022 & 7.209\(\,\pm\,\)0.019 & 0.502\(\,\pm\,\)0.018 \\
        $\mathrm{ID}_{\mathrm{cond}}$ & Top head & Video FT & 4.094\(\,\pm\,\)0.016 & 7.025\(\,\pm\,\)0.026 & 0.492\(\,\pm\,\)0.016 \\
        $\mathrm{ID}_{\mathrm{cond}}$ & Top head & Time-JEPA & \textbf{3.137\(\,\pm\,\)0.045} & \textbf{3.745\(\,\pm\,\)0.043} & \textbf{0.553\(\,\pm\,\)0.007} \\
        \addlinespace[2pt]
        $\mathrm{ID}_{\mathrm{cond}}$ & Left hand & Original & 3.805\(\,\pm\,\)0.012 & 7.397\(\,\pm\,\)0.022 & 0.516\(\,\pm\,\)0.017 \\
        $\mathrm{ID}_{\mathrm{cond}}$ & Left hand & Video FT & 3.751\(\,\pm\,\)0.018 & 7.252\(\,\pm\,\)0.017 & 0.541\(\,\pm\,\)0.030 \\
        $\mathrm{ID}_{\mathrm{cond}}$ & Left hand & Time-JEPA & \textbf{3.010\(\,\pm\,\)0.027} & \textbf{3.863\(\,\pm\,\)0.032} & \textbf{0.616\(\,\pm\,\)0.019} \\
        \addlinespace[2pt]
        $\mathrm{ID}_{\mathrm{cond}}$ & Right hand & Original & 4.167\(\,\pm\,\)0.029 & 6.432\(\,\pm\,\)0.015 & 0.578\(\,\pm\,\)0.031 \\
        $\mathrm{ID}_{\mathrm{cond}}$ & Right hand & Video FT & 4.131\(\,\pm\,\)0.029 & 6.312\(\,\pm\,\)0.016 & 0.584\(\,\pm\,\)0.017 \\
        $\mathrm{ID}_{\mathrm{cond}}$ & Right hand & Time-JEPA & \textbf{2.925\(\,\pm\,\)0.032} & \textbf{4.047\(\,\pm\,\)0.023} & \textbf{0.591\(\,\pm\,\)0.026} \\
        \midrule
        $\mathrm{OOD}_{\mathrm{cond}}$ & Left fisheye & Original & 4.228\(\,\pm\,\)0.040 & 7.568\(\,\pm\,\)0.051 & 0.455\(\,\pm\,\)0.013 \\
        $\mathrm{OOD}_{\mathrm{cond}}$ & Left fisheye & Video FT & 4.217\(\,\pm\,\)0.043 & 7.532\(\,\pm\,\)0.051 & 0.491\(\,\pm\,\)0.023 \\
        $\mathrm{OOD}_{\mathrm{cond}}$ & Left fisheye & Time-JEPA & \textbf{3.254\(\,\pm\,\)0.095} & \textbf{7.000\(\,\pm\,\)0.032} & \textbf{0.550\(\,\pm\,\)0.015} \\
        \addlinespace[2pt]
        $\mathrm{OOD}_{\mathrm{cond}}$ & Right fisheye & Original & 4.241\(\,\pm\,\)0.042 & 7.569\(\,\pm\,\)0.059 & 0.489\(\,\pm\,\)0.010 \\
        $\mathrm{OOD}_{\mathrm{cond}}$ & Right fisheye & Video FT & 4.259\(\,\pm\,\)0.037 & 7.415\(\,\pm\,\)0.057 & 0.532\(\,\pm\,\)0.014 \\
        $\mathrm{OOD}_{\mathrm{cond}}$ & Right fisheye & Time-JEPA & \textbf{4.183\(\,\pm\,\)0.027} & \textbf{4.402\(\,\pm\,\)0.074} & \textbf{0.588\(\,\pm\,\)0.024} \\
        \bottomrule
    \end{tabular}
\end{table}

We compute this direction only when the predicted shoulder-to-hand distance exceeds \(10^{-8}\,\mathrm{m}\). Before testing, we use the probe-training and validation splits to check the calibrated joint-to-hand mapping, the reach rate computed from ground-truth inputs, and the number of valid samples. These checks fix the validity threshold and the timestamp-based calculation of ground-truth reach rate. A separate probe predicts the hand velocity \(\widehat{\mathbf v}_{a,t}\). Projecting this velocity onto the shoulder-to-hand direction gives the signed reach rate:
\begin{equation}
    \widehat{\dot\rho}_{a,t}
    = \widehat{\mathbf u}_{a,t}^{\top}\widehat{\mathbf v}_{a,t}.
    \label{eq:agibot_reach_rate}
\end{equation}
We compute the ground-truth signed reach rate from raw shoulder-to-hand distances and source timestamps. For interior samples, we use the distance change between the preceding and following samples; at each sequence boundary, we use the nearest two samples.

The pairing diagnostic holds velocity fixed, substitutes the opposite-arm joint
state, and reports how often the correct pairing is closer to ground truth (half
credit for ties).

The left-fisheye/right-arm error remains comparatively high, while Time-JEPA
improves the macro errors and correct-pairing rate reported in
Section~\ref{sec:grounded_consistency}.
Passing ground-truth joint angles and endpoint velocities through the same
calibrated projection gives deterministic reference errors of 0.063 cm/s for the
left arm and 0.084 cm/s for the right arm on both evaluated camera families.

\FloatBarrier
\subsection{human motion: Cosine-Law Distance Reconstruction}

For elbows, $(L_1,L_2,d)$ denote upper-arm length, forearm length, and
shoulder--wrist distance. For knees they denote thigh length, shin length, and
hip--ankle distance. Ground-truth checks on the
probe-training and validation splits fix the internal-angle convention,
$[0,\pi]$ clipping, morphology rule, nonnegative square-root guard, and three
alternatives after verifying coverage, ground-truth-input error, law AUROC, and the
correct-law margin. The distance reconstruction is
\begin{equation}
    \widehat d^{\mathrm{law}}
    = \sqrt{\max\!\left(
        L_1^2+L_2^2-2L_1L_2
        \cos\!\left(\operatorname{clip}(\widehat\theta,0,\pi)\right),0
      \right)}.
    \label{eq:egoexo_cosine_closure}
\end{equation}
The alternatives use $\pi-\widehat\theta$, the opposite-side limb angle, or a
one-token temporal shift. Table~\ref{tab:app_egoexo_physics_limbs} reports AUROC averaged over all
activities.

\begin{table}[!htbp]
    \caption{\textbf{human motion cosine-law distance reconstruction.}
    Direct and cosine-law errors use MAE in cm. AUROC compares the correct
    relation with three fixed alternatives. Values give mean $\pm$ sample
    standard deviation across five probe seeds. Bold marks the best encoder
    mean for each limb, split, and metric.}
    \label{tab:app_egoexo_physics_limbs}
    \centering
    \fontsize{8.5}{10.5}\selectfont
    \setlength{\tabcolsep}{1.5pt}
    \begin{tabular*}{\textwidth}{@{\extracolsep{\fill}}cccccccc@{}}
        \toprule
        & & \multicolumn{3}{c}{$\mathrm{ID}_{\mathrm{cond}}$}
        & \multicolumn{3}{c}{$\mathrm{OOD}_{\mathrm{cond}}$} \\
        \cmidrule(lr){3-5}\cmidrule(lr){6-8}
        Limb & Encoder & Direct $\downarrow$ & Cosine $\downarrow$ & AUROC $\uparrow$
        & Direct $\downarrow$ & Cosine $\downarrow$ & AUROC $\uparrow$ \\
        \midrule
        Elbow L & Original & 7.017\(\,\pm\,\)0.089 & 6.657\(\,\pm\,\)0.103 & 0.561\(\,\pm\,\)0.004 & 7.111\(\,\pm\,\)0.110 & 6.868\(\,\pm\,\)0.178 & 0.554\(\,\pm\,\)0.006 \\
        Elbow L & Video FT & 7.028\(\,\pm\,\)0.091 & 6.671\(\,\pm\,\)0.110 & 0.562\(\,\pm\,\)0.004 & 7.108\(\,\pm\,\)0.113 & 6.845\(\,\pm\,\)0.189 & 0.556\(\,\pm\,\)0.006 \\
        Elbow L & Time-JEPA & \textbf{6.759\(\,\pm\,\)0.091} & \textbf{6.404\(\,\pm\,\)0.103} & \textbf{0.570\(\,\pm\,\)0.006} & \textbf{6.790\(\,\pm\,\)0.136} & \textbf{6.527\(\,\pm\,\)0.189} & \textbf{0.565\(\,\pm\,\)0.006} \\
        \addlinespace[3pt]
        Elbow R & Original & 6.786\(\,\pm\,\)0.025 & 6.597\(\,\pm\,\)0.020 & 0.568\(\,\pm\,\)0.002 & 6.880\(\,\pm\,\)0.117 & 6.749\(\,\pm\,\)0.066 & 0.564\(\,\pm\,\)0.004 \\
        Elbow R & Video FT & 6.799\(\,\pm\,\)0.037 & 6.628\(\,\pm\,\)0.023 & 0.567\(\,\pm\,\)0.002 & 6.884\(\,\pm\,\)0.121 & 6.771\(\,\pm\,\)0.072 & 0.563\(\,\pm\,\)0.004 \\
        Elbow R & Time-JEPA & \textbf{6.482\(\,\pm\,\)0.033} & \textbf{6.360\(\,\pm\,\)0.045} & \textbf{0.573\(\,\pm\,\)0.003} & \textbf{6.501\(\,\pm\,\)0.125} & \textbf{6.420\(\,\pm\,\)0.059} & \textbf{0.570\(\,\pm\,\)0.003} \\
        \addlinespace[3pt]
        Knee L & Original & 10.162\(\,\pm\,\)0.326 & 7.617\(\,\pm\,\)0.180 & \textbf{0.657\(\,\pm\,\)0.007} & 10.802\(\,\pm\,\)0.267 & 8.142\(\,\pm\,\)0.348 & 0.657\(\,\pm\,\)0.007 \\
        Knee L & Video FT & 10.155\(\,\pm\,\)0.325 & 7.570\(\,\pm\,\)0.206 & 0.657\(\,\pm\,\)0.009 & 10.763\(\,\pm\,\)0.290 & 8.070\(\,\pm\,\)0.371 & 0.657\(\,\pm\,\)0.009 \\
        Knee L & Time-JEPA & \textbf{9.034\(\,\pm\,\)0.291} & \textbf{6.590\(\,\pm\,\)0.170} & 0.655\(\,\pm\,\)0.007 & \textbf{9.515\(\,\pm\,\)0.318} & \textbf{6.865\(\,\pm\,\)0.275} & \textbf{0.659\(\,\pm\,\)0.008} \\
        \addlinespace[3pt]
        Knee R & Original & 9.369\(\,\pm\,\)0.499 & 6.636\(\,\pm\,\)0.280 & 0.641\(\,\pm\,\)0.008 & 10.047\(\,\pm\,\)0.538 & 7.156\(\,\pm\,\)0.528 & 0.637\(\,\pm\,\)0.008 \\
        Knee R & Video FT & 9.350\(\,\pm\,\)0.508 & 6.582\(\,\pm\,\)0.321 & 0.642\(\,\pm\,\)0.010 & 10.010\(\,\pm\,\)0.554 & 7.102\(\,\pm\,\)0.531 & 0.638\(\,\pm\,\)0.009 \\
        Knee R & Time-JEPA & \textbf{8.209\(\,\pm\,\)0.462} & \textbf{5.566\(\,\pm\,\)0.226} & \textbf{0.649\(\,\pm\,\)0.007} & \textbf{8.721\(\,\pm\,\)0.450} & \textbf{5.879\(\,\pm\,\)0.369} & \textbf{0.642\(\,\pm\,\)0.008} \\
        \bottomrule
    \end{tabular*}
\end{table}

Ground-truth angles give deterministic ground-truth-input errors of 1.975/2.036 cm for
left/right elbows and 3.221/2.153 cm for left/right knees.
$\mathrm{ID}_{\mathrm{cond}}$ and $\mathrm{OOD}_{\mathrm{cond}}$ share these
references because they show the same windows from different cameras.

\FloatBarrier
\section{Multi-Domain Training}
\label{app:mix_details}
\setcounter{table}{0}
\setcounter{figure}{0}

Multi-domain training shares one online encoder, EMA target encoder, and JEPA predictor
across domains.
Each domain retains its own variables, train-only normalization statistics,
pointwise auxiliary head, loss weight, validity mask, and warmup counter.
The video-fine-tuned control uses the same synchronized multi-domain schedule
and optimizes $\mathcal L_{\mathrm{JEPA}}$.

\subsection{Multi-Domain Training Schedule}

After distributed sharding, each domain contributes the same number of complete batches as in its corresponding single-domain loader.
All ranks select the same domain and its auxiliary head at each optimizer step.
Table~\ref{tab:app_mix_exposure} gives the exact accounting for eight GPUs and
batch size eight per GPU.

\begin{table}[!htbp]
    \caption{\textbf{Per-domain schedule quotas.}
    Each entry reports synchronized optimizer steps per epoch for the
    corresponding domain loader.}
    \label{tab:app_mix_exposure}
    \centering
    \fontsize{8.5}{10.5}\selectfont
    \setlength{\tabcolsep}{5pt}
    \begin{tabularx}{\textwidth}{@{}lXX@{}}
        \toprule
        Domain & Reference loader & Shared schedule \\
        \midrule
        autonomous driving & 98 steps / epoch & 98 steps / epoch \\
        robot manipulation & 234 steps / epoch & 234 steps / epoch \\
        human motion & 109 steps / epoch & 109 steps / epoch \\
        \bottomrule
    \end{tabularx}
\end{table}

\subsection{Matched Probes and Multi-Domain Results}

Within each domain and panel, we hold the split, variable definitions, probe
architecture, optimizer, seed, normalization, and aggregation fixed. We change
only the frozen encoder.

Table~\ref{tab:mix_primary} gives the exact values underlying the main-text
comparison in Figure~\ref{fig:multidomain_gain}.

\begin{table}[!htbp]
    \caption{\textbf{Single-domain and multi-domain Time-JEPA.}
    Single and Multi denote the corresponding Time-JEPA encoders.
    We compute changes before rounding and retain the matched supervised
    probes and variable/view aggregation from Table~\ref{tab:single_domain_main}. Bold marks the higher score.}
    \label{tab:mix_primary}
    \centering
    \fontsize{8.5}{10.5}\selectfont
    \setlength{\tabcolsep}{2pt}
    \renewcommand{\arraystretch}{1.06}
    \begin{tabular*}{\textwidth}{@{\extracolsep{\fill}}cccccc@{\hspace{8pt}}ccc@{\hspace{8pt}}ccc@{}}
        \toprule
         &  &  & \multicolumn{3}{c}{\textbf{autonomous driving}} & \multicolumn{3}{c}{\textbf{robot manipulation}} & \multicolumn{3}{c}{\textbf{human motion}} \\
        \cmidrule(lr){4-6}\cmidrule(lr){7-9}\cmidrule(lr){10-12}
        Condition & Variables & Metric & Single & Multi & $\Delta$ & Single & Multi & $\Delta$ & Single & Multi & $\Delta$ \\
        \midrule
        \multirow{4}{*}{\textbf{$\mathrm{ID}_{\mathrm{cond}}$}} & \multirow{2}{*}{$\mathrm{ID}_{\mathrm{var}}$} & \appmetric{r} & 0.916 & \textbf{0.988} & +0.072 & 0.845 & \textbf{0.912} & +0.066 & 0.589 & \textbf{0.698} & +0.109 \\
        & & \appmetric{R^2} & 0.849 & \textbf{0.976} & +0.127 & 0.725 & \textbf{0.840} & +0.115 & 0.345 & \textbf{0.506} & +0.161 \\
        & \multirow{2}{*}{$\mathrm{OOD}_{\mathrm{var}}$} & \appmetric{r} & 0.920 & \textbf{0.943} & +0.023 & 0.741 & \textbf{0.781} & +0.040 & 0.504 & \textbf{0.669} & +0.165 \\
        & & \appmetric{R^2} & 0.846 & \textbf{0.890} & +0.044 & 0.568 & \textbf{0.631} & +0.062 & 0.222 & \textbf{0.456} & +0.235 \\
        \midrule
        \multirow{4}{*}{\textbf{$\mathrm{OOD}_{\mathrm{cond}}$}} & \multirow{2}{*}{$\mathrm{ID}_{\mathrm{var}}$} & \appmetric{r} & 0.902 & \textbf{0.985} & +0.083 & 0.492 & \textbf{0.583} & +0.091 & 0.581 & \textbf{0.702} & +0.121 \\
        & & \appmetric{R^2} & 0.824 & \textbf{0.970} & +0.146 & 0.085 & \textbf{0.139} & +0.054 & 0.332 & \textbf{0.513} & +0.181 \\
        & \multirow{2}{*}{$\mathrm{OOD}_{\mathrm{var}}$} & \appmetric{r} & 0.937 & \textbf{0.961} & +0.025 & 0.555 & \textbf{0.609} & +0.054 & 0.515 & \textbf{0.672} & +0.156 \\
        & & \appmetric{R^2} & 0.876 & \textbf{0.923} & +0.047 & 0.205 & \textbf{0.267} & +0.063 & 0.225 & \textbf{0.456} & +0.231 \\
        \bottomrule
    \end{tabular*}
\end{table}

\begin{figure}[htbp]
    \centering
    \includegraphics[width=\textwidth]{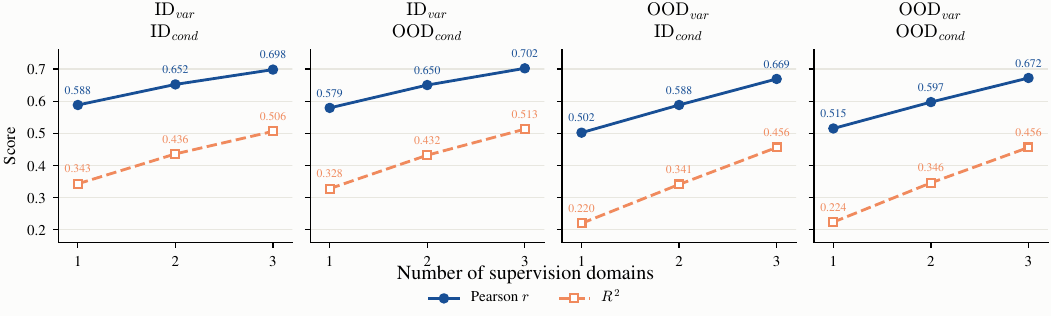}
    \caption{\textbf{Paired domain-addition trend in human motion.}
    Connected three-point plots trace the paired sequence trained with human
    motion only, autonomous driving + human motion, and all three domains.
    Dark blue reports Pearson $r$ and orange reports $R^2$.
    Point labels give the exact scores for all four variable--condition
    panels.}
    \label{fig:domain_addition}
\end{figure}

Across all four human motion panels, Pearson $r$ and $R^2$ increase from
human motion only to autonomous driving + human motion and again to multi-domain
Time-JEPA. For $\mathrm{OOD}_{\mathrm{var}}$ under
$\mathrm{OOD}_{\mathrm{cond}}$, Pearson follows
$0.515\!\rightarrow\!0.597\!\rightarrow\!0.672$, and $R^2$ follows
$0.224\!\rightarrow\!0.346\!\rightarrow\!0.456$. The other three panels show
the same monotonic direction in Figure~\ref{fig:domain_addition}.

\FloatBarrier

\subsection{Single- vs. Multi-Domain Time-JEPA for Future-State Prediction}

We compare single-domain and multi-domain Time-JEPA using the five-horizon
probes from Appendix~\ref{app:future_state}.
We keep the splits, variables, probe architecture, optimization, normalization,
and probe seed fixed and change the frozen encoder.
Table~\ref{tab:mix_future} reports all domain--variable--condition panels.

\begin{table}[!htbp]
    \caption{\textbf{Single-domain and multi-domain future-state prediction.}
    Single and Multi denote Time-JEPA encoders. We average the five horizons
    from 0.2 to 1.0 s equally and use the variable/view aggregation from
    Appendix~\ref{app:future_state}. Each encoder uses one probe run per
    variable group. Bold marks the better unrounded score.}
    \label{tab:mix_future}
    \centering
    \fontsize{8.5}{10.5}\selectfont
    \setlength{\tabcolsep}{3pt}
    \renewcommand{\arraystretch}{1.06}
    \begin{tabular*}{\textwidth}{@{\extracolsep{\fill}}ccccc@{\hspace{8pt}}cc@{\hspace{8pt}}cc@{}}
        \toprule
         & & & \multicolumn{2}{c}{\textbf{autonomous driving}}
         & \multicolumn{2}{c}{\textbf{robot manipulation}}
         & \multicolumn{2}{c}{\textbf{human motion}} \\
        \cmidrule(lr){4-5}\cmidrule(lr){6-7}\cmidrule(lr){8-9}
        Condition & Variables & Metric & Single & Multi & Single & Multi & Single & Multi \\
        \midrule
        \multirow{4}{*}{\textbf{$\mathrm{ID}_{\mathrm{cond}}$}} & \multirow{2}{*}{$\mathrm{ID}_{\mathrm{var}}$} & nRMSE $\downarrow$
        & 0.485 & \textbf{0.349} & 0.647 & \textbf{0.570} & 0.999 & \textbf{0.913} \\
        & & $R^2\uparrow$
        & 0.719 & \textbf{0.851} & 0.506 & \textbf{0.587} & 0.244 & \textbf{0.375} \\
        & \multirow{2}{*}{$\mathrm{OOD}_{\mathrm{var}}$} & nRMSE $\downarrow$
        & 0.560 & \textbf{0.488} & 0.771 & \textbf{0.722} & 0.978 & \textbf{0.894} \\
        & & $R^2\uparrow$
        & 0.667 & \textbf{0.748} & 0.379 & \textbf{0.444} & 0.190 & \textbf{0.321} \\
        \midrule
        \multirow{4}{*}{\textbf{$\mathrm{OOD}_{\mathrm{cond}}$}} & \multirow{2}{*}{$\mathrm{ID}_{\mathrm{var}}$} & nRMSE $\downarrow$
        & 0.460 & \textbf{0.333} & 0.960 & \textbf{0.952} & 1.010 & \textbf{0.902} \\
        & & $R^2\uparrow$
        & 0.680 & \textbf{0.812} & \textbf{0.008} & 0.006 & 0.226 & \textbf{0.391} \\
        & \multirow{2}{*}{$\mathrm{OOD}_{\mathrm{var}}$} & nRMSE $\downarrow$
        & 0.529 & \textbf{0.449} & \textbf{0.951} & 0.964 & 0.975 & \textbf{0.888} \\
        & & $R^2\uparrow$
        & 0.604 & \textbf{0.711} & \textbf{0.047} & -0.003 & 0.195 & \textbf{0.330} \\
        \bottomrule
    \end{tabular*}
\end{table}

Multi-domain training improves both metrics in all six
$\mathrm{ID}_{\mathrm{cond}}$ panels and in the autonomous driving and human
motion $\mathrm{OOD}_{\mathrm{cond}}$ panels.
For robot manipulation under $\mathrm{OOD}_{\mathrm{cond}}$, it lowers
$\mathrm{ID}_{\mathrm{var}}$ nRMSE but also lowers $R^2$, and the single-domain
encoder performs better on both $\mathrm{OOD}_{\mathrm{var}}$ metrics.
The corresponding horizon-macro values are $0.960\!\rightarrow\!0.952$ and
$0.008\!\rightarrow\!0.006$ for nRMSE and $R^2$ on $\mathrm{ID}_{\mathrm{var}}$,
and $0.951\!\rightarrow\!0.964$ and $0.047\!\rightarrow\!-0.003$ on
$\mathrm{OOD}_{\mathrm{var}}$.
Across the 24 displayed nRMSE and $R^2$ entries, multi-domain training is
better in 21; the three reversals are these robot-manipulation OOD entries.
Using the variable-count weights from Table~\ref{tab:future_aggregate},
multi-domain training improves both aggregate metrics for both variable groups
under both condition settings.

\subsection{Single- vs. Multi-Domain Time-JEPA for Kinematic Reconstruction}

We apply the reconstruction rules from Appendix~\ref{app:grounded_details} to
matched probes on the single-domain and multi-domain Time-JEPA encoders.
We keep the probe settings, evaluation windows, physical relations, and
aggregation fixed across each pair.
Multi-domain Time-JEPA lowers the primary error in all six panels
(Table~\ref{tab:mix_reconstruction}).
These reconstruction gains and the aggregate prediction gains support our use
of the multi-domain encoder for further physical evaluations.

\begin{table}[!htbp]
    \caption{\textbf{Single-domain and multi-domain kinematic reconstruction.}
    Single and Multi denote Time-JEPA encoders. Entries give mean $\pm$ sample
    standard deviation across five matched probe seeds. We average trajectory
    error over three horizons, reach-rate error over both arms, and distance
    error over four limbs within each seed. Bold marks the lower mean.}
    \label{tab:mix_reconstruction}
    \centering
    \fontsize{8.5}{10.5}\selectfont
    \setlength{\tabcolsep}{4pt}
    \renewcommand{\arraystretch}{1.06}
    \begin{tabular*}{\textwidth}{@{\extracolsep{\fill}}ccccc@{}}
        \toprule
        Condition & Domain & Metric $\downarrow$ & Single & Multi \\
        \midrule
        \multirow{3}{*}{$\mathrm{ID}_{\mathrm{cond}}$} & autonomous driving & Trajectory MAE (m)
        & 3.080\(\,\pm\,\)0.114 & \textbf{1.795\(\,\pm\,\)0.027} \\
        & robot manipulation & Reach-rate MAE (cm/s)
        & 3.454\(\,\pm\,\)0.016 & \textbf{2.290\(\,\pm\,\)0.006} \\
        & human motion & Distance MAE (cm)
        & 6.230\(\,\pm\,\)0.080 & \textbf{4.619\(\,\pm\,\)0.065} \\
        \midrule
        \multirow{3}{*}{$\mathrm{OOD}_{\mathrm{cond}}$} & autonomous driving & Trajectory MAE (m)
        & 3.572\(\,\pm\,\)0.312 & \textbf{1.858\(\,\pm\,\)0.144} \\
        & robot manipulation & Reach-rate MAE (cm/s)
        & 4.710\(\,\pm\,\)0.025 & \textbf{4.171\(\,\pm\,\)0.046} \\
        & human motion & Distance MAE (cm)
        & 6.423\(\,\pm\,\)0.129 & \textbf{4.557\(\,\pm\,\)0.053} \\
        \bottomrule
    \end{tabular*}
\end{table}

\FloatBarrier
\section{Future-State Prediction}
\label{app:future_state}
\setcounter{table}{0}
\setcounter{figure}{0}

\subsection{Five-Horizon Protocol}

Let $t$ be the final observed timestamp of the 64-frame, 5-FPS clip.
The probe predicts one state at each $t+h$ for
$\mathcal H=\{0.2,0.4,0.6,0.8,1.0\}$~s.
The frozen encoder processes the observed RGB frames. The probe
spatially averages each temporal feature, selects the final temporal feature,
and applies a two-layer pointwise nonlinear projection to predict all five
future states jointly.

For each of the three encoder configurations, we train a separate five-horizon
probe for 15 epochs using future-state labels from the probe-training split.
We estimate normalization statistics separately for
each variable and horizon on that split. The regression loss averages valid
state values within each horizon, then weights the five horizons equally.

$\mathrm{ID}_{\mathrm{var}}$ variables enter the encoder's time-series
supervision; $\mathrm{OOD}_{\mathrm{var}}$ variables enter only supervised
probe training. We compare original V-JEPA, the multi-domain video-fine-tuned
control, and multi-domain Time-JEPA on the same future-label splits. Metrics
are computed per variable and horizon on valid future labels, then averaged
equally across variables and horizons.

\subsection{In-Distribution Results}

Table~\ref{tab:future_state} breaks down the five-horizon
$\mathrm{ID}_{\mathrm{cond}}$ results by domain and variable group.
Table~\ref{tab:app_future_horizons} resolves the horizon trend, and
Tables~\ref{tab:app_future_covla}, \ref{tab:app_future_egoexo},
and~\ref{tab:app_future_agibot} list the individual variables. Time-JEPA has the lowest nRMSE and the highest $R^2$
in all six domain--variable-group rows; its advantage over Video FT holds at
every horizon in the three-domain aggregate.

\begin{table}[!htbp]
    \caption{\textbf{Five-horizon future-state prediction under
    $\mathrm{ID}_{\mathrm{cond}}$.} O/V/T denote original V-JEPA, the video-fine-tuned
    control, and Time-JEPA. Scores average variables and horizons equally
    within each domain and group. The final columns count horizon--variable
    cells with lower Time-JEPA RMSE than V/O (five per dimension). Each
    configuration and group uses one supervised joint state probe.}
    \label{tab:future_state}
    \centering
    \fontsize{8.5}{10.5}\selectfont
    \setlength{\tabcolsep}{3.4pt}
    \begin{tabular}{@{}ccc*{6}{c}cc@{}}
        \toprule
        & & & \multicolumn{3}{c}{nRMSE $\downarrow$}
        & \multicolumn{3}{c}{$R^2$ $\uparrow$}
        & \multicolumn{2}{c}{Time-JEPA RMSE wins} \\
        \cmidrule(lr){4-6}\cmidrule(lr){7-9}\cmidrule(l){10-11}
        Domain & Variables & Dims
        & O & V & T & O & V & T
        & vs. V & vs. O \\
        \midrule
        autonomous driving & $\mathrm{ID}_{\mathrm{var}}$ & 3
        & 0.580 & 0.541 & \textbf{0.349}
        & 0.630 & 0.668 & \textbf{0.851} & 15/15 & 15/15 \\
        & $\mathrm{OOD}_{\mathrm{var}}$ & 3
        & 0.640 & 0.598 & \textbf{0.488}
        & 0.575 & 0.625 & \textbf{0.748} & 15/15 & 15/15 \\
        \addlinespace[2pt]
        robot manipulation & $\mathrm{ID}_{\mathrm{var}}$ & 18
        & 0.821 & 0.806 & \textbf{0.570}
        & 0.260 & 0.284 & \textbf{0.587} & 90/90 & 90/90 \\
        & $\mathrm{OOD}_{\mathrm{var}}$ & 8
        & 0.876 & 0.865 & \textbf{0.722}
        & 0.209 & 0.228 & \textbf{0.444} & 40/40 & 40/40 \\
        \addlinespace[2pt]
        human motion & $\mathrm{ID}_{\mathrm{var}}$ & 9
        & 1.022 & 1.027 & \textbf{0.913}
        & 0.204 & 0.198 & \textbf{0.375} & 44/45 & 43/45 \\
        & $\mathrm{OOD}_{\mathrm{var}}$ & 5
        & 1.010 & 1.011 & \textbf{0.894}
        & 0.133 & 0.131 & \textbf{0.321} & 25/25 & 25/25 \\
        \bottomrule
    \end{tabular}
\end{table}

\begin{table}[!htbp]
    \caption{\textbf{Three-domain $\mathrm{ID}_{\mathrm{cond}}$ performance by future horizon.}
    Domains are weighted by variable count. Final columns count variables
    with lower Time-JEPA RMSE than Video FT/Original; bold marks the best score.}
    \label{tab:app_future_horizons}
    \centering
    \fontsize{8.5}{10.5}\selectfont
    \setlength{\tabcolsep}{2.2pt}
    \renewcommand{\arraystretch}{1.06}
    \begin{tabular*}{\textwidth}{@{\extracolsep{\fill}}ccccccccc@{}}
        \toprule
         & \multicolumn{3}{c}{nRMSE $\downarrow$} & \multicolumn{3}{c}{$R^2\uparrow$} & \multicolumn{2}{c}{Time-JEPA RMSE wins} \\
        \cmidrule(lr){2-4}\cmidrule(lr){5-7}\cmidrule(l){8-9}
        Horizon & Original & Video FT & Time-JEPA & Original & Video FT & Time-JEPA & vs. Video FT & vs. Original \\
        \midrule
        0.2 s & 0.832 & 0.818 & \textbf{0.612} & 0.293 & 0.312 & \textbf{0.582} & 46/46 & 45/46 \\
        0.4 s & 0.876 & 0.861 & \textbf{0.679} & 0.282 & 0.304 & \textbf{0.544} & 46/46 & 46/46 \\
        0.6 s & 0.851 & 0.839 & \textbf{0.669} & 0.271 & 0.288 & \textbf{0.517} & 46/46 & 46/46 \\
        0.8 s & 0.850 & 0.840 & \textbf{0.682} & 0.262 & 0.278 & \textbf{0.495} & 46/46 & 46/46 \\
        1.0 s & 0.905 & 0.896 & \textbf{0.752} & 0.247 & 0.259 & \textbf{0.458} & 45/46 & 45/46 \\
        \bottomrule
    \end{tabular*}
\end{table}

\begin{table}[!htbp]
    \caption{\textbf{autonomous driving: five-horizon future-state prediction by variable.}
    O/V/T denote Original/Video FT/Time-JEPA; scores average five horizons
    under $\mathrm{ID}_{\mathrm{cond}}$. Bold marks the best configuration.}
    \label{tab:app_future_covla}
    \centering
    \fontsize{8.5}{10.5}\selectfont
    \setlength{\tabcolsep}{2.2pt}
    \renewcommand{\arraystretch}{1.06}
    \begin{tabular*}{\textwidth}{@{\extracolsep{\fill}}c>{\centering\arraybackslash}p{0.30\textwidth}cccccc@{}}
        \toprule
         &  & \multicolumn{3}{c}{nRMSE $\downarrow$} & \multicolumn{3}{c}{$R^2\uparrow$} \\
        \cmidrule(lr){3-5}\cmidrule(lr){6-8}
        Group & Variable & O & V & T & O & V & T \\
        \midrule
        $\mathrm{ID}_{\mathrm{var}}$ & Longitudinal velocity & 0.280 & 0.245 & \textbf{0.110} & 0.899 & 0.923 & \textbf{0.984} \\
        $\mathrm{ID}_{\mathrm{var}}$ & Longitudinal acceleration & 0.917 & 0.894 & \textbf{0.601} & 0.278 & 0.313 & \textbf{0.685} \\
        $\mathrm{ID}_{\mathrm{var}}$ & Yaw rate & 0.541 & 0.483 & \textbf{0.335} & 0.714 & 0.769 & \textbf{0.882} \\
        \midrule
        $\mathrm{OOD}_{\mathrm{var}}$ & Steering angle & 0.685 & 0.638 & \textbf{0.522} & 0.590 & 0.643 & \textbf{0.760} \\
        $\mathrm{OOD}_{\mathrm{var}}$ & Lateral velocity & 0.604 & 0.550 & \textbf{0.435} & 0.656 & 0.713 & \textbf{0.818} \\
        $\mathrm{OOD}_{\mathrm{var}}$ & Lateral acceleration & 0.632 & 0.607 & \textbf{0.506} & 0.479 & 0.518 & \textbf{0.664} \\
        \bottomrule
    \end{tabular*}
\end{table}

\begin{table}[!htbp]
    \caption{\textbf{human motion: five-horizon future-state prediction by variable.}
    O/V/T denote Original/Video FT/Time-JEPA. We average metrics equally over
    five horizons under $\mathrm{ID}_{\mathrm{cond}}$.
    Bold marks the best configuration.}
    \label{tab:app_future_egoexo}
    \centering
    \fontsize{8.5}{10.5}\selectfont
    \setlength{\tabcolsep}{2.2pt}
    \renewcommand{\arraystretch}{1.06}
    \begin{tabular*}{\textwidth}{@{\extracolsep{\fill}}c>{\centering\arraybackslash}p{0.30\textwidth}cccccc@{}}
        \toprule
         &  & \multicolumn{3}{c}{nRMSE $\downarrow$} & \multicolumn{3}{c}{$R^2\uparrow$} \\
        \cmidrule(lr){3-5}\cmidrule(lr){6-8}
        Group & Variable & O & V & T & O & V & T \\
        \midrule
        $\mathrm{ID}_{\mathrm{var}}$ & Pelvis speed & 0.871 & 0.891 & \textbf{0.794} & 0.302 & 0.270 & \textbf{0.421} \\
        $\mathrm{ID}_{\mathrm{var}}$ & Left wrist relative speed & 1.008 & 1.012 & \textbf{1.003} & 0.214 & 0.207 & \textbf{0.223} \\
        $\mathrm{ID}_{\mathrm{var}}$ & Right wrist relative speed & 0.819 & 0.820 & \textbf{0.805} & 0.210 & 0.207 & \textbf{0.234} \\
        $\mathrm{ID}_{\mathrm{var}}$ & Left ankle relative speed & 1.844 & 1.853 & \textbf{1.821} & 0.092 & 0.082 & \textbf{0.121} \\
        $\mathrm{ID}_{\mathrm{var}}$ & Right ankle relative speed & 1.188 & 1.209 & \textbf{1.141} & 0.176 & 0.146 & \textbf{0.239} \\
        $\mathrm{ID}_{\mathrm{var}}$ & Left elbow flexion & 0.970 & 0.951 & \textbf{0.788} & 0.077 & 0.112 & \textbf{0.389} \\
        $\mathrm{ID}_{\mathrm{var}}$ & Right elbow flexion & 0.897 & 0.876 & \textbf{0.715} & 0.157 & 0.196 & \textbf{0.464} \\
        $\mathrm{ID}_{\mathrm{var}}$ & Left knee flexion & 0.792 & 0.817 & \textbf{0.587} & 0.394 & 0.355 & \textbf{0.667} \\
        $\mathrm{ID}_{\mathrm{var}}$ & Right knee flexion & 0.811 & 0.817 & \textbf{0.568} & 0.212 & 0.202 & \textbf{0.615} \\
        \midrule
        $\mathrm{OOD}_{\mathrm{var}}$ & Inter-wrist distance & 0.909 & 0.894 & \textbf{0.853} & 0.283 & 0.305 & \textbf{0.367} \\
        $\mathrm{OOD}_{\mathrm{var}}$ & Left wrist--shoulder distance & 1.018 & 1.033 & \textbf{0.894} & 0.066 & 0.038 & \textbf{0.282} \\
        $\mathrm{OOD}_{\mathrm{var}}$ & Right wrist--shoulder distance & 0.910 & 0.910 & \textbf{0.776} & 0.152 & 0.152 & \textbf{0.383} \\
        $\mathrm{OOD}_{\mathrm{var}}$ & Inter-ankle distance & 1.258 & 1.284 & \textbf{1.230} & 0.006 & \(-\)0.035 & \textbf{0.050} \\
        $\mathrm{OOD}_{\mathrm{var}}$ & Pelvis support height & 0.956 & 0.936 & \textbf{0.718} & 0.160 & 0.194 & \textbf{0.525} \\
        \bottomrule
    \end{tabular*}
\end{table}

\begin{table}[!htbp]
    \caption{\textbf{robot manipulation: five-horizon future-state prediction by variable.}
    O/V/T denote Original/Video FT/Time-JEPA; scores average five horizons
    under $\mathrm{ID}_{\mathrm{cond}}$. Bold marks the best configuration.}
    \label{tab:app_future_agibot}
    \centering
    \fontsize{8.5}{10.5}\selectfont
    \setlength{\tabcolsep}{2.2pt}
    \renewcommand{\arraystretch}{1.00}
    \begin{tabular*}{\textwidth}{@{\extracolsep{\fill}}c>{\centering\arraybackslash}p{0.30\textwidth}cccccc@{}}
        \toprule
         &  & \multicolumn{3}{c}{nRMSE $\downarrow$} & \multicolumn{3}{c}{$R^2\uparrow$} \\
        \cmidrule(lr){3-5}\cmidrule(lr){6-8}
        Group & Variable & O & V & T & O & V & T \\
        \midrule
        $\mathrm{ID}_{\mathrm{var}}$ & Base translation speed & 0.886 & 0.881 & \textbf{0.849} & 0.108 & 0.117 & \textbf{0.181} \\
        $\mathrm{ID}_{\mathrm{var}}$ & Base angular speed & 0.754 & 0.757 & \textbf{0.722} & 0.104 & 0.096 & \textbf{0.177} \\
        $\mathrm{ID}_{\mathrm{var}}$ & Waist PC1 & 0.557 & 0.532 & \textbf{0.272} & 0.700 & 0.726 & \textbf{0.928} \\
        $\mathrm{ID}_{\mathrm{var}}$ & Waist PC2 & 0.574 & 0.569 & \textbf{0.402} & 0.697 & 0.703 & \textbf{0.851} \\
        $\mathrm{ID}_{\mathrm{var}}$ & Left arm position $x$ & 0.803 & 0.796 & \textbf{0.346} & 0.253 & 0.266 & \textbf{0.860} \\
        $\mathrm{ID}_{\mathrm{var}}$ & Left arm position $y$ & 0.630 & 0.616 & \textbf{0.263} & 0.553 & 0.572 & \textbf{0.921} \\
        $\mathrm{ID}_{\mathrm{var}}$ & Left arm position $z$ & 0.720 & 0.701 & \textbf{0.298} & 0.377 & 0.411 & \textbf{0.891} \\
        $\mathrm{ID}_{\mathrm{var}}$ & Right arm position $x$ & 0.801 & 0.761 & \textbf{0.362} & 0.384 & 0.444 & \textbf{0.872} \\
        $\mathrm{ID}_{\mathrm{var}}$ & Right arm position $y$ & 0.729 & 0.697 & \textbf{0.335} & 0.466 & 0.511 & \textbf{0.885} \\
        $\mathrm{ID}_{\mathrm{var}}$ & Right arm position $z$ & 0.740 & 0.703 & \textbf{0.300} & 0.390 & 0.449 & \textbf{0.896} \\
        $\mathrm{ID}_{\mathrm{var}}$ & Left end velocity $x$ & 0.994 & 0.982 & \textbf{0.820} & \(-\)0.014 & 0.011 & \textbf{0.294} \\
        $\mathrm{ID}_{\mathrm{var}}$ & Left end velocity $y$ & 0.926 & 0.924 & \textbf{0.796} & 0.010 & 0.016 & \textbf{0.266} \\
        $\mathrm{ID}_{\mathrm{var}}$ & Left end velocity $z$ & 1.032 & 1.025 & \textbf{0.925} & \(-\)0.027 & \(-\)0.012 & \textbf{0.178} \\
        $\mathrm{ID}_{\mathrm{var}}$ & Right end velocity $x$ & 0.939 & 0.927 & \textbf{0.748} & 0.000 & 0.023 & \textbf{0.355} \\
        $\mathrm{ID}_{\mathrm{var}}$ & Right end velocity $y$ & 1.027 & 1.014 & \textbf{0.827} & 0.020 & 0.045 & \textbf{0.364} \\
        $\mathrm{ID}_{\mathrm{var}}$ & Right end velocity $z$ & 1.055 & 1.046 & \textbf{0.936} & 0.002 & 0.019 & \textbf{0.221} \\
        $\mathrm{ID}_{\mathrm{var}}$ & Left gripper state & 0.778 & 0.761 & \textbf{0.523} & 0.394 & 0.421 & \textbf{0.726} \\
        $\mathrm{ID}_{\mathrm{var}}$ & Right gripper state & 0.830 & 0.812 & \textbf{0.529} & 0.258 & 0.291 & \textbf{0.698} \\
        \midrule
        $\mathrm{OOD}_{\mathrm{var}}$ & Left joint-position norm & 0.660 & 0.663 & \textbf{0.435} & 0.471 & 0.466 & \textbf{0.770} \\
        $\mathrm{OOD}_{\mathrm{var}}$ & Right joint-position norm & 0.832 & 0.805 & \textbf{0.505} & 0.317 & 0.361 & \textbf{0.748} \\
        $\mathrm{OOD}_{\mathrm{var}}$ & Left joint-velocity norm & 0.911 & 0.890 & \textbf{0.744} & 0.149 & 0.188 & \textbf{0.427} \\
        $\mathrm{OOD}_{\mathrm{var}}$ & Right joint-velocity norm & 0.906 & 0.893 & \textbf{0.750} & 0.171 & 0.196 & \textbf{0.429} \\
        $\mathrm{OOD}_{\mathrm{var}}$ & Left joint-effort norm & 0.765 & 0.760 & \textbf{0.641} & 0.217 & 0.228 & \textbf{0.449} \\
        $\mathrm{OOD}_{\mathrm{var}}$ & Right joint-effort norm & 1.017 & 1.005 & \textbf{0.861} & 0.177 & 0.196 & \textbf{0.411} \\
        $\mathrm{OOD}_{\mathrm{var}}$ & Left wrench-force norm & 0.756 & 0.749 & \textbf{0.713} & 0.027 & 0.044 & \textbf{0.133} \\
        $\mathrm{OOD}_{\mathrm{var}}$ & Right wrench-force norm & 1.157 & 1.154 & \textbf{1.127} & 0.143 & 0.147 & \textbf{0.187} \\
        \bottomrule
    \end{tabular*}
\end{table}

\FloatBarrier
\subsection{Held-Out Conditions}

Table~\ref{tab:app_future_ood} evaluates the same fitted probes under
$\mathrm{OOD}_{\mathrm{cond}}$: stable rain for autonomous driving, both
fisheye views for robot manipulation, and exocentric \texttt{view03} for human
motion.

\begin{table}[htbp]
    \caption{\textbf{Multi-horizon future-state prediction under
    $\mathrm{OOD}_{\mathrm{cond}}$ by domain and variable group.}
    We average metrics equally over the five horizons.
Bold marks the best score for each domain, variable group, and metric.}
    \label{tab:app_future_ood}
    \centering
    \small
    \setlength{\tabcolsep}{4pt}
    \begin{tabular}{@{}cccccccc@{}}
        \toprule
        & & \multicolumn{3}{c}{nRMSE $\downarrow$}
          & \multicolumn{3}{c}{$R^2$ $\uparrow$} \\
        \cmidrule(lr){3-5}\cmidrule(l){6-8}
        Domain & Variables & Original & Video FT & Time-JEPA
                         & Original & Video FT & Time-JEPA \\
        \midrule
        autonomous driving & $\mathrm{ID}_{\mathrm{var}}$ & 0.548 & 0.515 & \textbf{0.333}
                                & 0.580 & 0.620 & \textbf{0.812} \\
                           & $\mathrm{OOD}_{\mathrm{var}}$ & 0.590 & 0.570 & \textbf{0.449}
                                & 0.512 & 0.543 & \textbf{0.711} \\
        \addlinespace[2pt]
        robot manipulation & $\mathrm{ID}_{\mathrm{var}}$ & 0.975 & 0.973 & \textbf{0.952}
                                & $-0.012$ & $-0.010$ & \textbf{0.006} \\
                           & $\mathrm{OOD}_{\mathrm{var}}$ & 1.007 & 0.980 & \textbf{0.964}
                                & $-0.042$ & \textbf{0.013} & $-0.003$ \\
        \addlinespace[2pt]
        human motion       & $\mathrm{ID}_{\mathrm{var}}$ & 1.045 & 1.047 & \textbf{0.902}
                                & 0.165 & 0.162 & \textbf{0.391} \\
                           & $\mathrm{OOD}_{\mathrm{var}}$ & 1.010 & 1.006 & \textbf{0.888}
                                & 0.134 & 0.142 & \textbf{0.330} \\
        \bottomrule
    \end{tabular}
\end{table}

Across the six domain--variable-group settings under
$\mathrm{OOD}_{\mathrm{cond}}$, Time-JEPA improves 11 of 12 reported metric
comparisons over Video FT: nRMSE decreases in all six settings, and $R^2$
increases in five.

\subsection{Long-Horizon Persistence Extension}

\begin{figure}[!htbp]
    \centering
    \includegraphics[width=0.9\textwidth]{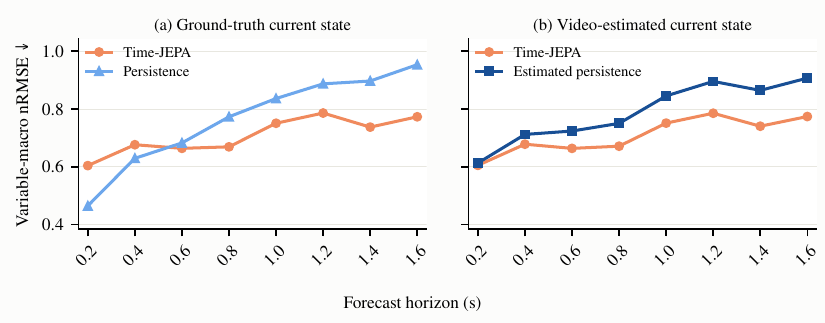}
    \caption{\textbf{Long-horizon persistence extension under
    $\mathrm{ID}_{\mathrm{cond}}$.}
    Panels (a) and (b) compare the eight-horizon Time-JEPA probe with
    persistence and estimated persistence, respectively. Curves give
    variable-macro nRMSE over 30 $\mathrm{ID}_{\mathrm{var}}$ and 16
    $\mathrm{OOD}_{\mathrm{var}}$ variables. Panel (a) uses valid current and
    future labels; panel (b) uses valid future labels.}
    \label{fig:future_state}
\end{figure}

We separately train six eight-horizon probes on frozen multi-domain Time-JEPA
features, one per domain and variable group, for
$t+\{0.2,0.4,\ldots,1.6\}$~s. Figure~\ref{fig:future_state} compares them
with \emph{persistence}, which holds the ground-truth state at $t$ fixed, and
\emph{estimated persistence}, which holds a separate video-based current-state
probe's estimate fixed.

The ground-truth comparison requires valid current and future labels; we
recompute the future-probe error on those same samples. The estimated
comparison requires only valid future labels. Each horizon reports nRMSE
averaged equally over 46 variables from $\mathrm{ID}_{\mathrm{cond}}$ test views.

Time-JEPA has lower variable-macro nRMSE than persistence at 0.6--1.6~s,
while persistence has lower error at 0.2--0.4~s. Time-JEPA has lower error
than estimated persistence at all eight horizons (Figure~\ref{fig:future_state}).
Together, the two comparisons suggest that this advantage reflects prediction
of future physical-state evolution rather than only current-state estimation:
the ground-truth-current baseline supplies the exact state at $t$, whereas
estimated persistence holds a video-based current-state estimate fixed, and
Time-JEPA outperforms them from 0.6--1.6~s and at all eight horizons,
respectively.

\FloatBarrier
\section{External Physical Evaluations}
\label{app:external_transfer}
\setcounter{table}{0}

We compare original V-JEPA, the multi-domain video-fine-tuned control, and
multi-domain Time-JEPA. GRASP directly evaluates the frozen online encoder,
target encoder, and predictor. Physion trains a supervised logistic probe on
frozen target-encoder features.

\subsection{GRASP Violation-of-Expectation Protocol}
\label{app:grasp}

GRASP Level 2 contains 16 test families with 128 matched
possible/impossible pairs per family, for 2,048 pairs in total
\citep{jassim2024grasp}. We sample the 50-FPS videos every tenth frame,
process 16-frame windows with stride 2, and apply the same $C=2$, $M=14$
prediction mask to every configuration. For each window, we compute the latent
$L_1$ prediction error between normalized predicted and target hierarchical
features, then average the window scores over the video. The higher-surprise video receives the
impossible label, and ties count as errors. The primary metric averages pair
accuracy equally over the eight non-empty intuitive-physics properties. A
multi-label family contributes to each property listed for that family.

\begin{table}[htbp]
    \caption{\textbf{GRASP eight-property macro pair accuracy.}
    We score all 2,048 Level~2 pairs with the fixed $C=2$, $M=14$ protocol.
    The primary metric averages the eight property accuracies equally.
    Differences use percentage points relative to Video FT. Bold marks the
    highest accuracy.}
    \label{tab:app_grasp_summary}
    \centering
    \fontsize{8.5}{10.5}\selectfont
    \setlength{\tabcolsep}{7pt}
    \begin{tabular}{@{}lcc@{}}
        \toprule
        Configuration & Pair accuracy (\%) $\uparrow$
        & $\Delta$ vs. Video FT (pp) \\
        \midrule
        Original V-JEPA & 50.658 & $-4.902$ \\
        Video FT & 55.560 & {--} \\
        Time-JEPA  & \textbf{57.751} & $\mathbf{+2.191}$ \\
        \bottomrule
    \end{tabular}
\end{table}

\begin{table}[htbp]
    \caption{\textbf{GRASP pair accuracy by intuitive-physics property.}
    Each row uses the matched pairs assigned to that property. Differences use
    percentage points relative to Video FT. Bold marks the highest encoder point
    estimate in each row.}
    \label{tab:app_grasp_property}
    \centering
    \fontsize{8.5}{10.5}\selectfont
    \setlength{\tabcolsep}{4.5pt}
    \begin{tabular}{@{}lrrrrr@{}}
        \toprule
        Property & Pairs & Original & Video FT & Time-JEPA
        & Time-JEPA $-$ Video FT \\
        \midrule
        Object Permanence & 384 & 56.510 & 65.885 & \textbf{70.052} & $+4.167$ \\
        Continuity        & 512 & 55.078 & 55.469 & \textbf{56.250} & $+0.781$ \\
        Color Constancy   & 256 & 32.422 & \textbf{58.984} & 53.125 & $-5.859$ \\
        Support           & 128 & \textbf{100.000} & 99.219 & 97.656 & $-1.563$ \\
        Inertia           & 512 & 36.328 & 38.672 & \textbf{41.016} & $+2.344$ \\
        Gravity           & 640 & 42.500 & 44.219 & \textbf{46.250} & $+2.031$ \\
        Solidity          & 256 & 82.422 & 80.469 & \textbf{86.719} & $+6.250$ \\
        Collision         & 128 & 0.000 & 1.563 & \textbf{10.938} & $+9.375$ \\
        \bottomrule
    \end{tabular}
\end{table}

Time-JEPA improves six of eight properties over the video-fine-tuned control.

\FloatBarrier
\subsection{Physion Object-Contact Prediction}
\label{app:physion_protocol}

Physion v1.5 Object Contact Prediction asks whether a designated agent and
patient object will contact after an observed RGB prefix.
The compact split contains 5,608 training and 1,035 test stimuli across
\emph{collide}, \emph{drop}, \emph{support}, \emph{link},
\emph{roll}, \emph{contain}, and \emph{dominoes}. The test positive rate
is 50.531\%.
Ordinary stimuli use source frames $[0,15,30,45]$ at 150-ms spacing, while the
fast-collision case uses $[0,0,0,15]$.

We spatially average the frozen target encoder's final-layer tokens within
each of two temporal tubelets and flatten the $[2,1408]$ array into a 2,816D
feature. Model selection includes feature standardization.
In each cross-validation fold, we compute every feature dimension's mean and
standard deviation from that fold's training portion and apply them to its
validation portion.
Five-fold cross-validation preserves the label ratio in every fold and selects
$C\in\{10^{-6},10^{-5},0.01,0.1,1,5,10,20\}$ by mean validation Accuracy.
$C$ is the inverse regularization strength, so larger values impose a weaker
penalty.
After selecting $C$, we recompute the standardization statistics on all
permitted training examples, refit the L2-regularized logistic classifier on
those standardized features, and evaluate it once on the corresponding test
set using the same final training statistics.

The \emph{all-scenario readout} fits one classifier using all seven training
scenarios and evaluates all test scenarios. The
\emph{leave-one-scenario-out transfer} protocol fits seven classifiers, each
excluding one scenario from readout training and evaluating only that scenario.
Supervised probe training excludes the scenario evaluated in each transfer task.

We fit all probes on chronological features. The temporally shuffled input
condition reorders the same four test frames and retains the labels and fitted
classifiers. This condition tests sensitivity to an OOD change in frame order.

\subsection{Physion Macro and Overall Results}

\begin{table}[htbp]
    \caption{\textbf{Physion scenario-macro Accuracy and AUROC.}
    We average percentages equally over seven scenarios. Chronological and
    temporally shuffled inputs reuse the same fitted probes.
    Bold marks the highest value per protocol and metric.}
    \label{tab:app_physion_macro}
    \centering
    \fontsize{8.5}{10.5}\selectfont
    \setlength{\tabcolsep}{4pt}
    \begin{tabular}{@{}cccccc@{}}
        \toprule
        & & \multicolumn{2}{c}{Chronological} & \multicolumn{2}{c}{Temporally shuffled} \\
        \cmidrule(lr){3-4}\cmidrule(lr){5-6}
        Configuration & Readout setting
        & Acc. $\uparrow$ & AUROC $\uparrow$ & Acc. $\uparrow$ & AUROC $\uparrow$ \\
        \midrule
        Original & All-scenario & 71.511 & 79.573 & 61.490 & 67.831 \\
        Video FT & All-scenario & \textbf{75.063} & 81.361 & \textbf{64.578} & 72.270 \\
        Time-JEPA & All-scenario & 75.062 & \textbf{82.543} & 63.470 & \textbf{72.523} \\
        \addlinespace[2pt]
        Original & \shortstack{Leave-one-\\scenario-out} & 59.368 & 63.076 & \textbf{54.504} & 58.146 \\
        Video FT & \shortstack{Leave-one-\\scenario-out} & 57.952 & 62.028 & 54.415 & 58.647 \\
        Time-JEPA & \shortstack{Leave-one-\\scenario-out} & \textbf{60.897} & \textbf{66.024} & 54.010 & \textbf{60.066} \\
        \bottomrule
    \end{tabular}
\end{table}

\begin{table}[!htbp]
    \caption{\textbf{Physion overall Accuracy and AUROC.}
    Percentages pool all 1,035 test stimuli. Larger scenarios receive more
    weight than in Table~\ref{tab:app_physion_macro}.
    Chronological and temporally shuffled inputs use the same fitted probes.
    Bold marks the highest value per condition and metric.}
    \label{tab:app_physion_overall}
    \centering
    \fontsize{8.5}{10.5}\selectfont
    \setlength{\tabcolsep}{2.2pt}
    \renewcommand{\arraystretch}{1.06}
    \begin{tabular*}{\textwidth}{@{\extracolsep{\fill}}ccccccccc@{}}
        \toprule
         & \multicolumn{2}{c}{\shortstack{All-scenario\\chronological}} & \multicolumn{2}{c}{\shortstack{All-scenario\\shuffled}} & \multicolumn{2}{c}{\shortstack{Leave-one-scenario-out\\chronological}} & \multicolumn{2}{c}{\shortstack{Leave-one-scenario-out\\shuffled}} \\
        \cmidrule(lr){2-3}\cmidrule(lr){4-5}\cmidrule(lr){6-7}\cmidrule(lr){8-9}
        Configuration & Acc. $\uparrow$ & AUROC $\uparrow$ & Acc. $\uparrow$ & AUROC $\uparrow$ & Acc. $\uparrow$ & AUROC $\uparrow$ & Acc. $\uparrow$ & AUROC $\uparrow$ \\
        \midrule
        Original & 71.401 & 80.100 & 61.739 & 67.737 & 59.034 & 61.430 & \textbf{54.106} & 55.592 \\
        Video FT & 74.686 & 82.029 & \textbf{64.638} & 71.675 & 57.681 & 60.914 & 54.010 & 57.212 \\
        Time-JEPA & \textbf{74.783} & \textbf{83.494} & 63.768 & \textbf{72.366} & \textbf{60.386} & \textbf{64.056} & 54.010 & \textbf{57.748} \\
        \bottomrule
    \end{tabular*}
\end{table}

Table~\ref{tab:app_physion_macro} weights scenarios equally, whereas
Table~\ref{tab:app_physion_overall} pools all stimuli. Under chronological
all-scenario evaluation, Time-JEPA changes Accuracy by only $-0.001$ macro
and $+0.097$ pooled points, while increasing AUROC by 1.182 and 1.465 points. Under leave-one-scenario-out transfer, Accuracy/AUROC improve by
$2.945/3.997$ macro points and $2.705/3.142$ pooled points.

For leave-one-scenario-out transfer, the matched Time-JEPA difference changes from
$+2.945/+3.997$ Accuracy/AUROC points on chronological inputs to
$-0.405/+1.419$ points after shuffling.
This drop suggests that the held-out-scenario readout uses temporal information
in addition to static appearance cues.

\begin{table}[!htbp]
    \caption{\textbf{Physion results by scenario with chronological input.}
    Changes give Time-JEPA minus Video FT in percentage points.
    The final two columns give Time-JEPA Accuracy and AUROC in percent.
    Bold marks the largest value in each column.}
    \label{tab:app_physion_scenarios}
    \centering
    \fontsize{8.5}{10.5}\selectfont
    \setlength{\tabcolsep}{2.2pt}
    \renewcommand{\arraystretch}{1.06}
    \begin{tabular*}{\textwidth}{@{\extracolsep{\fill}}ccccccc@{}}
        \toprule
         & \multicolumn{2}{c}{All-scenario change} & \multicolumn{2}{c}{Leave-one-scenario-out change} & \multicolumn{2}{c}{Time-JEPA leave-one-scenario-out} \\
        \cmidrule(lr){2-3}\cmidrule(lr){4-5}\cmidrule(lr){6-7}
        Scenario ($n$) & $\Delta$Acc. $\uparrow$ & $\Delta$AUROC $\uparrow$ & $\Delta$Acc. $\uparrow$ & $\Delta$AUROC $\uparrow$ & Acc. $\uparrow$ & AUROC $\uparrow$ \\
        \midrule
        Collide (105) & \(-\)1.905 & +0.472 & +6.667 & \textbf{+10.704} & 66.667 & 71.662 \\
        Contain (153) & +1.961 & +3.179 & +5.882 & +7.915 & 61.438 & 68.256 \\
        Dominoes (159) & \(-\)1.887 & +0.712 & +4.403 & +9.525 & 51.572 & 58.244 \\
        Drop (157) & \(-\)1.274 & +0.552 & \(-\)4.459 & \(-\)3.117 & 58.599 & 64.188 \\
        Link (157) & +1.911 & \(-\)1.186 & \(-\)1.274 & +1.186 & 56.688 & 63.759 \\
        Roll (157) & \textbf{+2.548} & \textbf{+3.805} & +1.911 & \(-\)1.366 & 49.682 & 47.154 \\
        Support (147) & \(-\)1.361 & +0.741 & \textbf{+7.483} & +3.130 & \textbf{81.633} & \textbf{88.907} \\
        \bottomrule
    \end{tabular*}
\end{table}

Under the all-scenario readout, Time-JEPA improves AUROC over the video-fine-tuned control in six of
seven scenarios while Accuracy changes have mixed signs.

Under leave-one-scenario-out transfer, Time-JEPA improves both metrics on
collide, contain, dominoes, and support. Both decrease on drop, while link and
roll show mixed changes. The scenario-macro result weights all seven
phenomena equally.

\FloatBarrier
\section{Design Ablations}
\label{app:ablations}
\setcounter{table}{0}
\setcounter{figure}{0}

The human motion ablations cover three design questions: temporal aggregation
in the auxiliary head, downstream probe capacity, and supervision breadth.
Unless stated otherwise, the evaluation uses the same nine current-state
$\mathrm{ID}_{\mathrm{var}}$ variables, five geometric
$\mathrm{OOD}_{\mathrm{var}}$ variables, seen exocentric
$\mathrm{ID}_{\mathrm{cond}}$ views, and held-out exocentric \texttt{view03}
$\mathrm{OOD}_{\mathrm{cond}}$ view as the main protocol.
We average metrics equally over views and variables. We remove each auxiliary
head and train a new probe on the frozen encoder.

\subsection{Temporal Aggregation in the Auxiliary Head}
\label{sec:ablation_temporal_head}

The main pointwise nonlinear auxiliary head applies the same nonlinear channel
projection independently at each temporal position. We compare it with near-parameter-matched heads (approximately
2M trainable parameters). Local convolution mixes three neighboring positions,
and dense multi-scale convolution uses temporal spans of 3, 5, and 9. Global
attention mixes all 32 temporal positions, local attention uses a
nine-position window, and the hybrid uses four local and four global attention
heads.
All variants retain the same final-feature interface, 9D tubelet-level state
values, training protocol, and pointwise nonlinear evaluation probe.

\begin{figure}[t]
    \centering
    \includegraphics[width=\textwidth]{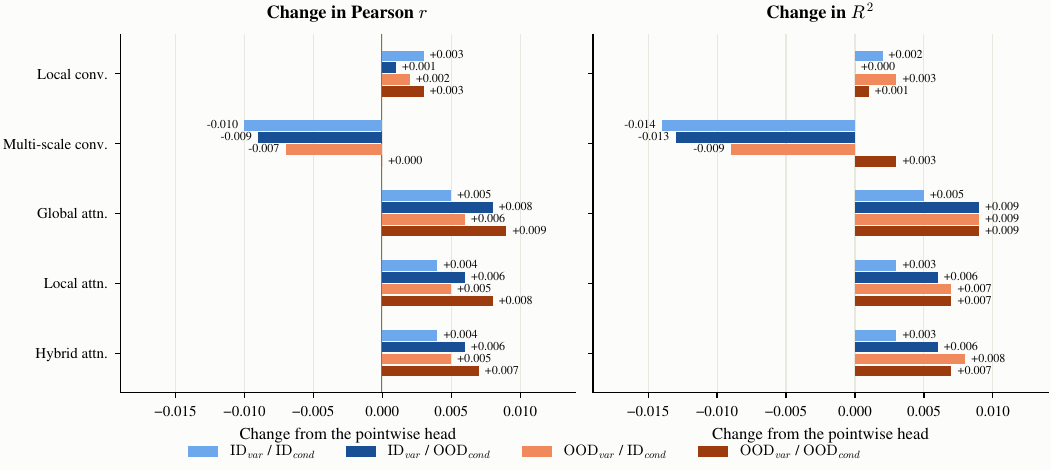}
    \caption{\textbf{human motion auxiliary-head temporal aggregation.}
    Grouped bars show changes from the pointwise nonlinear auxiliary head on the
    four variable--condition panels. Labels give each three-decimal change.
    The four solid colors denote the four variable--condition combinations.
    The left and right panels report Pearson $r$
    and $R^2$, respectively, and the gray vertical reference marks zero change.
    All frozen encoders use the matched pointwise nonlinear probe.}
    \label{fig:ablation_temporal_head}
\end{figure}

Global attention gives the highest Pearson $r$ and $R^2$ on all four panels
(Figure~\ref{fig:ablation_temporal_head}), exceeding the pointwise nonlinear
head by $0.005$--$0.009$ in each metric. Local and hybrid attention give similar
small gains, while dense multi-scale convolution degrades most panels.
The encoder's spatiotemporal self-attention already mixes information across
temporal positions, so further aggregation in the auxiliary head may be largely
redundant in this setting. We therefore retain the pointwise nonlinear head.

\subsection{Probe-Capacity Check}
\label{sec:ablation_probe_capacity}

We compare the pointwise nonlinear probe with a linear probe that spatially
pools each temporal feature and applies \texttt{LayerNorm + Linear}.
The linear probe has 15,497 parameters for $\mathrm{ID}_{\mathrm{var}}$ and
9,861 for $\mathrm{OOD}_{\mathrm{var}}$. The pointwise nonlinear probe has
1,996,553 and 1,990,917
parameters, respectively. Both operate independently at each temporal position.

\begin{table}[htbp]
    \caption{\textbf{human motion probe-capacity check.}
    Changes compare Time-JEPA with Video FT within each probe family.
    We report nRMSE reduction as Video FT minus Time-JEPA.
    Variable and condition notation follows Section~\ref{sec:protocol_domains}.
    $\mathrm{ID}_{\mathrm{cond}}$ averages seen exocentric views, and
    $\mathrm{OOD}_{\mathrm{cond}}$ uses held-out exocentric \texttt{view03}.
    Bold marks the larger matched gain.}
    \label{tab:ablation_probe_capacity}
    \centering
    \fontsize{8.5}{10.5}\selectfont
    \setlength{\tabcolsep}{5pt}
    \begin{tabular}{@{}>{\centering\arraybackslash}p{0.25\textwidth}*{6}{c}@{}}
        \toprule
        & \multicolumn{2}{c}{$\Delta r\uparrow$}
        & \multicolumn{2}{c}{$\Delta R^2\uparrow$}
        & \multicolumn{2}{c}{nRMSE reduction $\uparrow$} \\
        \cmidrule(lr){2-3}\cmidrule(lr){4-5}\cmidrule(l){6-7}
        Panel & Linear & \shortstack{Pointwise\\nonlinear} & Linear & \shortstack{Pointwise\\nonlinear} & Linear & \shortstack{Pointwise\\nonlinear} \\
        \midrule
        $\mathrm{ID}_{\mathrm{var}}$, $\mathrm{ID}_{\mathrm{cond}}$  & 0.066 & \textbf{0.086} & 0.084 & \textbf{0.114} & 0.048 & \textbf{0.068} \\
        $\mathrm{ID}_{\mathrm{var}}$, $\mathrm{OOD}_{\mathrm{cond}}$ & 0.074 & \textbf{0.096} & 0.095 & \textbf{0.133} & 0.055 & \textbf{0.076} \\
        $\mathrm{OOD}_{\mathrm{var}}$, $\mathrm{ID}_{\mathrm{cond}}$  & 0.045 & \textbf{0.069} & 0.051 & \textbf{0.089} & 0.022 & \textbf{0.035} \\
        $\mathrm{OOD}_{\mathrm{var}}$, $\mathrm{OOD}_{\mathrm{cond}}$ & 0.046 & \textbf{0.063} & 0.054 & \textbf{0.090} & 0.023 & \textbf{0.034} \\
        \bottomrule
    \end{tabular}
\end{table}

The linear probe preserves the Time-JEPA advantage on all four panels and
all three metrics (Table~\ref{tab:ablation_probe_capacity}).
Its matched gains span $0.045$--$0.074$ in Pearson and $0.051$--$0.095$ in
$R^2$, with nRMSE reductions of $0.022$--$0.055$.
The pointwise probe yields larger effects but the same encoder ordering.
The continued gains with a much smaller linear probe suggest that
they reflect the frozen encoder representation rather than the pointwise probe
architecture in this setting.

\subsection{Supervision Breadth}
\label{sec:ablation_supervision}

We compare aligned supervision on pelvis speed alone (1D), five motion-speed
quantities, and the full 9D set including four flexion angles.

\begin{table}[!htbp]
    \caption{\textbf{human motion supervision breadth.}
    We average Pearson $r$ and $R^2$ equally over views, then variables.
    Variable and condition notation follows Section~\ref{sec:protocol_domains}.
    Bold marks the best supervision condition per panel and metric.}
    \label{tab:ablation_supervision_macro}
    \centering
    \fontsize{8}{9.5}\selectfont
    \setlength{\tabcolsep}{1.2pt}
    \renewcommand{\arraystretch}{1.06}
    \begin{tabular*}{\textwidth}{@{\extracolsep{\fill}}>{\centering\arraybackslash}p{0.225\textwidth}cccccccc@{}}
        \toprule
         & \multicolumn{2}{c}{\shortstack{$\mathrm{ID}_{\mathrm{var}}$\\$\mathrm{ID}_{\mathrm{cond}}$}}
         & \multicolumn{2}{c}{\shortstack{$\mathrm{ID}_{\mathrm{var}}$\\$\mathrm{OOD}_{\mathrm{cond}}$}}
         & \multicolumn{2}{c}{\shortstack{$\mathrm{OOD}_{\mathrm{var}}$\\$\mathrm{ID}_{\mathrm{cond}}$}}
         & \multicolumn{2}{c}{\shortstack{$\mathrm{OOD}_{\mathrm{var}}$\\$\mathrm{OOD}_{\mathrm{cond}}$}} \\
        \cmidrule(lr){2-3}\cmidrule(lr){4-5}\cmidrule(lr){6-7}\cmidrule(lr){8-9}
        Encoder supervision & $r\uparrow$ & $R^2\uparrow$ & $r\uparrow$ & $R^2\uparrow$ & $r\uparrow$ & $R^2\uparrow$ & $r\uparrow$ & $R^2\uparrow$ \\
        \midrule
        Video FT & 0.503 & 0.231 & 0.486 & 0.199 & 0.434 & 0.133 & 0.452 & 0.135 \\
        Full 9D aligned & \textbf{0.589} & \textbf{0.345} & \textbf{0.581} & \textbf{0.332} & \textbf{0.504} & \textbf{0.222} & \textbf{0.515} & \textbf{0.225} \\
        Pelvis-only 1D aligned & 0.546 & 0.279 & 0.534 & 0.256 & 0.453 & 0.156 & 0.469 & 0.157 \\
        Motion-speed 5D aligned & 0.551 & 0.284 & 0.537 & 0.259 & 0.452 & 0.155 & 0.470 & 0.158 \\
        \bottomrule
    \end{tabular*}
\end{table}

\begin{table}[htbp]
    \caption{\textbf{Semantic decomposition of reduced supervision.}
    Changes give reduced minus full-aligned performance.
    Included/Excluded indicate whether a full-model
    $\mathrm{ID}_{\mathrm{var}}$ variable enters
    the reduced objective.}
    \label{tab:ablation_supervision_semantics}
    \centering
    \small
    \setlength{\tabcolsep}{3pt}
    \begin{tabular}{@{}cccccc@{}}
        \toprule
        Reduced supervision & Variable subset
        & $\mathrm{ID}_{\mathrm{cond}}$ $\Delta r\uparrow$
        & $\mathrm{ID}_{\mathrm{cond}}$ $\Delta R^2\uparrow$
        & $\mathrm{OOD}_{\mathrm{cond}}$ $\Delta r\uparrow$
        & $\mathrm{OOD}_{\mathrm{cond}}$ $\Delta R^2\uparrow$ \\
        \midrule
        Pelvis-only 1D & Included & +0.013 & +0.017 & +0.004 & \(-\)0.001 \\
        & Excluded & \(-\)0.050 & \(-\)0.077 & \(-\)0.054 & \(-\)0.085 \\
        Motion-speed 5D & Included & +0.004 & +0.005 & \(-\)0.001 & \(-\)0.001 \\
        & Excluded & \(-\)0.091 & \(-\)0.144 & \(-\)0.098 & \(-\)0.164 \\
        \bottomrule
    \end{tabular}
\end{table}

Both reduced aligned encoders improve both metrics over the video-fine-tuned control on all four
standard panels, but remain below full aligned supervision
(Table~\ref{tab:ablation_supervision_macro}).
The decomposition in Table~\ref{tab:ablation_supervision_semantics} shows that
focused supervision remains close to full supervision on included semantics.
Motion-only supervision loses most on excluded flexion variables, and all five
$\mathrm{OOD}_{\mathrm{var}}$ geometric variables also decline.
The 1D and 5D controls have similar scores. The larger gains from full
supervision suggest that coverage of complementary physical quantities
supports transfer.

\FloatBarrier
\section{Sensitivity Analysis}
\label{app:sensitivity_analysis}
\setcounter{table}{0}
\setcounter{figure}{0}

We test how temporal alignment and sensor noise affect the
human motion representation. Each experiment perturbs only the physical time
series used for encoder fine-tuning. Probe training and evaluation use clean
state labels. All encoders use the same full 9D state-variable set, training
protocol, and pointwise nonlinear auxiliary head. We remove that head and train
a matched pointwise nonlinear Readout on each frozen encoder. The results
average views equally before averaging the nine variables equally and report
seen exocentric views as $\mathrm{ID}_{\mathrm{cond}}$ and held-out exocentric
\texttt{view03} as $\mathrm{OOD}_{\mathrm{cond}}$.

\subsection{Temporal Misalignment}
\label{sec:sensitivity_misalignment}

The fixed-offset condition pairs each video time with the state sequence
0.8~s later. The cross-take condition pairs every video window with a 9D state
sequence from a different take through a one-to-one permutation. The latter
condition preserves the dataset-level state-value and validity-mask multisets
but removes local video--state correspondence.

\begin{table}[!htbp]
    \caption{\textbf{human motion sensitivity to temporal misalignment.}
    The pointwise nonlinear Readout uses clean labels for every frozen encoder.
    Metrics average views before the nine variables. Bold marks the best
    unrounded score in each row.}
    \label{tab:sensitivity_alignment}
    \centering
    \fontsize{8.5}{10.2}\selectfont
    \setlength{\tabcolsep}{5pt}
    \begin{tabular*}{\textwidth}{@{\extracolsep{\fill}}cccccc@{}}
        \toprule
        Condition & Metric & Video FT & Clean alignment & 0.8~s offset
        & Cross-take \\
        \midrule
        $\mathrm{ID}_{\mathrm{cond}}$ & $r\uparrow$ & 0.503 & \textbf{0.589} & 0.575 & 0.510 \\
        & $R^2\uparrow$ & 0.231 & \textbf{0.345} & 0.329 & 0.242 \\
        & nRMSE $\downarrow$ & 0.742 & \textbf{0.674} & 0.683 & 0.735 \\
        \midrule
        $\mathrm{OOD}_{\mathrm{cond}}$ & $r\uparrow$ & 0.486 & \textbf{0.581} & 0.570 & 0.498 \\
        & $R^2\uparrow$ & 0.199 & \textbf{0.332} & 0.320 & 0.219 \\
        & nRMSE $\downarrow$ & 0.761 & \textbf{0.685} & 0.692 & 0.750 \\
        \bottomrule
    \end{tabular*}
\end{table}

Cross-take misalignment causes a larger decline than the 0.8~s offset and
approaches the video-fine-tuned control. Across Pearson $r$ and $R^2$, it
retains only about $5$--$15\%$ of the clean-alignment gain over Video FT.

\subsection{Sensor Noise}
\label{sec:sensitivity_noise}

We add zero-mean Gaussian noise to each fine-tuning state variable with a
standard deviation equal to that variable's clean training-set standard
deviation. This gives the noise a standard deviation of 1 after state-value
normalization.

\begin{table}[!htbp]
    \caption{\textbf{human motion sensitivity to Gaussian sensor noise.}
    The pointwise nonlinear Readout uses clean labels. Metrics average views
    before the nine variables. Bold marks the better unrounded score in each
    row.}
    \label{tab:sensitivity_noise}
    \centering
    \small
    \setlength{\tabcolsep}{12pt}
    \begin{tabular}{@{}cccc@{}}
        \toprule
        Condition & Metric & Clean & 1$\sigma$ Gaussian noise \\
        \midrule
        $\mathrm{ID}_{\mathrm{cond}}$ & $r\uparrow$ & \textbf{0.589} & 0.580 \\
        & $R^2\uparrow$ & \textbf{0.345} & 0.332 \\
        & nRMSE $\downarrow$ & \textbf{0.674} & 0.682 \\
        \midrule
        $\mathrm{OOD}_{\mathrm{cond}}$ & $r\uparrow$ & \textbf{0.581} & 0.570 \\
        & $R^2\uparrow$ & \textbf{0.332} & 0.315 \\
        & nRMSE $\downarrow$ & \textbf{0.685} & 0.695 \\
        \bottomrule
    \end{tabular}
\end{table}

The evaluation covers temporal pairing errors and noisy state values under
both seen and held-out views; each perturbation reduces performance.

\FloatBarrier
\section{Alternative Forms of Physical Supervision}
\label{app:alternative_supervision}
\setcounter{table}{0}
\setcounter{figure}{0}

We compare supervision on individual frames or frame pairs with Time-JEPA in
robot manipulation, then add an explicit physical-law constraint to Time-JEPA
in autonomous driving. All evaluations fit supervised, video-only probes on
frozen representations; $\mathrm{OOD}_{\mathrm{var}}$ variables still have probe
labels (Section~\ref{sec:protocol_domains}). Time-JEPA and Time-JEPA+Law use
their exponential-moving-average encoders. Future-state $R^2$ and nRMSE average
the five horizons $t+\{0.2,0.4,0.6,0.8,1.0\}$~s equally, where $t$ is the
final observed frame; nRMSE uses each variable's probe-training standard
deviation. Current- and future-state tables use one probe run per fixed
encoder; Table~\ref{tab:law_grounded} uses five paired probe seeds.

\subsection{Supervision on Individual Frames or Frame Pairs}
\label{app:frame_supervision}

\paragraph{Training objectives.}
We adapt PSG-JEPA~\citep{yan2026psgjepa} to the released V-JEPA~2.1 ViT-g/16
image branch. Both variants independently encode frames, pool spatial tokens,
and project them; both retain the action-conditioned latent predictor and
SIGReg regularizer. Each 64-frame window supplies 16 four-frame sequences.

\emph{Static} regresses the aligned 18D state from each frame.
\emph{Full} adds two heads on paired frame latents: one predicts changes in ten
configuration coordinates (two waist components, six end-effector positions,
and two gripper positions) across all frame pairs in each four-frame sequence,
averaging the three separations equally. The other predicts base translation
and angular speed and six end-effector velocities from adjacent pairs at the
earlier frame. All terms use mean squared error on valid, standardized values:
\begin{align}
    \mathcal L_{\mathrm{Static}}
    &= \mathcal L_{\mathrm{forward}}
       + 0.09\mathcal L_{\mathrm{SIGReg}}
       + 0.1\mathcal L_{\mathrm{state}}, \\
    \mathcal L_{\mathrm{Full}}
    &= \mathcal L_{\mathrm{Static}}
       + 0.1\bigl(\mathcal L_{\mathrm{change}}
                  + \mathcal L_{\mathrm{velocity}}\bigr).
    \label{eq:psg_supervision_objectives}
\end{align}
Here, $\mathcal L_{\mathrm{forward}}$ is latent next-step prediction.
Time-JEPA instead supervises temporally contextualized video features at each
aligned tubelet, alongside masked spatiotemporal JEPA prediction.

\paragraph{Matched data and probes.}
All models start from the released checkpoint and use the same 5,000 physical
windows rendered in three views (15,000 view-specific windows), 64 frames at 5 FPS,
$384\times384$ resolution, 18D state panel and normalization, 20 epochs,
global batch size 64, and encoder seed 239. Static and Full use their final
checkpoints; the future-state Time-JEPA encoder is single-domain.

We remove training heads and freeze the representations. Static and Full keep
their image encoder and projector, averaging adjacent frames into 32 temporal
positions. Matched pointwise nonlinear probes predict all five future horizons
jointly from the final feature. They use 15 epochs, learning rate $10^{-4}$,
global batch size 16, seed 0, and validation-loss selection. Current-state
train/validation/test counts are 4,500/1,500/1,000 windows per view; requiring
future labels gives 4,440/1,476/989.

\paragraph{Variables and aggregation.}
We evaluate 18 $\mathrm{ID}_{\mathrm{var}}$ and 8 $\mathrm{OOD}_{\mathrm{var}}$
variables (Appendix~\ref{app:datasets}). Current-state $\mathrm{ID}_{\mathrm{cond}}$
averages three seen views; future-state $\mathrm{ID}_{\mathrm{cond}}$ uses the
head camera. $\mathrm{OOD}_{\mathrm{cond}}$ averages two held-out fisheye views. We average selected variables and views equally. Unlike
Table~\ref{tab:single_domain_main}.

\begin{table}[!htbp]
    \caption{\textbf{Current-state decoding in robot manipulation.}
    $\mathrm{ID}_{\mathrm{cond}}$ averages three seen views;
    $\mathrm{OOD}_{\mathrm{cond}}$ averages variables across two fisheye
    views. Higher $R^2$ and Pearson $r$ are better; bold marks the best score.}
    \label{tab:frame_supervision_current}
    \centering
    \small
    \setlength{\tabcolsep}{6pt}
    \begin{tabular*}{\textwidth}{@{\extracolsep{\fill}}cccccccc@{}}
        \toprule
        & & \multicolumn{2}{c}{Static} & \multicolumn{2}{c}{Full}
        & \multicolumn{2}{c}{Time-JEPA} \\
        \cmidrule(lr){3-4}\cmidrule(lr){5-6}\cmidrule(lr){7-8}
        Condition & Variables & $R^2\uparrow$ & $r\uparrow$
        & $R^2\uparrow$ & $r\uparrow$ & $R^2\uparrow$ & $r\uparrow$ \\
        \midrule
        \multirow{2}{*}{$\mathrm{ID}_{\mathrm{cond}}$} & $\mathrm{ID}_{\mathrm{var}}$ & 0.273 & 0.493 & 0.397 & 0.628 & \textbf{0.725} & \textbf{0.845} \\
        & $\mathrm{OOD}_{\mathrm{var}}$ & 0.270 & 0.512 & 0.373 & 0.602 & \textbf{0.568} & \textbf{0.741} \\
        \midrule
        \multirow{2}{*}{$\mathrm{OOD}_{\mathrm{cond}}$} & $\mathrm{ID}_{\mathrm{var}}$ & \(-\)0.073 & 0.059 & \(-\)0.137 & 0.118 & \textbf{0.269} & \textbf{0.688} \\
        & $\mathrm{OOD}_{\mathrm{var}}$ & \(-\)0.061 & 0.056 & \(-\)0.111 & 0.123 & \textbf{0.205} & \textbf{0.555} \\
        \bottomrule
    \end{tabular*}
\end{table}

\begin{table}[!htbp]
    \caption{\textbf{Future-state prediction in robot manipulation.}
    $\mathrm{ID}_{\mathrm{cond}}$ uses the head camera;
    $\mathrm{OOD}_{\mathrm{cond}}$ averages variables across two fisheye
    views. Five horizons span 0.2--1.0~s. Higher $R^2$ and lower nRMSE are
    better; bold marks the best score.}
    \label{tab:frame_supervision_future}
    \centering
    \small
    \setlength{\tabcolsep}{6pt}
    \begin{tabular*}{\textwidth}{@{\extracolsep{\fill}}cccccccc@{}}
        \toprule
        & & \multicolumn{2}{c}{Static} & \multicolumn{2}{c}{Full}
        & \multicolumn{2}{c}{Time-JEPA} \\
        \cmidrule(lr){3-4}\cmidrule(lr){5-6}\cmidrule(lr){7-8}
        Condition & Variables & $R^2\uparrow$ & nRMSE $\downarrow$
        & $R^2\uparrow$ & nRMSE $\downarrow$ & $R^2\uparrow$ & nRMSE $\downarrow$ \\
        \midrule
        \multirow{2}{*}{$\mathrm{ID}_{\mathrm{cond}}$} & $\mathrm{ID}_{\mathrm{var}}$ & 0.217 & 0.853 & 0.302 & 0.804 & \textbf{0.506} & \textbf{0.647} \\
        & $\mathrm{OOD}_{\mathrm{var}}$ & 0.167 & 0.900 & 0.243 & 0.857 & \textbf{0.379} & \textbf{0.771} \\
        \midrule
        \multirow{2}{*}{$\mathrm{OOD}_{\mathrm{cond}}$} & $\mathrm{ID}_{\mathrm{var}}$ & \(-\)0.081 & 1.014 & \(-\)0.116 & 1.029 & \textbf{0.100} & \textbf{0.919} \\
        & $\mathrm{OOD}_{\mathrm{var}}$ & \(-\)0.053 & 1.012 & \(-\)0.077 & 1.023 & \textbf{0.047} & \textbf{0.951} \\
        \bottomrule
    \end{tabular*}
\end{table}

\paragraph{Results.}
Time-JEPA leads on both metrics in all eight settings
(Tables~\ref{tab:frame_supervision_current}--\ref{tab:frame_supervision_future})
and at each future horizon under both condition groups. Full improves on Static
in all four $\mathrm{ID}_{\mathrm{cond}}$ settings but lowers $R^2$ in all
four $\mathrm{OOD}_{\mathrm{cond}}$ settings; its current-state Pearson gains
under held-out views coexist with poorer absolute fit. Time-JEPA's
$\mathrm{OOD}_{\mathrm{var}}$/$\mathrm{OOD}_{\mathrm{cond}}$ $R^2$ reaches
$-0.012$ at 1.0~s despite its positive five-horizon average.
This advantage is consistent with the value of supervising evolving physical
state on features that incorporate the video's full temporal context.

\subsection{Adding Explicit Physical-Law Constraints}
\label{app:law_supervision}

\paragraph{Training objective.}
We augment the single-domain autonomous driving Time-JEPA with a longitudinal
kinematic constraint, denoting the variant \emph{Time-JEPA+Law}.
It reuses the existing speed, acceleration, and yaw-rate auxiliary head and
adds no trainable parameters. We convert its speed and acceleration back to physical units using the original training-set statistics, denoting them
$\widehat v_k$ and $\widehat a_k$ at tubelet $k$. For adjacent tubelets, we
penalize the trapezoidal integration residual
\begin{equation}
    r_k = \widehat v_{k+1}-\widehat v_k
        - \frac{\Delta t_k}{2}
          \bigl(\widehat a_k+\widehat a_{k+1}\bigr).
    \label{eq:training_law_residual}
\end{equation}
We obtain $\Delta t_k$ from decoded video timestamps at tubelet centers.
Valid pairs require speed and acceleration on all four frames, increasing
frame times, and $0<\Delta t_k\leq0.6$~s. State-label alignment and validity
masks remain unchanged.

For the set $\mathcal P$ of valid pairs, the added term and total objective are
\begin{align}
    \mathcal L_{\mathrm{law}}
    &= \frac{1}{|\mathcal P|}\sum_{k\in\mathcal P}
       \mathrm{SmoothL1}_{0.5}\!\left(\frac{r_k}{s_{\Delta v}},0\right),
       \label{eq:training_law_loss}\\
    \mathcal L_{\mathrm{Time\text{-}JEPA+Law}}
    &= \Ljepa + \lambda\bigl(\Ltime
                        + \alpha\mathcal L_{\mathrm{law}}\bigr),
       \qquad \alpha=0.5.
       \label{eq:time_jepa_law_objective}
\end{align}
The fixed $s_{\Delta v}$ is the sample standard deviation of valid ground-truth
speed differences from one clip per each of 6,305 training videos.
The original $\lambda=0.2$ and 250-step warmup apply to both auxiliary terms;
the law term's final effective weight is 0.1.

\paragraph{Evaluation protocol.}
The models share the training split, backbone, state variables, normalization,
schedule, and final 20-epoch checkpoint. Matched probes use 15 epochs, global
batch size 32, seed 239, and validation-loss selection on 1,349 train and 449
validation videos. Tests contain 900 $\mathrm{ID}_{\mathrm{cond}}$ and 521
rainy $\mathrm{OOD}_{\mathrm{cond}}$ videos. Both tasks evaluate the three
$\mathrm{ID}_{\mathrm{var}}$ and three $\mathrm{OOD}_{\mathrm{var}}$ variables
in Appendix~\ref{app:datasets}. The Time-JEPA current-state probes are reused;
future-state probes use the same single-domain encoder.

\begin{table}[!htbp]
    \caption{\textbf{Current-state decoding in autonomous driving.}
    Each variable group averages three variables; $\mathrm{OOD}_{\mathrm{cond}}$
    is rainy weather. Higher $R^2$ and Pearson $r$ are better; bold marks the
    better score.}
    \label{tab:law_supervision_current}
    \centering
    \small
    \setlength{\tabcolsep}{9pt}
    \begin{tabular*}{\textwidth}{@{\extracolsep{\fill}}cccccc@{}}
        \toprule
        & & \multicolumn{2}{c}{Time-JEPA}
        & \multicolumn{2}{c}{Time-JEPA+Law} \\
        \cmidrule(lr){3-4}\cmidrule(lr){5-6}
        Condition & Variables & $R^2\uparrow$ & $r\uparrow$
        & $R^2\uparrow$ & $r\uparrow$ \\
        \midrule
        \multirow{2}{*}{$\mathrm{ID}_{\mathrm{cond}}$} & $\mathrm{ID}_{\mathrm{var}}$ & \textbf{0.849} & \textbf{0.916} & 0.809 & 0.890 \\
        & $\mathrm{OOD}_{\mathrm{var}}$ & \textbf{0.846} & \textbf{0.920} & 0.831 & 0.912 \\
        \midrule
        \multirow{2}{*}{$\mathrm{OOD}_{\mathrm{cond}}$} & $\mathrm{ID}_{\mathrm{var}}$ & \textbf{0.824} & \textbf{0.902} & 0.780 & 0.878 \\
        & $\mathrm{OOD}_{\mathrm{var}}$ & \textbf{0.876} & \textbf{0.937} & 0.863 & 0.930 \\
        \bottomrule
    \end{tabular*}
\end{table}

\begin{table}[!htbp]
    \caption{\textbf{Future-state prediction in autonomous driving.}
    Scores average three variables per group and five horizons equally;
    $\mathrm{OOD}_{\mathrm{cond}}$ is rainy weather. Higher $R^2$ and lower
    nRMSE are better; bold marks the better score.}
    \label{tab:law_supervision_future}
    \centering
    \small
    \setlength{\tabcolsep}{9pt}
    \begin{tabular*}{\textwidth}{@{\extracolsep{\fill}}cccccc@{}}
        \toprule
        & & \multicolumn{2}{c}{Time-JEPA}
        & \multicolumn{2}{c}{Time-JEPA+Law} \\
        \cmidrule(lr){3-4}\cmidrule(lr){5-6}
        Condition & Variables & $R^2\uparrow$ & nRMSE $\downarrow$
        & $R^2\uparrow$ & nRMSE $\downarrow$ \\
        \midrule
        \multirow{2}{*}{$\mathrm{ID}_{\mathrm{cond}}$} & $\mathrm{ID}_{\mathrm{var}}$ & \textbf{0.719} & \textbf{0.485} & 0.699 & 0.503 \\
        & $\mathrm{OOD}_{\mathrm{var}}$ & \textbf{0.667} & \textbf{0.560} & 0.648 & 0.576 \\
        \midrule
        \multirow{2}{*}{$\mathrm{OOD}_{\mathrm{cond}}$} & $\mathrm{ID}_{\mathrm{var}}$ & \textbf{0.680} & \textbf{0.460} & 0.667 & 0.469 \\
        & $\mathrm{OOD}_{\mathrm{var}}$ & \textbf{0.604} & \textbf{0.529} & 0.594 & 0.536 \\
        \bottomrule
    \end{tabular*}
\end{table}

\paragraph{Results.}
At $\alpha=0.5$ and the fixed budget, the constraint lowers both current-state
metrics and future-state $R^2$, while raising future-state nRMSE, in all four
settings (Tables~\ref{tab:law_supervision_current}
and~\ref{tab:law_supervision_future}). Future nRMSE rises at every horizon
for both variable and condition groups.
\paragraph{Law satisfaction and trajectory reconstruction.}
Eleven independent scalar probes (seeds 0--4) on the fixed encoders decode
longitudinal/lateral velocity, yaw rate, six trajectory coordinates, speed,
and acceleration. We integrate motion states and compare endpoints with ground
truth as in Appendix~\ref{app:grounded_details}; integration-law AUROC compares
this reconstruction with five fixed alternatives. We also measure the mean
absolute speed--acceleration residual at $H\in\{0.5,1.0,2.0\}$~s, using
trapezoidal integration and 0.5~s tubelet intervals:
\begin{equation*}
    r(t,H)=\widehat v(t+H)-\widehat v(t)
        -\int_t^{t+H}\widehat a(\tau)\,d\tau.
\end{equation*}
\begin{table}[!htbp]
    \caption{\textbf{Law satisfaction and trajectory reconstruction in
    autonomous driving.} AUROC and trajectory MAE average 1.00, 2.00, and
    2.95~s; AUROC also averages five alternatives. Residuals are mean absolute
    values. Entries are mean $\pm$ sample standard deviation over five paired
    probe seeds on one fixed encoder per method; bold marks the better mean.}
    \label{tab:law_grounded}
    \centering
    \small
    \setlength{\tabcolsep}{3.5pt}
    \begin{tabular*}{\textwidth}{@{\extracolsep{\fill}}ccccc@{}}
        \toprule
        & \multicolumn{2}{c}{\textbf{$\mathrm{ID}_{\mathrm{cond}}$}}
        & \multicolumn{2}{c}{\textbf{$\mathrm{OOD}_{\mathrm{cond}}$}} \\
        \cmidrule(lr){2-3}\cmidrule(lr){4-5}
        Metric & Time-JEPA & Time-JEPA+Law & Time-JEPA & Time-JEPA+Law \\
        \midrule
        Integration-law AUROC $\uparrow$
            & \textbf{0.694\(\,\pm\,\)0.004} & 0.683\(\,\pm\,\)0.005
            & \textbf{0.672\(\,\pm\,\)0.009} & 0.661\(\,\pm\,\)0.010 \\
        Integrated trajectory MAE (m) $\downarrow$
            & \textbf{3.080\(\,\pm\,\)0.114} & 3.252\(\,\pm\,\)0.128
            & \textbf{3.572\(\,\pm\,\)0.312} & 3.875\(\,\pm\,\)0.359 \\
        Direct trajectory MAE (m) $\downarrow$
            & \textbf{3.482\(\,\pm\,\)0.068} & 3.534\(\,\pm\,\)0.066
            & \textbf{3.967\(\,\pm\,\)0.565} & 4.072\(\,\pm\,\)0.677 \\
        \addlinespace[3pt]
        $v$--$a$ residual, 0.5~s (m/s) $\downarrow$
            & 0.523\(\,\pm\,\)0.020 & \textbf{0.374\(\,\pm\,\)0.016}
            & 0.658\(\,\pm\,\)0.038 & \textbf{0.464\(\,\pm\,\)0.031} \\
        $v$--$a$ residual, 1.0~s (m/s) $\downarrow$
            & 0.695\(\,\pm\,\)0.027 & \textbf{0.531\(\,\pm\,\)0.024}
            & 0.790\(\,\pm\,\)0.042 & \textbf{0.594\(\,\pm\,\)0.038} \\
        $v$--$a$ residual, 2.0~s (m/s) $\downarrow$
            & 0.957\(\,\pm\,\)0.039 & \textbf{0.774\(\,\pm\,\)0.039}
            & 1.016\(\,\pm\,\)0.045 & \textbf{0.814\(\,\pm\,\)0.045} \\
        \bottomrule
    \end{tabular*}
\end{table}

The constraint lowers local residuals at all horizons, yet integration-law
AUROC falls and trajectory MAE rises by 5.6\%/8.5\% under
$\mathrm{ID}_{\mathrm{cond}}$/$\mathrm{OOD}_{\mathrm{cond}}$. This pattern is
consistent with explicit constraints favoring specified dynamics over other
video changes, as discussed in the Introduction.

\end{document}